\documentclass{bmvc2k}

\usepackage{mathtools} 
\usepackage{booktabs} 
\usepackage{bbm}
\usepackage{amssymb}
\usepackage{tikz}
\usepackage{amsmath,amsfonts}
\usepackage{algorithm}
\usepackage{algpseudocode}
\usepackage{enumitem}
\usepackage{array}
\usepackage{textcomp}
\usepackage{stfloats}
\usepackage{url}
\usepackage{verbatim}
\usepackage{xcolor}
\usepackage{hyperref}
\usepackage{lipsum}  
\usepackage{multicol}
\usepackage{multirow}
\usepackage{adjustbox}
\usepackage{graphicx}          
\usepackage{subcaption}

\usepackage{mathtools} 
\usepackage{booktabs} 
\usepackage{bbm}
\usepackage{amssymb}
\usepackage{tikz}
\usetikzlibrary{shapes.geometric} 

\usepackage{oplotsymbl}
\usepackage{wasysym}
\usepackage{float}

\title{SAE: Single Architecture Ensemble Neural Networks}

\addauthor{Martin Ferianc}{martin.ferianc.19@ucl.ac.uk}{1}
\addauthor{Hongxiang Fan}{h.fan17@imperial.ac.uk}{2}
\addauthor{Miguel Rodrigues}{m.rodrigues@ucl.ac.uk}{1}

\addinstitution{
 Department of Electronic and Electrical Engineering,\\ 
 University College London,\\
 London,\\
 United Kingdom
}
\addinstitution{
 Department of Computing,\\ 
 Imperial College London,\\
 London,\\
 United Kingdom
}

\runninghead{M. Ferianc, H. Fan, M.Rodrigues}{SAE: Single Architecture Ensemble}

\newcommand{\revision}[1]{#1}

\DeclareRobustCommand{\PentagonMarker}{
    \begin{tikzpicture}[baseline=-0.75ex] 
        \node[regular polygon, regular polygon sides=5, draw, fill, scale=8, inner sep=0pt] {};
    \end{tikzpicture}
}
\DeclareRobustCommand{\HexagonMarker}{
    \begin{tikzpicture}[baseline=-0.75ex] 
        \node[regular polygon, regular polygon sides=6, draw, fill, scale=8, inner sep=0pt, rotate=30] {};
    \end{tikzpicture}
}
\DeclareRobustCommand{\OctagonMarker}{
    \begin{tikzpicture}[baseline=-0.75ex] 
        \node[regular polygon, regular polygon sides=8, draw, fill, scale=8, inner sep=0pt] {};
    \end{tikzpicture}
}
\DeclareRobustCommand{\PlusMarker}{
    \begin{tikzpicture}[baseline=-0.75ex]
        \draw[line width=4pt] (0,0.15) -- (0,-0.15); 
        \draw[line width=4pt] (-0.15,0) -- (0.15,0); 
    \end{tikzpicture}
}
\DeclareRobustCommand{\TimesMarker}{
    \begin{tikzpicture}[baseline=-0.75ex, rotate=45]
        \draw[line width=4pt] (0,0.15) -- (0,-0.15); 
        \draw[line width=4pt] (-0.15,0) -- (0.15,0); 
    \end{tikzpicture}
}

\definecolor{Gray}{HTML}{595959}
\definecolor{Blue}{HTML}{2F528F}
\definecolor{Orange}{HTML}{C55A11}
\definecolor{Cyan}{HTML}{00BFBF}
\definecolor{Pink}{HTML}{BF00BF}

\begin{document}

\maketitle

\begin{abstract}
    Ensembles of separate neural networks (NNs) have shown superior accuracy and confidence calibration over single NN across tasks.
    To improve the hardware efficiency of ensembles of separate NNs, recent methods create ensembles within a single network via adding early exits or considering multi input multi output approaches.
    However, it is unclear which of these methods is the most effective for a given task, needing a manual and separate search through each method.
    Our novel Single Architecture Ensemble (SAE) framework enables an automatic and joint search through the early exit and multi input multi output configurations and their previously unobserved in-between combinations.
    SAE consists of two parts: a scalable search space that generalises the previous methods and their in-between configurations, and an optimisation objective that allows learning the optimal configuration for a given task.
    Our image classification and regression experiments show that with SAE we can automatically find diverse configurations that fit the task, achieving competitive accuracy or confidence calibration to baselines while reducing the compute operations or parameter count by up to $1.5{\sim}3.7\times$.
\end{abstract}
\section{Introduction}\label{sec:introduction}
\revision{Ensemble of independently trained NNs achieves superior accuracy and confidence calibration over a single NN across tasks~\citep{ovadia2019can, lakshminarayanan2017simple}.
In a naive ensemble, each NN is trained independently, which enables the ensemble members to learn independent features and make diverse predictions~\citep{fort2019deep}.
At evaluation time, each NN is queried with the same input, and the overall prediction is obtained by aggregating the individual predictions, leading to improved algorithmic performance.
The computational cost of training and querying an NN ensemble increases linearly with $N$, the number of NNs in the ensemble as $\mathcal{O}(N)$.
Therefore, as the number of NNs in the ensemble increases, the computational cost of training and querying the ensemble becomes prohibitive.}

Recent advances have introduced more hardware-efficient solutions, encapsulating the ensemble within a singular network architecture.
They do so by either adding early exits (EEs) to intermediate layers of the NN~\citep{antoran2020depth, qendro2021early, matsubara2022split, laskaridis2021adaptive}, feeding the NN multiple inputs and simultaneously expecting multiple outputs (MIMO)~\citep{havasi2020training}, or a blend of both: multi input massive multi output (MIMMO)~\citep{ferianc2023mimmo}.
These techniques mimic the naive ensemble of NNs by simultaneously training multiple predictors in one training round and NN architecture and collecting their predictions in a single forward pass for the final predictions.
These optimisations reduce the computational cost of training and inference to $\mathcal{O}(1)$, making the resultant ensembles hardware-efficient.

However, the current methodologies are fragmented and limited to their specific configurations e.g. only considering early exits - EE, multiple inputs and outputs - MIMO, or an extreme combination of both in MIMMO, and it is unclear which method is the most effective for a given task.
In this work, we hypothesise that it is necessary to unify these methods and introduce a search space that encompasses them and their previously unexplored in-between configurations to find the optimal configuration for a given task.
Therefore we introduce the novel Single Architecture Ensemble (SAE) framework consisting of a scalable search space and an optimisation objective that allows learning the optimal hardware-efficient ensemble configuration for a given task.
We summarise our contributions as follows:

\begin{itemize}[leftmargin=*, topsep=-0.12cm, itemsep=-0.1cm]
    \item[\textit{1.)}] A novel search space and problem formulation, encompassing the early exit, multiple inputs and outputs methods and their new in-between configurations.
    \item[\textit{2.)}] An optimisation objective that allows learning an optimal configuration for a given task, facilitating automatic and efficient exploration of the proposed search space.
    \item[\textit{3.)}] An empirical evaluation on image classification and regression tasks demonstrating that there is no one-size-fits-all hardware-efficient ensemble configuration out of the previous methods, confirming our hypothesis.
    However, with our novel SAE framework, we can find diverse configurations, achieving competitive accuracy or confidence calibration to baselines while reducing the compute operations or parameter count by up to $1.5{\sim}3.7\times$.
\end{itemize}
\section{Related Work}\label{sec:related_work}

Modern NNs are becoming increasingly complex, with billions of parameters and hundreds of layers~\cite{zhao2023survey}.
Nevertheless, it can be shown that most of these parameters are unused, indicating that the network is overparameterised~\citep{allen2019learning}.
Hardware-efficient NN ensembles utilise this overparameterisation to train a single NN that simulates the naive ensemble by providing the practitioner with diverse predictions in a single pass, thus improving the algorithmic performance while maintaining the hardware efficiency of a single NN.
These approaches include early exit (EE) networks~\citep{antoran2020depth, qendro2021early, matsubara2022split, laskaridis2021adaptive}, multi input multi output (MIMO) networks~\citep{havasi2020training}, and multi input massive multi output (MIMMO) networks~\citep{ferianc2023mimmo}.

We illustrate these approaches by considering a network consisting of $D\geq1$ layers, and processing $N\geq1$ inputs and outputs.
A standard single exit (SE) NN processes a single input and output, $N=1$, and gives a single prediction at the maximum depth of the network, $D$.

\textbf{Early Exit (EE)} EE networks~\citep{antoran2020depth, qendro2021early, matsubara2022split, laskaridis2021adaptive} introduce minimally-parametrised auxiliary exits at all or some intermediate layers within $D$ layers of the NN.
These exits predict the output at their respective depth, and the final prediction for a particular input is obtained by aggregating the predictions of all exits.
The EE network leverages models of varying capacity by ensembling the exits' predictions at different network depths.
For EE, the free parameter is the number and position of the exits within the network's depth $D$.

\textbf{Multi Input Multi Output (MIMO)} MIMO networks~\citep{li2018learning,havasi2020training,soflaei2020aggregated,murahari2022datamux,rame2021mixmo, sun2022towards, sunadapting} process multiple inputs in a single architecture while being competitive with rank-1 Bayesian NNs and Batch Ensemble approaches~\citep{wen2020batchensemble,dusenberry2020efficient,wenzel2020hyperparameter}.
A MIMO network processes $N$ inputs and outputs simultaneously, such that $N$ inputs are concatenated before the first layer into a single feature tensor.
At the end of the network at depth $D$, the network gives $N$ predictions, a separate prediction for each input.
During training all the $N$ input and target pairs are distinct, and at test time, the predictions for $N$ repeated identical inputs are aggregated to give a single prediction.
MIMO learns a representation that encodes the information from all inputs simultaneously, utilising the network's capacity more efficiently.
For MIMO, the free parameter is the number of inputs $N$ that the network should process simultaneously.

\textbf{Multi Input Massive Multi Output (MIMMO)}
MIMMO networks~\citep{ferianc2023mimmo} combine MIMO and EE by adding exactly $D-1$ reshaping connectors to the network, feeding in the intermediate layers' outputs to the prediction head at the final layer.
MIMMO processes $N$ inputs and outputs for each exit, resulting in $N \times D$ predictions per forward pass through the network.
MIMMO trains to match the predictions between all exits and targets, and at test time, the predictions are averaged across all exits for repeated inputs.
While MIMMO utilises all the exits' predictions within the network's depth $D$, the number of inputs $N$ is a free parameter.

The current methodologies in hardware-efficient NN ensembles, such as EE, MIMO, and MIMMO, are specialised and constrained to particular settings. 
For instance, EE focuses on the arrangement and number of exits within a network's depth, MIMO on processing multiple inputs simultaneously, and MIMMO combines features of both, optimising the usage of all exits and the number of inputs.
Our work argues that there is not a universal, optimal configuration or method that suits all problem settings, but it depends on the type and size of the NN and the complexity of the dataset.
Therefore, it is necessary to define a scalable search space that encompasses these methods  and introduces their novel in-between configurations, and an optimisation objective that allows its efficient exploration.
In the Supplementary Material, we provide an even more detailed comparison of the generalised methods, and baseline methods compared to in the experiments.

\section{Single Architecture Ensemble (SAE)}\label{sec:sae}

To address the limitations of the current methodologies, we introduce the SAE framework.
It consists of a scalable search space and an optimisation objective that allows learning the optimal hardware-efficient ensemble configuration for a given task.

\subsection{Search Space of SAE}\label{sec:sae:search_space}

We first consider a naive generalisation of the EE, MIMO, and MIMMO methods and their novel in-between configurations by considering a network with $D$ layers and $N$ inputs and outputs per exit.
If we consider that a network processes $N$ inputs and can have up to $D$ exits, we can represent the search space of the generalised methods and their in-between configurations as ${(2^{D}-1)}^N$, where ${(2^{D}-1)}$ exits are turned on or off for each input in $N$.
Manual exploration of this search space is impractical, and it is necessary to introduce a simplification that allows automatic and efficient exploration of the search space.

\begin{figure}
    \centering
    
    \includegraphics[width=0.5\linewidth]{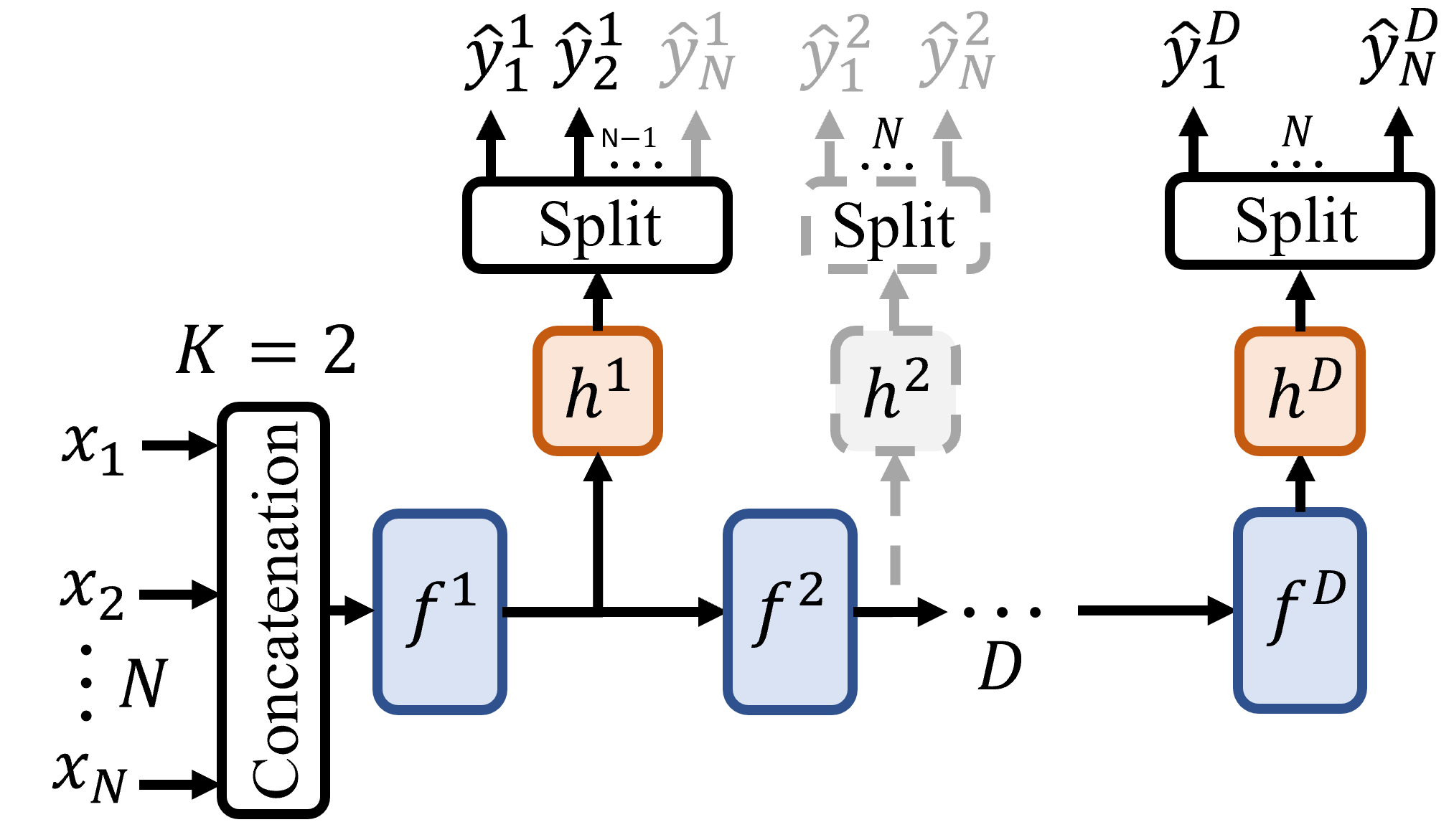}
    \vspace{0.1cm}
    \caption{The Single Architecture Ensemble (SAE). The filled rectangles stand for learnable layers $\{f^j(\cdot)\}^D_{j=1}$ and prediction heads $\{h^j(\cdot)\}^D_{j=1}$, while the empty rectangle represents a non-parametric operation. 
    The $D$ is the network depth, $N$ is the number of separate inputs in the ensemble, and $K$ is the maximum number of active exits during evaluation for each input.
    The $\{x_i\}_{i=1}^N$ are the $N$ inputs and $\{\hat{y}_i^j\}^{N,D}_{i,j=1}$ represent the predictions from the $D$ exits for $N$ inputs. 
    The arrows represent the flow of information.
    The dashed and greyed boxes and arrows represent exits that were active during training but inactive during evaluation because of top $K$ exits identified during training.}
    \label{fig:sae}
\end{figure}

We simplify the search space by considering the number of inputs $N$ and the \textit{maximum} number of exits $1 \leq K \leq D$ that can be active for each input in $N$.
The new efficient search space is reduced to $N \times D \ll {(2^{D}-1)}^N$ where a practitioner needs to try $N\times D$ configurations via setting $N$ and $K$.
By setting $N$ and $K$, we recover the generalised methods as:
\begin{multicols}{2}
\begin{itemize}[topsep=-0.12cm, itemsep=-0.1cm]
    \item if $N=1$, $K=1$, this is a Standard Single exit (SE) NN;
    \item if $N=1$, $K\geq 2$, this is an EE NN;
    \item if $N \geq 2$, $K=1$, this is a MIMO NN;
    \item if $N \geq 2$, $K=D$, this is a MIMMO NN;
    \item if $N \geq 2$, $1 < K < D$, this is a previously unexplored In-Between (I/B) configuration.
\end{itemize}
\end{multicols}
The SAE search space can be illustrated as a network architecture in Figure~\ref{fig:sae}.
The network processes $N$ inputs via a widened input layer, after which the refined features of the $D$ layers are processed by exits at different depths.
During training, all the exits are active, and the network produces $N \times D$ predictions.
However, during evaluation, the network produces $N \times K$ predictions by selecting the top $K$ exits for each input.
Next, we present the optimisation objective that allows learning which $K$ exits to use for each input in $N$ and the network weights at the same time.

\subsection{Optimisation Objective of SAE}\label{sec:sae:optimisation}

\revision{The optimisation is built on jointly learning the network weights and the depth distribution of the $K$ most suitable exits for each input in $N$ via variational inference~\citep{blei2017variational}.
Through defining the preferences for the exits for each input as learnable variables, we avoid having to manually search through the search space of $(2^{D}-1)^N$ configurations, and instead the depth preference is learned during training, similarly to learning operations in differentiable neural architecture search (DNAS)~\citep{liu2018darts}.
In contrast to DNAS, the final configuration does not need to be discretised and retrained.
To fully explore the search space, the user needs to run the training only $N \times D$ times, where up to $D$ exits can be active for each input, making the traversal of the search space more efficient.}

We define the SAE's optimisation objective, initially considering $N$ independent ensemble inputs.
Denote the dataset $\mathcal{D} = \{x_i,y_i\}_{i=1}^{|\mathcal{D}|}$, where $x_i$ is the input and $y_i$ is the ground truth for input $i$ and $|\mathcal{D}|$ is the size of the dataset $\mathcal{D}$.
We assume initially that our model processes $N$ inputs $\{x_i\}^{N}_{i=1}$ and produces $N$ matching outputs $\{\hat{y}_i\}^{N}_{i=1}$.
During training, we assume that the $N$ inputs $\{x_i\}^{N}_{i=1}$ and targets $\{y_i\}^{N}_{i=1}$ are sampled independently from the dataset $\mathcal{D}$ in $N$, hence assuming independence between $p(\{y_i\}^{N}_{i=1}|\{x_i\}^{N}_{i=1}, \mathbf{w}) = \prod_{i=1}^N p(y_i|\{x_z\}^{N}_{z=1}, \mathbf{w})$~\citep{havasi2020training}, where $\mathbf{w}$ are the learnable weights. 
By independence, we mean that the pairs $(x_i, y_i)$ are sampled independently from the other $N-1$ pairs. 
There is no mixing of inputs and targets and $y_i$ is the ground truth associated with $x_i$.
We denote $\mathbf{x}_i=\{x_z\}^{N}_{z=1}$ as the input for $i$, containing $x_i$ and $N-1$ random sampled inputs from the dataset $\mathcal{D}$, matching the $\{y_i\}^{N-1}_{i=1}$.

With the independence assumption, we introduce the exits, which enable the NN to make a prediction for each input $i$ at $D$ different depths as $p(y_i|\mathbf{x}_i, d_i=j, \mathbf{w}) =p(y_i|\hat{y}_i^j); \hat{y}_i^j = h^j(f^j(f^{j-1}(\dots f^1(\mathbf{x}_i))))$ where $1 \leq j \leq D$ and $f^j(\cdot)$ is the $j$-th layer and $h^j(\cdot)$ is the prediction head for the $j$-th layer.
The $\hat{y}_i^j$ is the prediction for input $i$ at depth $j$ and $d_i$ is the latent variable for $D$ exits for input $i$.
The computation is visualised in Figure~\ref{fig:sae}.

The evidence of the data for a single tuple is $p(y_i|\mathbf{x}_i, \mathbf{w}) = \sum_{j=1}^D p(y_i|\mathbf{x}_i, d_i=j, \mathbf{w}) p(d_i=j)$ for a categorical prior for the latent variable $d_i$, $p(d_i)$ where $p(d)=\prod_{i=1}^N p(d_i)$.
We introduce a categorical variational distribution $q(d|\mathbf{\theta})=\prod_{i=1}^N q(d_i|\mathbf{\theta}_i)$ with parameters $\mathbf{\theta}=\{\theta_i^j\}^{N,D}_{i,j=1}$, where $\theta_i$ are the parameters of the weightings of all the exits for input $i$.
Our aim is to learn $\mathbf{\theta}$ to determine the $K$ active exits for each input during evaluation together with $\mathbf{w}$.

~\citet{antoran2020depth} demonstrated for $N=1$ that by choosing the prior and the variational distribution to be categorical, it is possible to jointly optimise $\mathbf{w}$ and $\mathbf{\theta}$ and maximise the evidence lower bound of the data via stochastic gradient ascent.
To derive the evidence lower bound for all $N$ independent predictors, we minimise the Kulback-Leibler (KL) divergence~\citep{csiszar1975divergence} between the variational distribution $q(d|\mathbf{\theta})$ and the posterior distribution $p(d|\mathcal{D}, \mathbf{w}) = \prod_{i=1}^N p(d_i|\mathcal{D}, \mathbf{w})$ as:
\begin{align}
     &KL(q(d|\mathbf{\theta})\,||\,p(d|\mathcal{D}, \mathbf{w})) = \sum_{i=1}^N KL(q(d_i|\mathbf{\theta}_i)\,||\,p(d_i|\mathcal{D}, \mathbf{w})) \nonumber \\
     =&\sum_{i=1}^N \mathbb{E}_{q(d_i|\mathbf{\theta}_i)} \left[ \log q(d_i|\mathbf{\theta}_i) - \log \frac{p(Y_i|\mathbf{X}_i, d_i, \mathbf{w}) p(d_i)}{p(Y_i|\mathbf{X}_i, \mathbf{w})} \right] \nonumber \\
    &= \underbrace{\sum_{i=1}^N \mathbb{E}_{q(d_i|\mathbf{\theta}_i)} \left[ - \log p(Y_i|\mathbf{X}_i, d_i, \mathbf{w}) \right]}_{\text{Data-fit in } -\mathcal{L}(\mathbf{w}, \mathbf{\theta})} +  \underbrace{KL(q(d_i|\mathbf{\theta}_i)\,||\,p(d_i))}_{\text{Regulariser in } -\mathcal{L}(\mathbf{w}, \mathbf{\theta})} + \underbrace{\sum_{i=1}^N \log p(Y_i|\mathbf{X}_i, \mathbf{w})}_{\text{Evidence}} \label{eq:elbo_1} 
\end{align}
The KL divergence in Equation~\ref{eq:elbo_1} is decomposed into the expectation of the log-likelihood, or the data-fit term, of the data $Y_i=\{y_i^{(j)}\}^{|D|}_{j=1}$, given input tuples $\mathbf{X}_i=\{\mathbf{x}_i^{(j)}\}^{|D|}_{j=1}$ for dataset size $|\mathcal{D}|$, the exits $d_i$ and the weights $\mathbf{w}$, the KL divergence between the variational distribution $q(d_i|\mathbf{\theta}_i)$ and the prior distribution $p(d_i)$, or the regulariser term, and the evidence of the data.
The KL divergence is non-negative, and the evidence of the data is maximised when the $\mathcal{L}(\mathbf{w}, \mathbf{\theta})$ is maximised as $\mathcal{L}(\mathbf{w}, \mathbf{\theta}) \leq \sum_{i=1}^N \log p(Y_i|\mathbf{X}_i, \mathbf{w})$.
The objective $\mathcal{L}(\mathbf{w}, \mathbf{\theta})$ can be further decomposed with respect to all the inputs $N$ and the whole dataset $\mathcal{D}$ as in Equation~\ref{eq:elbo_2}.
\begin{align}
    \mathcal{L}(\mathbf{w}, \mathbf{\theta}) &= \sum_{i=1}^{N} \mathbb{E}_{q(d_i|\mathbf{\theta}_i)} \left[ \log p(Y_i|\mathbf{X}_i, d_i, \mathbf{w}) \right] - \sum_{i,j=1}^{N, D} \theta_i^j \log \frac{\theta_i^j}{D^{-1}}  \notag \\
     &= \sum_{i,b=1}^{N, |\mathcal{D}|} \mathbb{E}_{q(d_i|\mathbf{\theta}_i)} \left[ \log p(y_i^{(b)}|\mathbf{x}_i^{(b)}, d_i, \mathbf{w}) \right] - \sum_{i,j=1}^{N, D} \theta_i^j \log \frac{\theta_i^j}{D^{-1}}  \label{eq:elbo_2}
\end{align}
The $b$ is the index of the input tuple $\mathbf{x}_i^{(b)}$ and the target $y_i^{(b)}$ sampled from the dataset $\mathcal{D}$.

We assume a uniform categorical prior, meaning that the KL divergence between the variational distribution $q(d|\mathbf{\theta})$ and the prior distribution $p(d)$ can be computed in closed form.
Equation~\ref{eq:elbo_3} approximates the evidence lower bound, where $B$ is the batch size.
\begin{align}
    \mathcal{L}(\mathbf{w}, \mathbf{\theta}) &\approx \frac{|\mathcal{D}|}{B} \sum_{i,b=1}^{N, B} \mathbb{E}_{q(d_i|\mathbf{\theta}_i)} \left[ \log p(y_i^{(b)}|\mathbf{x}_i^{(b)}, d_i, \mathbf{w}) \right] - \sum_{i,j=1}^{N,D} \theta_i^j \log \frac{\theta_i^j}{D^{-1}} \label{eq:elbo_3}
\end{align}

We propose to sample the depth variable $d_i$ from the variational distribution $q(d_i|\mathbf{\theta}_i)$ using a top $K$ sampling strategy~\citep{kool2019stochastic} which enables us to optimise the $\mathbf{\theta}$ towards the top $K$ exits per input.
We use one Monte Carlo sample per batch to approximate the expectation, the sum of the $D$ log-likelihoods multiplied by the sampled probabilities of $d_i$.
The $\mathbf{\theta}$ are implemented as zero-initialised logits divided by a temperature $T$ to which softmax was applied to normalise the logits as $\theta_i^j = \frac{\exp(l_i^j/T)}{\sum_{k=1}^D \exp(l_i^k/T)}$ and $l_i^j$ is the $j$-th exit logit for input $i$.

We empirically observed that the training of the SAE might be compromised by overregularisation, which can be caused by the regulariser term in Equation~\ref{eq:elbo_3} or by the size of the dataset $|\mathcal{D}|$.
We propose to multiply the regulariser term by a factor $\alpha(t)$, where $0 \leq \alpha(t) \leq 1$ is a linear interpolation depending on the training step $t$, and replace the $\frac{|\mathcal{D}|}{B}$ with $1$ in Equation~\ref{eq:elbo_3}.
Furthermore, we propose to linearly interpolate between starting and ending values of the temperature $0 < T(t)\leq1$ and the input repetition factor $0 \leq i(t)\leq1$ during training, where the starting and ending values can be optimised.
The $T(t)$ determines the sharpness of the probabilities over the auxiliary exits during sampling, which enables the framework to gradually focus on the $K$ most important auxiliary exits for each input in the architecture.
The $i(t)$ determines a portion of the batch $B$ where the same input is repeated across all the $N$ inputs, relaxing the independence assumption~\citep{havasi2020training}.
The linear interpolation of hyperparameters aims to enable the framework to iterate over multiple configurations during training, trading-off exploration and exploitation of the search space.
\revision{We empirically observed that the linear hyperparameter schedules were not too disruptive to the training process, however, other strategies can be considered.}

During the evaluation, the SAE framework produces $N \times K$ predictions $\{\hat{y}_i^j\}^{N,K}_{i,j=1}$, where $K$ is the number of exits used for each input in $N$.
It does so through only keeping the top $K$ largest logits and setting the rest to negative infinity to give $\mathbf{\theta}^* = \{\theta_i^{*j}\}^{N,K}_{i,j=1}$.
If no inputs select an exit at some level $j$, it does not need to be implemented as shown in Figure~\ref{fig:sae} in grey.
The input sample is repeated $N$ times as $\mathbf{x}^{*}=\{x_{i}\}^{N}_{i=1}$ and the predictions from the active exits are collected. 
The final prediction is obtained by averaging the predictions from the exits and their $\theta_i^{*j}$ as $\hat{y}^* = \frac{1}{N} \sum_{i,j=1}^{N,K} \hat{y}_i^j \theta_i^{*j}$ which approximates the marginal likelihood. 

In summary, the training objective $\mathcal{L}(\mathbf{w}, \mathbf{\theta})$ allows us to learn the exit preferences $\mathbf{\theta}$ for each input $i\in N$ and each exit $j\in D$ along the network's weights $\mathbf{w}$ via gradient ascent.
A practitioner can set the $N\geq1$ and $1\leq K \leq D$ to explore the search space of the generalised methods and their novel in-between configurations by optimising the $\mathbf{w}$ and $\mathbf{\theta}$ via the $\mathcal{L}(\mathbf{w}, \mathbf{\theta})$.
The resultant $\mathbf{w}$ and $\mathbf{\theta}^*$ define the optimal configuration for a given task, maximally exploiting the network's capacity through the search space exploration.
In the Supplementary Material, we provide an Algorithm summarising the training and evaluation as well as the implementation details for the input and early exits per architecture type.

\begin{figure*}[t]
    \centering
    
    \begin{subfigure}[t]{0.22\textwidth}
        \centering
        \includegraphics[width=\textwidth]{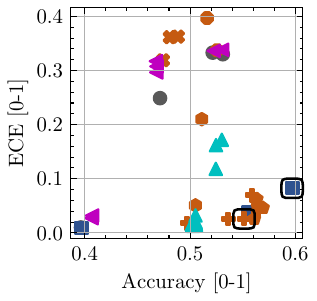}
        \caption{TinyImageNet ACC, ECE.}
        \label{fig:baselines:tinyimagenet:id:accuracy_calibration}
    \end{subfigure}
    \hfill
    \begin{subfigure}[t]{0.24\textwidth}
        \centering
        \includegraphics[width=\textwidth]{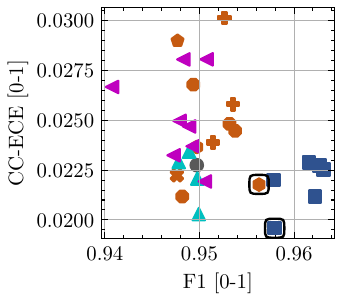}
        \caption{BloodMNIST F1, CC-ECE.}
        \label{fig:baselines:bloodmnist:id:f1_ccece}
    \end{subfigure}
    \hfill
    \begin{subfigure}[t]{0.235\textwidth}
        \centering
        \includegraphics[width=\textwidth]{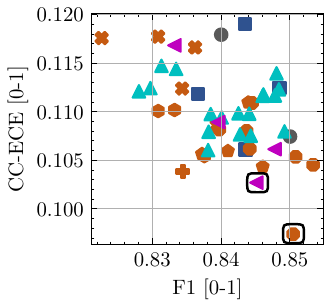}
        \caption{PneumoniaMNIST F1, CC-ECE.}
        \label{fig:baselines:pneumoniamnist:id:f1_ccece}
    \end{subfigure}
    \hfill
    \begin{subfigure}[t]{0.235\textwidth}
        \centering
        \includegraphics[width=\textwidth]{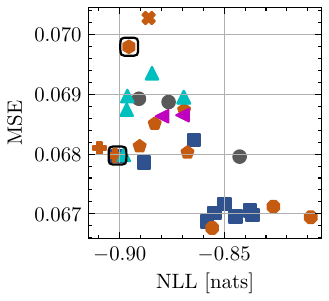}
        \caption{RetinaMNIST NLL and MSE.}
        \label{fig:baselines:retinamnist:id:nll_mse}
    \end{subfigure}
    
    \begin{subfigure}[t]{0.21\textwidth}
        \centering
        \includegraphics[width=\textwidth]{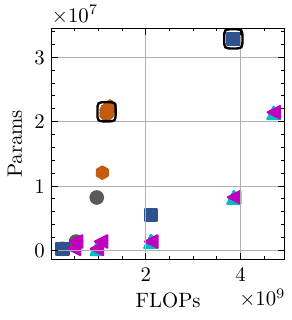}
        \caption{TinyImageNet FLOPs, Params.}
        \label{fig:baselines:tinyimagenet:id:flops_params}
    \end{subfigure}
    \hfill
    \begin{subfigure}[t]{0.22\textwidth}
        \centering
        \includegraphics[width=\textwidth]{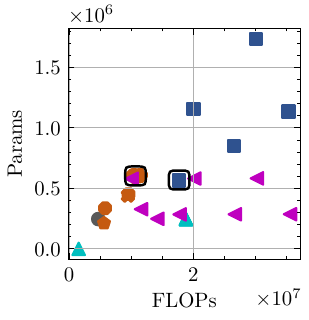}
        \caption{BloodMNIST FLOPs, Params.}
        \label{fig:baselines:bloodmnist:id:flops_params}
    \end{subfigure}
    \hfill
    \begin{subfigure}[t]{0.21\textwidth}
        \centering
        \includegraphics[width=\textwidth]{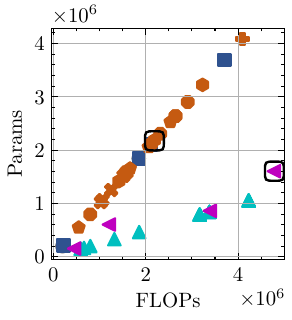}
        \caption{PneumoniaMNIST FLOPs, Params.}
        \label{fig:baselines:pneumoniamnist:id:flops_params}
    \end{subfigure}
    \hfill
    \begin{subfigure}[t]{0.21\textwidth}
        \centering
        \includegraphics[width=\textwidth]{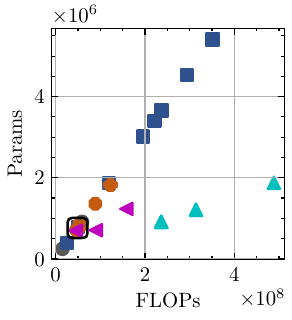}
        \caption{RetinaMNIST FLOPs, Params.}
        \label{fig:baselines:retinamnist:id:flops_params}
    \end{subfigure}
    \vspace{0.2cm}
    \caption{Comparison on ID test sets, with respect to \textbf{\textcolor{Gray}{Standard NN $\newmoon$}}, \textbf{\textcolor{Blue}{NN Ensemble $\blacksquare$}}, 
    \textbf{\textcolor{Orange}{SAE: 
    I/B: $N\geq 2, 2 \leq K < D$ \PlusMarker, 
    EE: $N=1, K \geq 2$ \OctagonMarker, 
    MIMMO: $N\geq 2, K=D$ \PentagonMarker, 
    MIMO: $N\geq 2, K=1$ \HexagonMarker, 
    SE NN: $N=1, K=1$ \TimesMarker}}, 
    \textbf{\textcolor{Cyan}{MCD $\blacktriangle$}}, 
    \textbf{\textcolor{Pink}{BE $\blacktriangleleft$}}.
    The black outlines denote the configurations compared in the text.}
    \label{fig:baseline:id_plots}
\end{figure*}

\section{Experiments}\label{sec:experiments}

We perform experiments on four datasets: TinyImageNet~\citep{le2015tiny}, BloodMNIST, PneumoniaMNIST and RetinaMNIST,~\citep{yang2023medmnist} for both classification and regression tasks.
Our architecture backbones are ResNet~\cite{he2016deep} for TinyImageNet, ViT~\cite{dosovitskiy2020image} for RetinaMNIST, VGG~\cite{simonyan2014very} for BloodMNIST, and a residual fully connected net with batch normalisation~\cite{ioffe2015batch} and ReLU activations (FC) for PneumoniaMNIST.
We compare approaches under SAE with Monte Carlo Dropout (MCD)~\citep{gal2016dropout}, where we insert dropout~\cite{srivastava2014dropout} layers before each linear or convolutional layer.
We also compare to Batch Ensemble (BE)~\citep{wen2020batchensemble}, where the BE layers replace all linear and convolutional layers in the network.
The algorithmic lower bound is a standard NN.
The algorithmic upper bound is a naive ensemble of NNs.
We employ multi-objective Bayesian optimisation (MOBO) to perform hyperparameter optimisation (HPO) for SAE and: $1 \leq N \leq 4, 1 \leq K \leq D$ the input repetition, alpha and temperature start and end values and the width of the network for all the evaluated metrics simultaneously on the validation dataset.
For all the other methods, we enumerate networks of all widths, depths, or $N$ for BE and NN ensembles.
Exceptions are MCD, for which we set $N=4$ to make it competitive and we fix the  network width for TinyImageNet.
We use MOBO to search for the MCD or SAE's parameters.
We minimise unweighted negative log-likelihood (NLL) loss for all tasks except for SAE.
For classification, we measure F1 score $[\uparrow 0,1]$, Accuracy (ACC) $[\uparrow 0,1]$, Expected Calibration Error (ECE) $[\downarrow 0,1]$, Class Conditional ECE (CC-ECE) $[\downarrow 0,1]$ and the NLL $[\downarrow 0,\infty)$ as evaluation metrics.
For regression, we use Gaussian NLL $[\downarrow -\infty,\infty)$ and mean squared error (MSE) $[\downarrow 0,\infty)$ as evaluation metrics.
From the hardware perspective, we measure the number of floating point operations (FLOPs) $[\downarrow 0,\infty)$ and the number of parameters $[\downarrow 0,\infty)$.
We test on ID and OOD data created by applying augmentations to the test set, such as Gaussian noise, motion blur, and contrast changes~\citep{hendrycks2019benchmarking}.
\revision{We chose versatile architectures paired with datasets of varying complexity to demonstrate the SAE's effectiveness across different tasks and datasets.}
The Supplementary Material details architectures, datasets, metrics, HPO runs, additional results on the correlation of of hyperparameters, including $N$, $K$, $\alpha(t)$, $i(t)$ and $T(t)$ and the performance, and OOD experiments.

\subsection{Baseline Comparison}\label{sec:experiments:baseline}

\begin{figure*}[t]
    \centering
    \begin{subfigure}[t]{0.24\textwidth}
    \centering
    \includegraphics[width=\textwidth]{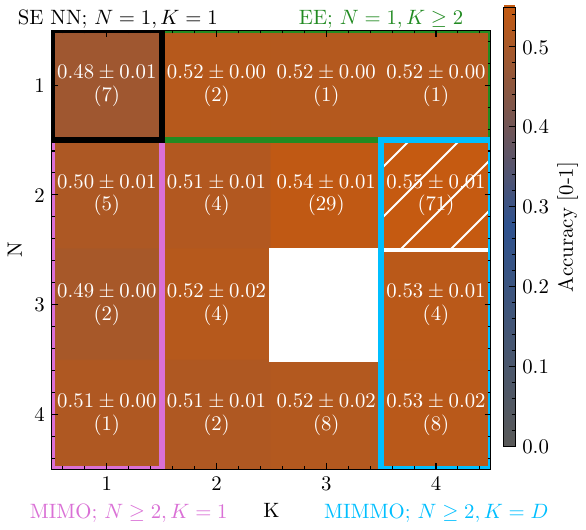}
    \vspace{-0.45cm}
    \caption{TinyImageNet ACC.}
    \label{fig:changing_k_n:tinyimagenet:accuracy:id}
    \end{subfigure}
    \hfill
    \begin{subfigure}[t]{0.24\textwidth}
    \centering
    \includegraphics[width=\textwidth]{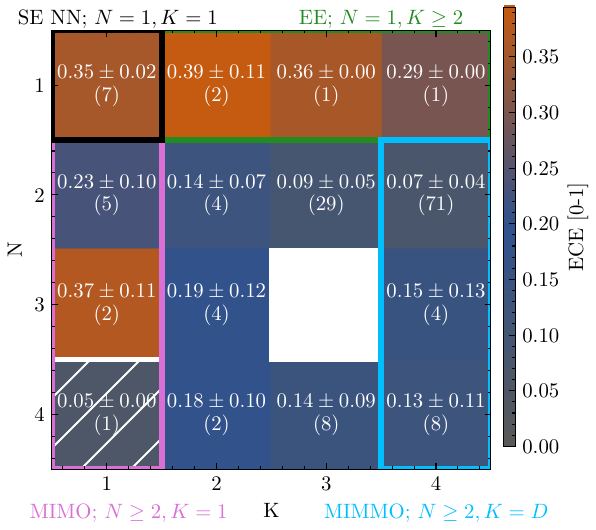}
    \vspace{-0.45cm}
    \caption{TinyImageNet ECE.}
    \label{fig:changing_k_n:tinyimagenet:ece:id}
    \end{subfigure}
    \hfill
    \begin{subfigure}[t]{0.24\textwidth}
    \centering
    \includegraphics[width=\textwidth]{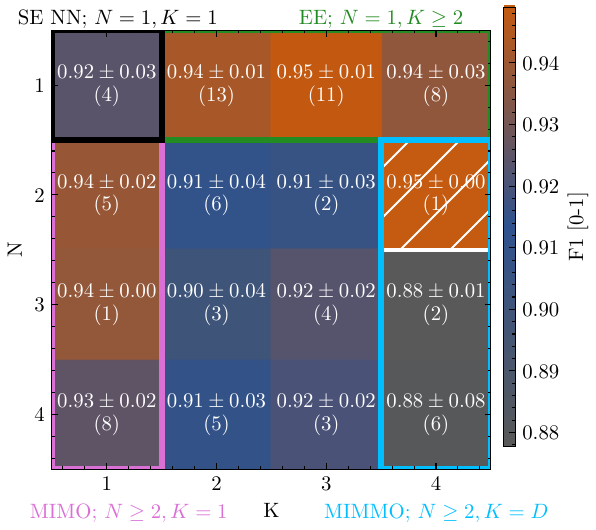}
    \vspace{-0.45cm}
    \caption{BloodMNIST F1.}
    \label{fig:changing_k_n:bloodmnist:f1:id}
    \end{subfigure}
    \hfill
    \begin{subfigure}[t]{0.24\textwidth}
    \centering
    \includegraphics[width=\textwidth]{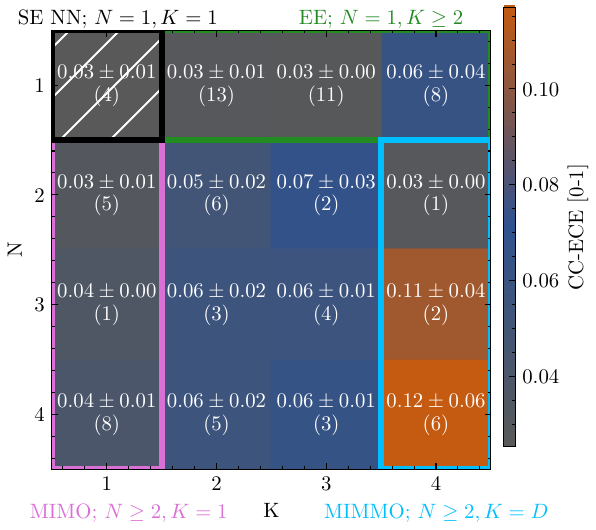}
    \vspace{-0.45cm}
    \caption{BloodMNIST CC-ECE.}
    \label{fig:changing_k_n:bloodmnist:ccece:id}
    \end{subfigure}
    \newline 
    \begin{subfigure}[t]{0.24\textwidth}
    \centering
    \includegraphics[width=\textwidth]{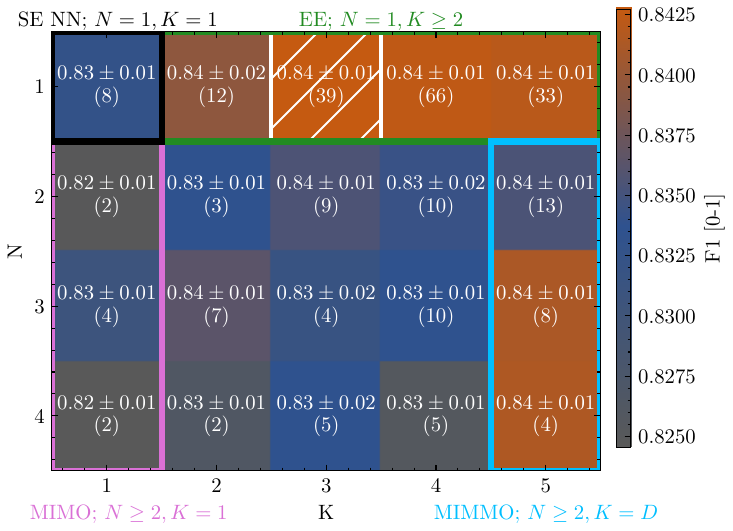}
    \vspace{-0.45cm}
    \caption{PneumoniaMNIST F1.}
    \label{fig:changing_k_n:pneumoniamnist:f1:id}
    \end{subfigure}
    \hfill
    \begin{subfigure}[t]{0.24\textwidth}
    \centering
    \includegraphics[width=\textwidth]{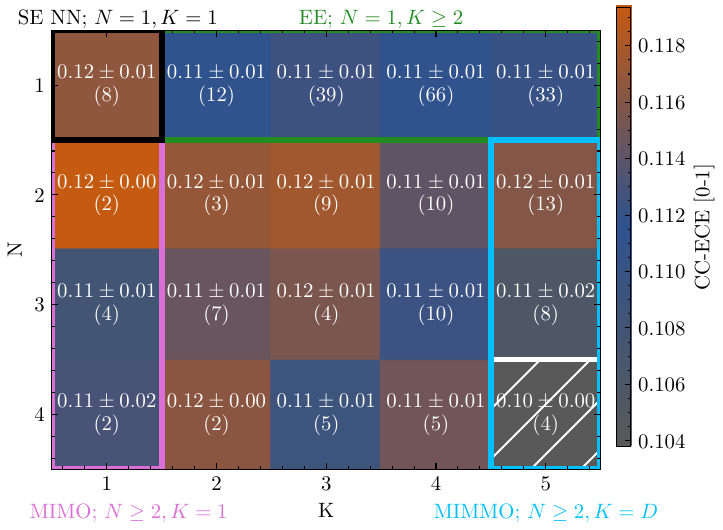}
    \vspace{-0.45cm}
    \caption{PneumoniaMNIST CC-ECE.}
    \label{fig:changing_k_n:pneumoniamnist:ccece:id}
    \end{subfigure}
    \hfill
    \begin{subfigure}[t]{0.24\textwidth}
    \centering
    \includegraphics[width=\textwidth]{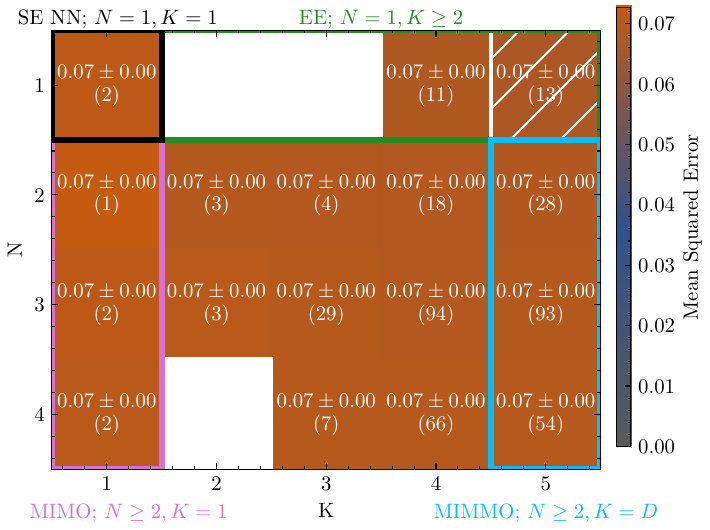}
    \vspace{-0.45cm}
    \caption{RetinaMNIST MSE.}
    \label{fig:changing_k_n:retinamnist:mse:id}
    \end{subfigure}
    \hfill
    \begin{subfigure}[t]{0.24\textwidth}
    \centering
    \includegraphics[width=\textwidth]{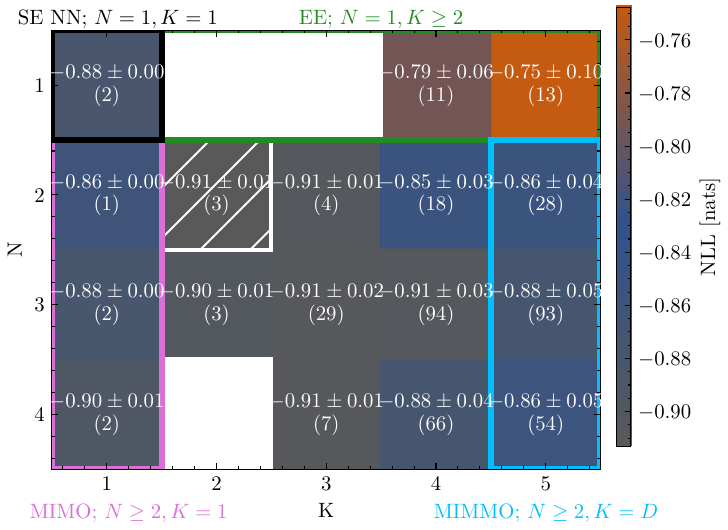}
    \vspace{-0.45cm}
    \caption{RetinaMNIST NLL.}
    \label{fig:changing_k_n:retinamnist:nll:id}
    \end{subfigure}
    \vspace{0.2cm}
    \caption{Varying $N,K$ on across ID test sets.
    The upper number is the average performance over $N$, $K$ combinations. 
    The number in brackets is the number of sampled configurations by HPO. 
    White box means no configurations sampled for that $N$, $K$. 
    Pattern signals best average performance.
    The coloured outlines signal the special cases for the generalised methods.}
    \label{fig:changing_k_n:id:accuracy_f1_mse_ece_ccece_nll}
    \vspace{-0.2cm}
\end{figure*}

In Figure~\ref{fig:baseline:id_plots}, we compare Pareto optimal configurations identified via HPO across all metrics and datasets.
In the upper row, we show the results on ACC/F1/MSE and ECE/CC-ECE/NLL and in the lower row, we show the results on FLOPs and parameters.
We split the methods in SAE into 5 categories based on the $N$ and $K$: {\textcolor{Orange}{I/B: $N\geq 2, 2 \leq K < D$ \PlusMarker}}, {\textcolor{Orange}{EE: $N=1, K \geq 2$ \OctagonMarker}}, {\textcolor{Orange}{MIMMO: $N\geq 2, K=D$ \PentagonMarker}}, {\textcolor{Orange}{MIMO: $N\geq 2, K=1$ \HexagonMarker}}, {\textcolor{Orange}{SE NN: $N=1, K=1$ \TimesMarker}}.
The black outlines around the markers in the Figures denote the configurations compared in the text.
The results are averaged across 4  random seeds.
The Supplementary Material contains numerical results for OOD datasets, the dataset's best configurations and NLL comparison.

The overall results show that SAE's Pareto points cluster around the best algorithmic and hardware performance, while providing trade-offs between the two.
For TinyImageNet, in Figures~\ref{fig:baselines:tinyimagenet:id:accuracy_calibration} and~\ref{fig:baselines:tinyimagenet:id:flops_params}, it can be seen that the {\textcolor{Blue}{ensemble $\blacksquare$}} achieves the best accuracy.
However, SAE can find an \textcolor{Orange}{I/B \PlusMarker} configuration which is within 3\% of accuracy, but 4\% better ECE than the best \textcolor{Blue}{ensemble $\blacksquare$} with $3.2\times$ fewer FLOPs and $1.5\times$ fewer parameters.
For BloodMNIST, in Figures~\ref{fig:baselines:bloodmnist:id:f1_ccece} and~\ref{fig:baselines:bloodmnist:id:flops_params}, it can be seen that the {\textcolor{Blue}{ensemble $\blacksquare$}} achieves the best algorithmic performance. 
At the same time, the SAE can find an \textcolor{Orange}{I/B \PlusMarker} configuration which is within 1\% of the F1 score or CC-ECE with a comparable number of parameters but approximately $1.5\times$ fewer FLOPs.
For PneumoniaMNIST, in Figures~\ref{fig:baselines:pneumoniamnist:id:f1_ccece} and~\ref{fig:baselines:pneumoniamnist:id:flops_params}, it can be seen that the \textcolor{Orange}{EE \OctagonMarker} configuration is the best-performing method with approximately half FLOPs than the closest \textcolor{Pink}{BE $\blacktriangleleft$}.
For RetinaMNIST, in Figures~\ref{fig:baselines:retinamnist:id:nll_mse} and~\ref{fig:baselines:retinamnist:id:flops_params}, it can be seen that in terms of NLL the \textcolor{Orange}{MIMO \HexagonMarker} configuration is comparable to the \textcolor{Blue}{ensemble $\blacksquare$} with marginally fewer FLOPs and parameters. 

\begin{figure}
    \centering
    \begin{subfigure}[t]{0.24\textwidth}
    \centering
    \includegraphics[width=\textwidth]{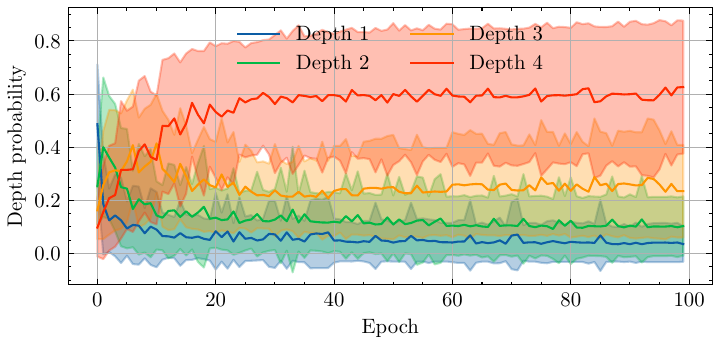}
    \vspace{-0.60cm}
    \caption{TinyImageNet.}
    \label{fig:depth:tinyimagenet:average}
    \end{subfigure}
    \hfill
    \begin{subfigure}[t]{0.24\textwidth}
    \centering
    \includegraphics[width=\textwidth]{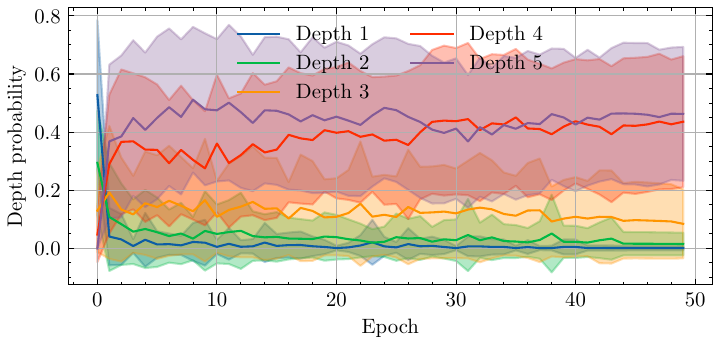}
    \vspace{-0.60cm}
    \caption{BloodMNIST.}
    \label{fig:depth:bloodmnist:average}
    \end{subfigure}
    \hfill
    \begin{subfigure}[t]{0.24\textwidth}
    \centering
    \includegraphics[width=\textwidth]{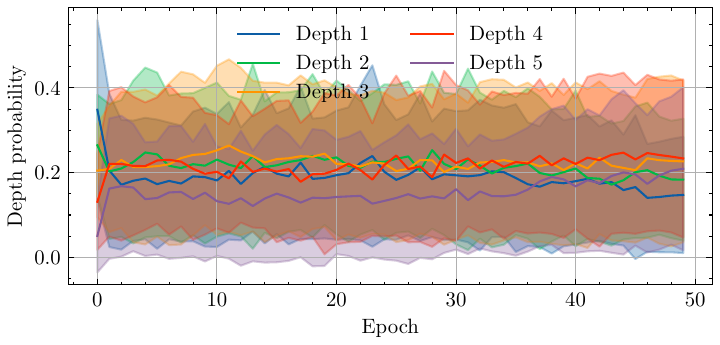}
    \vspace{-0.60cm}
    \caption{PneumoniaMNIST.}
    \label{fig:depth:pneumoniamnist:average}
    \end{subfigure}
    \hfill
    \begin{subfigure}[t]{0.24\textwidth}
    \centering
    \includegraphics[width=\textwidth]{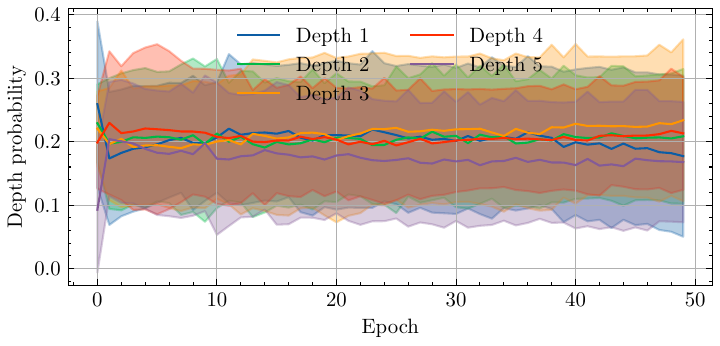}
    \vspace{-0.60cm}
    \caption{RetinaMNIST.}
    \label{fig:depth:retinamnist:average}
    \end{subfigure}
    \vspace{0.2cm}
    \caption{Depth preference during training when averaging all $N$ and $K$ for all datasets.
    The lines denote the mean trend, and the shaded regions denote the standard deviation across configurations.}
    \label{fig:depth:averages}
\end{figure}

In summary, SAE can find a set of competitive or better configurations than the baselines across all datasets, tasks, architectures with various capacities, and evaluation metrics using fewer or comparable FLOPs and parameters.
\revision{We hypothesise that this is due to the regularisation effect of the proposed objective, which prevents overfitting and instead focuses on fitting multiple sub-networks, each learning diverse representations, thus maximally utilising the capacity of the network.
Combination of the representations, reflected via the diverse predictions ultimately leads to better algorithmic performance.}
In the Supplementary Material, we rank the methods based on their mean performance across all their Pareto optimal points, datasets and metrics and show that SAE achieves best or competitive ranks across the algorithmic metrics.
SAE does not need custom operations as in BE, random number generators as in MCD, or more training rounds than a standard NN, MCD, or BE, it only needs to add compute-efficient early exits and an enlarged input layer capable of processing multiple inputs.
\revision{The Supplementary Material provides a detailed hardware cost breakdown of SAE per each architecture type.}

\subsection{Ablations}\label{sec:experiments:changing_k_n}
In Figure~\ref{fig:changing_k_n:id:accuracy_f1_mse_ece_ccece_nll}, we investigate the influence of $N$ and $K$ on the algorithmic performance across ID test sets across all datasets.
The Figures show the average performance across all $N$ and $K$ combinations across the individual HPO runs for each dataset.

As seen across the Figures, the best combinations of $N$ and $K$ can vary for the same dataset but different metrics, requiring a detailed search based on the specific use case.
Figures~\ref{fig:changing_k_n:tinyimagenet:accuracy:id} and~\ref{fig:changing_k_n:tinyimagenet:ece:id} show that for TinyImageNet, $N=2,4$ and $K=4,1$ are the best configurations for accuracy and ECE, respectively.
Figures~\ref{fig:changing_k_n:bloodmnist:f1:id} and~\ref{fig:changing_k_n:bloodmnist:ccece:id} show that for BloodMNIST, $N=2,1$ and $K=4,1$ are the best configurations for F1, CC-ECE, respectively.
Figures~\ref{fig:changing_k_n:pneumoniamnist:f1:id} and~\ref{fig:changing_k_n:pneumoniamnist:ccece:id} show that for PneumoniaMNIST, $N=1,4$ and $K=3,5$ are the best configurations for F1, CC-ECE, respectively.
Lastly, Figures~\ref{fig:changing_k_n:retinamnist:mse:id} and~\ref{fig:changing_k_n:retinamnist:nll:id} show that for RetinaMNIST, $N=1,2$ and $K=5,2$ are the best configurations for MSE and NLL, respectively.

An interesting observation is that $N>1$ leads to better algorithmic performance, which differs from~\citep{ferianc2023mimmo, havasi2020training} where $N>1$ consistently leads to worse performance.
On one hand, we observed that for easier problems, for example, PneumoniaMNIST the model's capacity is large in comparison to the complexity of the problem, hence the network was able to learn representations for multiple predictors in the same architecture resulting in $N>1$ and $K>1$ configurations being beneficial.
On the other hand, for harder problems, for example, TinyImageNet, the model's capacity was not sufficient to learn the representations for too many predictors, hence the performance peaks at a certain $N$ and $K$ combination.

Furthermore, we look at the learnt depth preference during training across all $N$ and $K$ for all datasets in Figure~\ref{fig:depth:averages}.
As discussed in the previous paragraph, if the model's capacity is large in comparison to the complexity of the problem, the depth preference is more uniform across the depths, as seen in Figures~\ref{fig:depth:bloodmnist:average},~\ref{fig:depth:pneumoniamnist:average}, and~\ref{fig:depth:retinamnist:average} for BloodMNIST, PneumoniaMNIST, and RetinaMNIST, respectively.
However, if the model's capacity is not sufficient to learn the representations for too many predictors, the depth preference is strongly biased towards deeper depths, as seen in Figure~\ref{fig:depth:tinyimagenet:average} for TinyImageNet.
These observations further validate our initial claim that there is no single best multi input multi output or early exit configuration across all datasets, tasks, and architectures, necessitating the need for a thorough and automatic search depending on the use case.

\revision{In terms of the backbone architectures and their impact on performance, we compare BloodMNIST and RetinaMNIST which have a similar complexity but use different architectures: VGG and ViT, respectively. 
We notice that for both problems the algorithmic performance improvements are similar across the datasets. 
However as discussed in the previous paragraphs, the optimal $N$ and $K$, along with other hyperparameters, as well as the depth preference differ between the two datasets.
Despite the different backbone architectures, the SAE can find competitive configurations for both datasets and architectures.}
The Supplementary Material contains more results on changing $N, K$ or and depth preference.

\section{Discussion}\label{sec:conclusion}

\revision{In this work we proposed the Single Architecture Ensemble (SAE). 
This unified framework melds the strengths of diverse hardware-efficient ensemble methods, outperforming traditional baselines in confidence calibration, accuracy and versatility across varied datasets, tasks, and architectures. 
Our findings underscore the necessity of SAE due to the lack of a universally superior method across all datasets, tasks, and architectures.
Our method demonstrated that different network depths were used in different architectures and problem complexities, necessitating the need to learn the appropriate depth usage.
Furthermore, we showed the different requirements for different architectures and the usefulness of adaptive hyperparameter scheduling.}

\revision{There are multiple limitations of our study. 
In order to gain full benefits from SAE, the early exits, which introduce branching in the architecture, need to be implemented efficiently.
An inefficient implementation can lead to a significant increase in the real-world hardware cost both during training and evaluation.
Furthermore, we assumed an independence assumption between the inputs, which might not hold in practice.
A theoretical investigation is needed to understand more complex assumptions and their impact on the performance and the trade-offs between the network's representation capacity and quality of the learned representations.
Lastly, we focused on image classification or regression tasks commonly solved in the community, which could fit our computational budget.}

\revision{Therefore, in future work, we plan to investigate the theoretical properties of the SAE, such as the impact of the independence assumption on the performance and the trade-offs between the network's representation capacity and the quality of the learned representations.
Furthermore, to minimise the number of sub-optimal configurations, we plan to investigate the automatic learning of $N$ and $K$, standing for the number of inputs and exits, respectively on tasks beyond image classification and regression.}
\newpage
\section*{Acknowledgements}

Martin Ferianc was sponsored through a scholarship from the Institute of Communications and Connected Systems at UCL. 
Martin wants to thank syne-tune and team Matthias Seeger for their work in maintaining and improving the repository and advice regarding HPO.
\bibliography{sample-bib}

\begin{thebibliography}{42}
\providecommand{\natexlab}[1]{#1}
\providecommand{\url}[1]{\texttt{#1}}
\expandafter\ifx\csname urlstyle\endcsname\relax
  \providecommand{\doi}[1]{doi: #1}\else
  \providecommand{\doi}{doi: \begingroup \urlstyle{rm}\Url}\fi

\bibitem[Allen-Zhu et~al.(2019)Allen-Zhu, Li, and Liang]{allen2019learning}
Zeyuan Allen-Zhu, Yuanzhi Li, and Yingyu Liang.
\newblock Learning and generalization in overparameterized neural networks, going beyond two layers.
\newblock \emph{Advances in neural information processing systems}, 32, 2019.

\bibitem[Antor{\'a}n et~al.(2020)Antor{\'a}n, Allingham, and Hern{\'a}ndez-Lobato]{antoran2020depth}
Javier Antor{\'a}n, James Allingham, and Jos{\'e}~Miguel Hern{\'a}ndez-Lobato.
\newblock Depth uncertainty in neural networks.
\newblock \emph{Advances in neural information processing systems}, 33:\penalty0 10620--10634, 2020.

\bibitem[Blei et~al.(2017)Blei, Kucukelbir, and McAuliffe]{blei2017variational}
David~M Blei, Alp Kucukelbir, and Jon~D McAuliffe.
\newblock Variational inference: A review for statisticians.
\newblock \emph{Journal of the American statistical Association}, 112\penalty0 (518):\penalty0 859--877, 2017.

\bibitem[Csisz{\'a}r(1975)]{csiszar1975divergence}
Imre Csisz{\'a}r.
\newblock I-divergence geometry of probability distributions and minimization problems.
\newblock \emph{The annals of probability}, pages 146--158, 1975.

\bibitem[Deng et~al.(2009)Deng, Dong, Socher, Li, Li, and Fei-Fei]{deng2009imagenet}
Jia Deng, Wei Dong, Richard Socher, Li-Jia Li, Kai Li, and Li~Fei-Fei.
\newblock Imagenet: A large-scale hierarchical image database.
\newblock In \emph{2009 IEEE conference on computer vision and pattern recognition}, pages 248--255. Ieee, 2009.

\bibitem[Dosovitskiy et~al.(2020)Dosovitskiy, Beyer, Kolesnikov, Weissenborn, Zhai, Unterthiner, Dehghani, Minderer, Heigold, Gelly, et~al.]{dosovitskiy2020image}
Alexey Dosovitskiy, Lucas Beyer, Alexander Kolesnikov, Dirk Weissenborn, Xiaohua Zhai, Thomas Unterthiner, Mostafa Dehghani, Matthias Minderer, Georg Heigold, Sylvain Gelly, et~al.
\newblock An image is worth 16x16 words: Transformers for image recognition at scale.
\newblock \emph{arXiv preprint arXiv:2010.11929}, 2020.

\bibitem[Dusenberry et~al.(2020)Dusenberry, Jerfel, Wen, Ma, Snoek, Heller, Lakshminarayanan, and Tran]{dusenberry2020efficient}
Michael~W Dusenberry, Ghassen Jerfel, Yeming Wen, Yi-an Ma, Jasper Snoek, Katherine Heller, Balaji Lakshminarayanan, and Dustin Tran.
\newblock Efficient and scalable {B}ayesian neural nets with rank-1 factors.
\newblock In \emph{ICML}, 2020.

\bibitem[Ferianc and Rodrigues(2023)]{ferianc2023mimmo}
Martin Ferianc and Miguel Rodrigues.
\newblock Mimmo: Multi-input massive multi-output neural network.
\newblock In \emph{Proceedings of the IEEE/CVF Conference on Computer Vision and Pattern Recognition}, pages 4563--4568, 2023.

\bibitem[Ferianc and Rodrigues(2024)]{ferianc2024yamle}
Martin Ferianc and Miguel Rodrigues.
\newblock Yamle: Yet another machine learning environment.
\newblock \emph{arXiv preprint arXiv:2402.06268}, 2024.

\bibitem[Fort et~al.(2019)Fort, Hu, and Lakshminarayanan]{fort2019deep}
Stanislav Fort, Huiyi Hu, and Balaji Lakshminarayanan.
\newblock Deep ensembles: A loss landscape perspective.
\newblock \emph{arXiv preprint arXiv:1912.02757}, 2019.

\bibitem[Gal and Ghahramani(2016)]{gal2016dropout}
Yarin Gal and Zoubin Ghahramani.
\newblock Dropout as a bayesian approximation: Representing model uncertainty in deep learning.
\newblock In \emph{international conference on machine learning}, pages 1050--1059. PMLR, 2016.

\bibitem[Girden(1992)]{girden1992anova}
Ellen~R Girden.
\newblock \emph{ANOVA: Repeated measures}.
\newblock Number~84. sage, 1992.

\bibitem[Guo et~al.(2017)Guo, Pleiss, Sun, and Weinberger]{guo2017calibration}
Chuan Guo, Geoff Pleiss, Yu~Sun, and Kilian~Q Weinberger.
\newblock On calibration of modern neural networks.
\newblock In \emph{International conference on machine learning}, pages 1321--1330. PMLR, 2017.

\bibitem[Havasi et~al.(2020)Havasi, Jenatton, Fort, Liu, Snoek, Lakshminarayanan, Dai, and Tran]{havasi2020training}
Marton Havasi, Rodolphe Jenatton, Stanislav Fort, Jeremiah~Zhe Liu, Jasper Snoek, Balaji Lakshminarayanan, Andrew~M Dai, and Dustin Tran.
\newblock Training independent subnetworks for robust prediction.
\newblock \emph{arXiv preprint arXiv:2010.06610}, 2020.

\bibitem[He et~al.(2016)He, Zhang, Ren, and Sun]{he2016deep}
Kaiming He, Xiangyu Zhang, Shaoqing Ren, and Jian Sun.
\newblock Deep residual learning for image recognition.
\newblock In \emph{Proceedings of the IEEE conference on computer vision and pattern recognition}, pages 770--778, 2016.

\bibitem[Hendrycks and Dietterich(2019)]{hendrycks2019benchmarking}
Dan Hendrycks and Thomas Dietterich.
\newblock Benchmarking neural network robustness to common corruptions and perturbations.
\newblock \emph{arXiv preprint arXiv:1903.12261}, 2019.

\bibitem[Ioffe and Szegedy(2015)]{ioffe2015batch}
Sergey Ioffe and Christian Szegedy.
\newblock Batch normalization: Accelerating deep network training by reducing internal covariate shift.
\newblock In \emph{International conference on machine learning}, pages 448--456. pmlr, 2015.

\bibitem[Kool et~al.(2019)Kool, Van~Hoof, and Welling]{kool2019stochastic}
Wouter Kool, Herke Van~Hoof, and Max Welling.
\newblock Stochastic beams and where to find them: The gumbel-top-k trick for sampling sequences without replacement.
\newblock In \emph{International Conference on Machine Learning}, pages 3499--3508. PMLR, 2019.

\bibitem[Lakshminarayanan et~al.(2017)Lakshminarayanan, Pritzel, and Blundell]{lakshminarayanan2017simple}
Balaji Lakshminarayanan, Alexander Pritzel, and Charles Blundell.
\newblock Simple and scalable predictive uncertainty estimation using deep ensembles.
\newblock \emph{Advances in neural information processing systems}, 30, 2017.

\bibitem[Laskaridis et~al.(2021)Laskaridis, Kouris, and Lane]{laskaridis2021adaptive}
Stefanos Laskaridis, Alexandros Kouris, and Nicholas~D Lane.
\newblock Adaptive inference through early-exit networks: Design, challenges and directions.
\newblock In \emph{Proceedings of the 5th International Workshop on Embedded and Mobile Deep Learning}, pages 1--6, 2021.

\bibitem[Le and Yang(2015)]{le2015tiny}
Ya~Le and Xuan Yang.
\newblock Tiny imagenet visual recognition challenge.
\newblock \emph{CS 231N}, 7\penalty0 (7):\penalty0 3, 2015.

\bibitem[Li and Liang(2018)]{li2018learning}
Yuanzhi Li and Yingyu Liang.
\newblock Learning overparameterized neural networks via stochastic gradient descent on structured data.
\newblock \emph{Advances in neural information processing systems}, 31, 2018.

\bibitem[Liu et~al.(2018)Liu, Simonyan, and Yang]{liu2018darts}
Hanxiao Liu, Karen Simonyan, and Yiming Yang.
\newblock Darts: Differentiable architecture search.
\newblock \emph{arXiv preprint arXiv:1806.09055}, 2018.

\bibitem[Loshchilov and Hutter(2016)]{loshchilov2016sgdr}
Ilya Loshchilov and Frank Hutter.
\newblock Sgdr: Stochastic gradient descent with warm restarts.
\newblock \emph{arXiv preprint arXiv:1608.03983}, 2016.

\bibitem[Matsubara et~al.(2022)Matsubara, Levorato, and Restuccia]{matsubara2022split}
Yoshitomo Matsubara, Marco Levorato, and Francesco Restuccia.
\newblock Split computing and early exiting for deep learning applications: Survey and research challenges.
\newblock \emph{ACM Computing Surveys}, 55\penalty0 (5):\penalty0 1--30, 2022.

\bibitem[Murahari et~al.(2022)Murahari, Jimenez, Yang, and Narasimhan]{murahari2022datamux}
Vishvak Murahari, Carlos~E Jimenez, Runzhe Yang, and Karthik Narasimhan.
\newblock Datamux: Data multiplexing for neural networks.
\newblock \emph{arXiv preprint arXiv:2202.09318}, 2022.

\bibitem[Ovadia et~al.(2019)Ovadia, Fertig, Ren, Nado, Sculley, Nowozin, Dillon, Lakshminarayanan, and Snoek]{ovadia2019can}
Yaniv Ovadia, Emily Fertig, Jie Ren, Zachary Nado, David Sculley, Sebastian Nowozin, Joshua Dillon, Balaji Lakshminarayanan, and Jasper Snoek.
\newblock Can you trust your model's uncertainty? evaluating predictive uncertainty under dataset shift.
\newblock \emph{Advances in neural information processing systems}, 32, 2019.

\bibitem[Paria et~al.(2020)Paria, Kandasamy, and P{\'o}czos]{paria2020flexible}
Biswajit Paria, Kirthevasan Kandasamy, and Barnab{\'a}s P{\'o}czos.
\newblock A flexible framework for multi-objective bayesian optimization using random scalarizations.
\newblock In \emph{Uncertainty in Artificial Intelligence}, pages 766--776. PMLR, 2020.

\bibitem[Qendro et~al.(2021)Qendro, Campbell, Lio, and Mascolo]{qendro2021early}
Lorena Qendro, Alexander Campbell, Pietro Lio, and Cecilia Mascolo.
\newblock Early exit ensembles for uncertainty quantification.
\newblock In \emph{Machine Learning for Health}, pages 181--195. PMLR, 2021.

\bibitem[Ram{\'e} et~al.(2021)Ram{\'e}, Sun, and Cord]{rame2021mixmo}
Alexandre Ram{\'e}, R{\'e}my Sun, and Matthieu Cord.
\newblock Mixmo: Mixing multiple inputs for multiple outputs via deep subnetworks.
\newblock In \emph{Proceedings of the IEEE/CVF International Conference on Computer Vision}, pages 823--833, 2021.

\bibitem[Salinas et~al.(2022)Salinas, Seeger, Klein, Perrone, Wistuba, and Archambeau]{salinas2022syne}
David Salinas, Matthias Seeger, Aaron Klein, Valerio Perrone, Martin Wistuba, and Cedric Archambeau.
\newblock Syne tune: A library for large scale hyperparameter tuning and reproducible research.
\newblock In \emph{International Conference on Automated Machine Learning, AutoML 2022}, 2022.
\newblock URL \url{https://proceedings.mlr.press/v188/salinas22a.html}.

\bibitem[Simonyan and Zisserman(2014)]{simonyan2014very}
Karen Simonyan and Andrew Zisserman.
\newblock Very deep convolutional networks for large-scale image recognition.
\newblock \emph{arXiv preprint arXiv:1409.1556}, 2014.

\bibitem[Soflaei et~al.(2020)Soflaei, Guo, Al-Bashabsheh, Mao, and Zhang]{soflaei2020aggregated}
Masoumeh Soflaei, Hongyu Guo, Ali Al-Bashabsheh, Yongyi Mao, and Richong Zhang.
\newblock Aggregated learning: A vector-quantization approach to learning neural network classifiers.
\newblock In \emph{Proceedings of the AAAI Conference on Artificial Intelligence}, pages 5810--5817, 2020.

\bibitem[Srivastava et~al.(2014)Srivastava, Hinton, Krizhevsky, Sutskever, and Salakhutdinov]{srivastava2014dropout}
Nitish Srivastava, Geoffrey Hinton, Alex Krizhevsky, Ilya Sutskever, and Ruslan Salakhutdinov.
\newblock Dropout: a simple way to prevent neural networks from overfitting.
\newblock \emph{The journal of machine learning research}, 15\penalty0 (1):\penalty0 1929--1958, 2014.

\bibitem[Sun et~al.(2022{\natexlab{a}})Sun, Masson, Thome, and Cord]{sunadapting}
R{\'e}my Sun, Cl{\'e}ment Masson, Nicolas Thome, and Matthieu Cord.
\newblock Adapting multi-input multi-output schemes to vision transformers.
\newblock In \emph{CVPR 2022 workshop on Efficient Deep Learning for Computer Vision (ECV)}, 2022{\natexlab{a}}.

\bibitem[Sun et~al.(2022{\natexlab{b}})Sun, Ram{\'e}, Masson, Thome, and Cord]{sun2022towards}
R{\'e}my Sun, Alexandre Ram{\'e}, Cl{\'e}ment Masson, Nicolas Thome, and Matthieu Cord.
\newblock Towards efficient feature sharing in mimo architectures.
\newblock In \emph{Proceedings of the IEEE/CVF Conference on Computer Vision and Pattern Recognition}, pages 2697--2701, 2022{\natexlab{b}}.

\bibitem[Vaswani et~al.(2017)Vaswani, Shazeer, Parmar, Uszkoreit, Jones, Gomez, Kaiser, and Polosukhin]{vaswani2017attention}
Ashish Vaswani, Noam Shazeer, Niki Parmar, Jakob Uszkoreit, Llion Jones, Aidan~N Gomez, {\L}ukasz Kaiser, and Illia Polosukhin.
\newblock Attention is all you need.
\newblock \emph{Advances in neural information processing systems}, 30, 2017.

\bibitem[Wen et~al.(2020)Wen, Tran, and Ba]{wen2020batchensemble}
Yeming Wen, Dustin Tran, and Jimmy Ba.
\newblock {BatchEnsemble}: an alternative approach to efficient ensemble and lifelong learning.
\newblock In \emph{International Conference on Learning Representations}, 2020.

\bibitem[Wenzel et~al.(2020)Wenzel, Snoek, Tran, and Jenatton]{wenzel2020hyperparameter}
Florian Wenzel, Jasper Snoek, Dustin Tran, and Rodolphe Jenatton.
\newblock Hyperparameter ensembles for robustness and uncertainty quantification.
\newblock In \emph{Neural Information Processing Systems}, 2020.

\bibitem[Yang et~al.(2023)Yang, Shi, Wei, Liu, Zhao, Ke, Pfister, and Ni]{yang2023medmnist}
Jiancheng Yang, Rui Shi, Donglai Wei, Zequan Liu, Lin Zhao, Bilian Ke, Hanspeter Pfister, and Bingbing Ni.
\newblock Medmnist v2-a large-scale lightweight benchmark for 2d and 3d biomedical image classification.
\newblock \emph{Scientific Data}, 10\penalty0 (1):\penalty0 41, 2023.

\bibitem[Zar(2005)]{zar2005spearman}
Jerrold~H Zar.
\newblock Spearman rank correlation.
\newblock \emph{Encyclopedia of biostatistics}, 7, 2005.

\bibitem[Zhao et~al.(2023)Zhao, Zhou, Li, Tang, Wang, Hou, Min, Zhang, Zhang, Dong, et~al.]{zhao2023survey}
Wayne~Xin Zhao, Kun Zhou, Junyi Li, Tianyi Tang, Xiaolei Wang, Yupeng Hou, Yingqian Min, Beichen Zhang, Junjie Zhang, Zican Dong, et~al.
\newblock {A Survey of Large Language Models}.
\newblock \emph{arXiv preprint arXiv:2303.18223}, 2023.

\end{thebibliography}
\newpage
\newpage

\appendix

In the Supplementary Material we first reexamine the notation used in the paper in Section~\ref{sec:appendix:notation}.
Then we provide a detailed comparison of SAE to generalised methods and baselines in Section~\ref{sec:appendix:comparison}.
Then we summarise the training and evaluation of SAE as an Algorithm in Section~\ref{sec:appendix:algorithm}.
We provide additional details on the experimental setup in Section~\ref{sec:appendix:experiments_setup}.
Then we describe the practical implementation details and hardware cost of SAE in Section~\ref{sec:appendix:hardware}.
We provide additional results for the baseline performance evaluation in Section~\ref{sec:appendix:baseline}.
In Section~\ref{sec:appendix:changing_k_n}, we investigate the impact of changing the number of inputs $N$ and the number of exits $K$ on the OOD test sets.
In Section~\ref{sec:appendix:depth}, we examine the impact of $N$ and $K$ on the depth preference of the learnt exit weights.
Lastly, in Section~\ref{sec:appendix:hyperparameter_influence}, we detail the impact of the introduced hyperparameters on the performance of SAE.

\section{Notation}\label{sec:appendix:notation}

The Table~\ref{tab:notation} summarises the notation used in the paper.
In this work, we assume a standard single exit NN, a singular member of the naive ensemble, to be a function which maps an input $x$ to an output $\hat{y}$ for target $y$, with learnable weights $\mathbf{w}$.
The function can be decomposed into $D$ layers $h^D(\cdot) \circ f^D(\cdot) \circ f^{D-1}(\cdot) \circ \dots \circ f^1(\cdot)$, where $f^j(\cdot)$ is the $j$-th layer and $h^D(\cdot)$ is the prediction head at the end of the network giving the output $p(y|x, \mathbf{w}) = p(y|\hat{y});\hat{y}=h^D(f^D(f^{D-1}(\dots f^1(x))))$.
For generality, we assume a layer $f^j(\cdot)$ to be an arbitrary weighted operation, such as a combination of convolution, batch normalisation, pooling or activation layers~\citep{simonyan2014very}, residual block~\citep{he2016deep} or a transformer encoder layer~\citep{vaswani2017attention} and a prediction head $h(\cdot)$ to be a fully-connected layer preceded by a global average pooling layer and any necessary parametric or non-parametric reshaping operation to convert the hidden representation into the desired output shape.

\begin{table}[ht]
\centering
\begin{sc}
\begin{adjustbox}{max width=\linewidth}
\begin{tabular}{|l|c|}
\hline
\textbf{Symbol} & \textbf{Description}  \\
\hline\hline 
$NN$            & Neural Network \\
$\mathcal{D}$   & Dataset \\
$|\mathcal{D}|$ & Number of samples in the dataset \\
$D$             & Depth of the network \\
$\{f^j(\cdot)\}_{j=1}^D$ & Layers of the NN \\
$\{h^j(\cdot)\}_{j=1}^D$ & Prediction heads of the NN \\
$\mathbf{w}$    & Weights of the NN \\
$N$             & Number of simultaneously processed inputs/number of submembers \\
$K$             & Maximum number of exits used by submember \\
$\{d_i\}_{i=1}^N$ & Latent depth variables for each submember in $N$ \\
$\{\theta_i^j\}_{i,j=1}^{N,K}$ & Weights of the $j$-th exit of the $i$-th submember \\
$\mathbf{\theta}^*$ & Largest depth weights selected with respect to $K$ \\
$\{l_i^j\}_{i,j=1}^{N,K}$ & Logits of the $j$-th exit of the $i$-th submember \\
$x$             & Input to the NN \\
$y$             & Target output \\
$\{\hat{y}_i^j\}_{i,j=1}^{N,K}$ & Predicted outputs of the $j$-th exit of the $i$-th submember \\
$\mathbf{x} = \{x_i\}_{i=1}^N$ & $N$-member input \\
$\mathbf{X}_i = \{\mathbf{x}_{i}^{(j)}\}_{j=1}^{|\mathcal{D}|}$ & Input samples for the $i$-th submember across the dataset $\mathcal{D}$ \\
$Y_i = \{y_{i}^{(j)}\}_{j=1}^{|\mathcal{D}|}$ & Target outputs for the $i$-th submember across the dataset $\mathcal{D}$ \\
$\mathcal{L}$   & Loss function \\
$q(\cdot)$      & Variational distribution \\
$B$             & Batch size \\
$H$             & Image height \\
$W$             & Image width \\
$C$             & Image channels \\
$F$             & Feature size \\
$S$             & Sequence length \\
$P$             & Patch size \\
$O$             & Number of outputs \\
$\alpha(t)$        & Re-weighting factor between data fit and regularization at time $t$ \\
$T(t)$             & Temperature parameter at time $t$ \\
$i(t)$             & Input repetition probability at time $t$ \\
\hline
\end{tabular}
\end{adjustbox}
\end{sc}
\vspace{0.2cm}
\caption{Notation used in the paper.}
\label{tab:notation}
\end{table}

\section{Detailed Comparison to Generalised Methods and Baselines}\label{sec:appendix:comparison}

\textbf{Generalised Methods:}
A distinctive feature of SAE is its ability to determine the most effective exit depth for each input, a capability previously investigated in SE NN, EE and MIMMO~\citep{antoran2020depth, ferianc2023mimmo}, but not in MIMO~\cite{havasi2020training}.
MIMMO~\cite{ferianc2023mimmo}, where $N \geq 2$ and $K=D$, is the closest approach to SAE. 
However, MIMMO did not consider creating a search space generalising the previous methods or the introduction of the new in-between approaches.
It also learns the exit depth for each input but uses all the exits during evaluation.
As empirically observed, small capacity NNs might not have enough capacity to make meaningful predictions at all depths, hence the need for the $K$ parameter.
SAE introduces the $K$ parameter and has to choose an appropriate sampling strategy for the depth variables, which allows for the selection of the most effective exits for each input, features not present in MIMMO.
Additionally, SAE diverges from the MIMMO by not adopting a common prediction head across all exits, which we empirically found to degrade performance.
Moreover, SAE introduced the hyperparameter scheduling of $i(t)$ and the additional regularisation terms $\alpha(t)$, $T(t)$, which were not considered in MIMMO.
Last but not least, MIMMO was demonstrated only for classification using a convolutional architecture, while SAE is demonstrated for classification and regression using a variety of architectures.

In essence, SAE's objective is not to rival the generalised methods but to amalgamate them into a cohesive framework facilitated by setting $N$ and $K$ and empirically investigate the trade-offs between the generalised approaches and their novel in-between counterparts explorable through our proposed search space and problem formulation. 

\paragraph{Naive NN Ensemble}\label{sec:appendix:comparison:naive_ensemble}
The naive ensemble~\cite{lakshminarayanan2017simple} comprises $N$ independent models, each with the same architecture as the base model.
Each model is trained independently on the same training data, scaling the training time linearly with the number of inputs $N$.
During the evaluation, input is processed through each model, and the predictions are averaged across all models, scaling the inference time or consumed memory linearly with the number of inputs $N$.

\paragraph{Batch Ensemble}\label{sec:appendix:comparison:batch_ensemble}
The Batch Ensemble~\cite{wen2020batchensemble} is contained in a single architecture, making it more efficient than the naive ensemble and requiring only a single training and inference run.
However, to apply elementwise rank-1 weight matrices to the input and output of each layer, scaling with the number of inputs $N$, all the linear and convolutional layers must be re-implemented.
This re-implementation might require custom kernels for the forward and backward passes for efficient computation.
This increases the number of FLOPs and marginally increases the number of parameters for each layer.
Practically, Batch Ensemble defines the parameters for each input in $N$ manually through the rank-1 weight decomposition.

\paragraph{Monte Carlo Dropout}\label{sec:appendix:comparison:mc_dropout}
Monte Carlo Dropout~\cite{gal2016dropout} is contained in a single architecture, making it more efficient than the naive ensemble and requiring only a single training run.
A dropout layer~\cite{srivastava2014dropout} is added before each linear and convolutional layer, randomly dropping out a fraction of the input nodes.
Therefore, dropout requires an efficient random number generator and dropout mask application before each linear or convolutional layer.
The repeated forward passes are used to obtain the Monte Carlo samples for the predictions, scaling the inference time linearly with the number of inputs $N$. 
At the same time, the memory consumed by parameters stays constant.

Compared to the baseline methods, SAE is trained and evaluated in a single run, making it more efficient than the compared methods.
At the same time, it does not require any re-implementation of the layers or random number generators, making it more efficient than Batch Ensemble and Monte Carlo Dropout.
It automatically learns the weights which react to the features of the input data, not requiring manual parameterisation, making it more efficient than Batch Ensemble.

\section{Algorithm}\label{sec:appendix:algorithm}

The Algorithm~\ref{alg:sae} summarises the training and evaluation procedures for the Single Architecture Ensemble (SAE) framework.
The SAE training operates over a fixed number of steps $t_{end}$. 
For each step, it iterates over a batch size $B$, $\mathbf{x}^{(b)}=\{\mathbf{x}_{i}^{(b)}\}^{N}_{i=1}$ and $\mathbf{y}^{(b)}=\{y_{i}^{(b)}\}^{N}_{i=1}$, sampled from the dataset $\mathcal{D}$ according to the input repetition factor $i(t)$ and $N$.
It then computes predictions $\{\hat{y}_{i}^{j,(b)}\}^{N,D}_{i,j=1}$ using the current model and samples the depth variables from a distribution $q(d|\mathbf{\theta})$ for all $N$.
Using these predictions and the sampled depth variables, the loss $\mathcal{L}(\mathbf{w}, \mathbf{\theta})$ is calculated, and $\mathbf{w}, \mathbf{\theta}$ are updated through backpropagation.
The evaluation involves retaining the top $K$ parameters of $\mathbf{\theta}$ while setting the rest of the logits to negative infinity. 
It repeats input $\mathbf{x}^{*}=\{x_{i}\}^{N}_{i=1}$ $N$ times and computes the weighted prediction $\hat{y}^*$.

\begin{algorithm}[t]
    \caption{Single Architecture Ensemble (SAE)}\label{alg:sae}
    \begin{algorithmic}[1]
    \Procedure{Train}{$\mathcal{D}$,$\mathbf{w}$,$\mathbf{\theta}$,$B$,$N$,$D$,$K$,$\alpha(t)$,$T(t)$,$i(t)$, $t_{end}$}
    \For{$t=1$ \textbf{to} $t_{end}$}
        \For{$b=1$ \textbf{to} $B$}
                \State Sample $\mathbf{x}^{(b)}$, $\mathbf{y}^{(b)}$ from $\mathcal{D}$ at batch index $b$ with respect to input repetition factor $i(t)$ and $N$ 
        \State Predict $\{\hat{y}_{i}^{j,(b)}\}^{N,D}_{i,j=1}$
        \EndFor
        \State Sample $\{d_i\}_{i=1}^N$ 
        \State Compute $\mathcal{L}(\mathbf{w}, \mathbf{\theta})$ using Equation~\ref{eq:elbo_3}
        \State Backpropagate and update $\mathbf{w}$, $\mathbf{\theta}$
    \EndFor
    \EndProcedure
    
    \Procedure{Evaluate}{$\mathcal{D}$,$\mathbf{w}$,$\mathbf{\theta}$,$N$,$D$,$K$,$T(t_{end})$}
    \State Keep top $K$ $\mathbf{\theta}^*$ and set the rest to -inf for all $i\in N$; initialise top $K$ active exits
    \State Repeat input $\mathbf{x}^{*}=\{x_{i}\}^{N}_{i=1}$ $N$ times
    \State Predict $\{\hat{y}_{i}^{j}\}^{N,K}_{i,j=1}$
        \State Predict $\hat{y}^* = \frac{1}{N} \sum_{i,j=1}^{N,K} \hat{y}_i^j \theta_i^{*j}$
    \EndProcedure
\end{algorithmic}
\end{algorithm}

\section{Experimental Setup}\label{sec:appendix:experiments_setup}

Our code is implemented in PyTorch. 
No specific hardware optimisations were used to accelerate the training or evaluation for any methods.
We fixed the number of epochs to 50 for all datasets except TinyImageNet, for which we set it to 100.
We use Adam optimiser, starting with a learning rate of 3e-4 and a weight decay of 1e-5, except TinyImageNet, for which we set the weight decay to 0, and cosine annealing learning rate scheduler~\citep{loshchilov2016sgdr} across all datasets.
We used batch size 64 for PneumoniaMNIST and RetinaMNIST and 128 for TinyImageNet and BloodMNIST.
For Batch Ensemble, we had to find a batch size close to the other experiments' batch size and a multiple of $1\leq N \leq 4$. 
Therefore, for experiments where batch size 64 was used, we used 72 for Batch Ensemble, and for experiments where batch size 128 was used, we used 144 for Batch Ensemble.
The batch repetition factor~\citep{havasi2020training} for SAE was set to 2, except for TinyImageNet, where we set it to 1. 
We used gradient clipping with a maximum norm of 5.0 to avoid exploding gradients across all experiments.
We used the default PyTorch initialisation for all layers. 
We fixed the base  hyparameters (HP)s, such as the learning rate, weight decay, batch size, and number of epochs for all methods, to the same values for fair comparison and to demonstrate the ease of implementing SAE based on existing solutions.

We used the validation data to guide the multi-objective Bayesian optimisation (MOBO) from syne-tune~\citep{salinas2022syne,paria2020flexible} to perform HPO.
We used 10\% of the training data as validation data across all datasets.
We enable 50 random initialisations for exploration.
We used 2 GTX 1080 and 2 RTX 2080 GPUs in one machine, giving us 4 workers for the HPO.
We run for 8 hours on 4 GPUs for BloodMNIST, PneumoniaMNIST, and RetinaMNIST and 8 days on 4 GPUs for TinyImageNet.
For SAE we set the search space for $i_{start}(t)$ to $[0.0, 1.0]$ and $i_{end}(t)$ to $[0.0, 1.0]$.
For $\alpha_{start}(t)$ and $\alpha_{end}(t)$ we set the search space to $[0.00001, 1.0]$ on a log scale.
For $T_{start}(t)$ and $T_{end}(t)$ we set the search space to $[0.001, 1.0]$ on a log scale.
We set the search space for the dropout rate to $[0.0, 1.0]$ on a linear scale.
Dropout was not added before the first layer and after the last layer in any architecture.
$N=4$ Monte Carlo samples are used during evaluation for Dropout.
For ViT, it was added only where dropout would usually be added in the original architecture.

For TinyImageNet, we performed experiments on a ResNet with [3,4,6,3] residual blocks per stage, strides [2,2,2,2] and channels [64,128,256,512] within which we only enabled the search for the number of inputs $N$ and a depth $ D \in \{1,4\}$, in addition to the regularisation parameters.
Here, depth refers to the stage depth, not the overall depth of the network.
With $D=4$, this is the standard ResNet-34 architecture.

For ViT and RetinaMNIST, we enable to vary the encoder depth $D \in \{1,5\}$, width multiplier for base hidden dimension 192 as $W \in \{1,4\}$, the patch size 4, 4 heads and embedding dimension 192 and no dropout enabled by default.
The depth refers to the number of encoder layers in ViT.
The width multiplier multiplies the base hidden dimension of the ViT encoder.
For ViT, we add a token for each additional input, but we design the final layer as a pooling of all the tokens to arrive at the prediction. 
We empirically observed that this leads to better performance than just the added tokens for prediction. 

For VGG and BloodMNIST, we enable to vary $D \in \{1,5\}$ and $W \in \{1,4\}$ with base width $[4,8,16,32,32]$ per stage and $[1,1,1,1,1], [1,1,2,2,2]$ blocks per stage, we omit the repeated memory-expensive linear layers at the end of the network and replace them with a single predictive linear layer.
Each block consisted of a convolutional layer, batch normalisation, and ReLU; the first three stages included a max pooling layer in the last block of each stage.
The depth refers to the number of stages in VGG.
The width multiplier multiplies the base width of the VGG network per stage.

For FC and PneumoniaMNIST, we design a residual fully connected network where the input is processed through an initial linear layer, upscaled to the hidden dimension, and then processed through a series of residual blocks, of a linear layer, batch normalisation, ReLU, we enable to vary $D \in \{1,5\}$ and $W \in \{1,4\}$ with base width 128.
The depth refers to the number of residual blocks in FC.
The width multiplier multiplies the base width of the FC network.
We varied the search space size from the smallest, for example, on TinyImageNet, to the largest, for instance, on BloodMNIST, to demonstrate the adaptability of SAE to different search spaces.
Note that for SAE, the depth is permanently fixed to the maximum.

The classification on TinyImageNet, BloodMNIST and RetinaMNIST was performed by processing the outputs of the last layer through a softmax function.
The mean over all predictions was used as the final prediction for all the methods.
For regression on RetinaMNIST, one output node was used as the mean, and the second output was processed through an exponential function to obtain the variance for minimising the Gaussian negative log-likelihood per example.
The mean over all predictions was computed as the mean across all the mean outputs. 
The variance was computed as the mean of the variances plus the variance of the means by the law of total variance.

\revision{The SAE implementation is available via YAMLE~\citep{ferianc2024yamle} at \url{https://github.com/martinferianc/yamle}.}

\subsection{Datasets}\label{sec:appendix:datasets}

We used the following datasets for our experiments:

\paragraph{MedMNIST}\label{sec:appendix:datasets:medmnist}

MedMNIST~\citep{yang2023medmnist} offers diverse medical imaging tasks: binary/multi-class classification and regression in a standardised, MNIST-like format that enables benchmarking with a reasonable computational cost. 
Given its multi-fidelity nature covering grayscale or RGB images, class imbalance and varying number of samples in the datasets, it presents a more realistic and challenging benchmark than other clean datasets such as MNIST, FashionMNIST or CIFAR-10/100.

\begin{itemize}[leftmargin=*, itemsep=0em]
    \item \textbf{PneumoniaMNIST} is derived from a dataset of pediatric chest X-ray images aimed at binary classification between pneumonia and normal cases. 
    It includes 5,856 images originally pre-processed to $28 \times 28$ grayscale images.
    \item \textbf{BloodMNIST} consists of 17,092 blood cell microscope images from individuals without infection or disease, categorised into 8 classes representing different cell types.
    The images are RGB and originally pre-processed to $28 \times 28$.
    \item \textbf{RetinaMNIST} originates from the DeepDRiD challenge, offering 1,600 retina fundus images for regression to grade diabetic retinopathy severity on a five-level scale.
    The images are RGB and originally pre-processed to $28 \times 28$.
\end{itemize}

In pre-processing, we pad the images to 32 pixels, random cropping and normalisation are applied to the training data, and normalisation is only used for the validation and test data.
For RetinaMNIST, we normalised the ordered levels of the regression target to $[0,1]$, and we modelled the prediction as parameters of a Gaussian distribution with a mean and variance output to enable us to model the uncertainty in the predictions and benchmark the negative log-likelihood.

\paragraph{TinyImageNet}\label{sec:appendix:datasets:tinyimagenet}

TinyImageNet~\citep{le2015tiny} is a subset of the ImageNet dataset~\citep{deng2009imagenet} with 200 classes and 500 images per class, resulting in 100,000 training images and 10,000 test images.
Compared to MedMNIST, TinyImageNet is a more challenging dataset with higher-resolution images and more classes.
However, the dataset is balanced, and the images are relatively clean.
The images are RGB and pre-processed initially to $64 \times 64$.
In pre-processing, random cropping and normalisation are applied to the training data, and normalisation is only used for the validation and test data.
For TinyImageNet, we used 10\% of the training data as validation data.

We considered datasets of varying image sizes ($64 \times 64$ against $28 \times 28$) and varying numbers of classes (200 against 2 and 8), fidelity (RGB against grayscale) and class imbalance (balanced against imbalanced), sizes (100,000, 17,092, 5,856 and 1,600) to demonstrate the adaptability of SAE to different datasets and environments.
Naturally, given the dataset size, architecture, and parameterisation, the training time for the experiments varied, which also varied the number of HPO runs we performed within our budget. 
This aimed to demonstrate the robustness of SAE to different search budgets.

\paragraph{OOD Datasets}\label{sec:appendix:datasets:ood}

To create the OOD test data, we used the augmentations from~\citep{hendrycks2019benchmarking} and applied them to the test data before normalisation.
These corruptions include, for example, adding snow or fog to the image, changing the brightness or saturation of the image or blurring the image across 5 intensities.
We averaged the performance across all the augmentations and severities except FSGM to obtain scalar metrics for the OOD test data.
For PneumoniaMNIST, we had to convert the grayscale image to RGB, apply the augmentations, and then convert it back to grayscale.
The OOD test aimed to demonstrate the robustness of SAE to OOD data where naive ensembles are known to outperform many state-of-the-art methods~\citep{ovadia2019can}.

We chose feed-forward, residual, convolutional, fully connected, or transformer architectures of various capacities, a combination of classification and regression tasks, and balanced and unbalanced datasets in RGB or grayscale, with many or few data samples across ID and OOD data to demonstrate the versatility of our framework in different applications.

\subsection{Algorithmic Metrics}\label{sec:appendix:algorithmic_metrics}

We used the following metrics to evaluate the performance of the methods:

\paragraph{Accuracy}\label{sec:appendix:metrics:accuracy}

Accuracy is the percentage of correctly classified samples defined in Equation~\ref{eq:accuracy}.
We want to maximise accuracy for classification tasks.
\begin{equation}\label{eq:accuracy}
    \text{Accuracy} = \frac{1}{|\mathcal{D}|}\sum_{i=1}^{|\mathcal{D}|} \mathbbm{1}_{\hat{y}_i = y_i}
\end{equation}
where $\mathcal{D}$ is the dataset, $\hat{y}_i$ is the predicted label for the $i$-th sample and $y_i$ is the ground truth label for the $i$-th sample.

\paragraph{F1 Score}\label{sec:appendix:metrics:f1_score}

F1 score is the harmonic mean of precision and recall defined in Equation~\ref{eq:f1_score}.
It is more suitable for imbalanced datasets than accuracy.
We want to maximise the F1 score for classification tasks.
\begin{equation}\label{eq:f1_score}
    \text{F1 Score} = \frac{2}{\frac{1}{\text{Precision}} + \frac{1}{\text{Recall}}} = \frac{2 \times \text{Precision} \times \text{Recall}}{\text{Precision} + \text{Recall}}
\end{equation}
where $\text{Precision} = \frac{\text{TP}}{\text{TP} + \text{FP}}$ and $\text{Recall} = \frac{\text{TP}}{\text{TP} + \text{FN}}$.
TP, FP and FN are the number of true positives, false positives and false negatives, respectively.
We measured the macro F1 score, the average of the F1 scores for each class, in a one-vs-all fashion.

\paragraph{Expected Calibration Error}\label{sec:appendix:metrics:ece}

Expected calibration error (ECE)~\citep{guo2017calibration} measures a model's confidence calibration.
A model is well-calibrated if the confidence of the model's predictions matches the accuracy of the predictions.
We want to minimise ECE for classification tasks.
ECE is defined in Equation~\ref{eq:ece}.

\begin{equation}\label{eq:ece}
    \text{ECE} = \sum_{m=1}^M \frac{|B_m|}{|\mathcal{D}|} \left| \text{acc}(B_m) - \text{conf}(B_m) \right|
\end{equation}

where $\mathcal{D}$ is the dataset, $M$ is the number of bins, $B_m$ is the $M$-th bin, $|B_m|$ is the number of samples in the $M$-th bin, $\text{acc}(B_m)$ is the accuracy of the $M$-th bin and $\text{conf}(B_m)$ is the confidence of the $M$-th bin.
We used 15 bins for all datasets.

\paragraph{Class-Conditional Expected Calibration Error}\label{sec:appendix:metrics:cc_ece}

The class-conditional expected calibration error (CC-ECE)~\citep{guo2017calibration} measures the calibration of a model for each class.
It is better suited for imbalanced datasets than ECE.
We want to minimise CC-ECE for classification tasks.
CC-ECE is defined in Equation~\ref{eq:cc_ece}.
\begin{equation}\label{eq:cc_ece}
    \text{CC-ECE} = \sum_{c=1}^C \frac{|\mathcal{D}_c|}{|\mathcal{D}|} \text{ECE}(\mathcal{D}_c)
\end{equation}
where $\mathcal{D}_c$ is the set of samples with the highest probability of class $c$ and $C$ is the number of classes.

\paragraph{Negative Log-Likelihood}\label{sec:appendix:metrics:nll}

Negative log-likelihood (NLL) is a proper scoring rule for probabilistic models.
It measures the average negative log probability of the ground truth labels for classification and the average negative log-likelihood of the ground truth values for regression.
We want to minimise NLL for classification and regression tasks.
NLL for classification is defined in Equation~\ref{eq:nll_classification}, and NLL for regression is described in Equation~\ref{eq:nll_regression}.

\begin{equation}\label{eq:nll_classification}
    \text{NLL} = -\frac{1}{|\mathcal{D}|}\sum_{i=1}^{|\mathcal{D}|} \mathbbm{1}_{\hat{y}_i = y_i} \log \hat{y}_{i}
\end{equation}
where $\hat{y}_{i}$ is the predicted probability of the ground truth label for the $i$-th sample.

\begin{equation}\label{eq:nll_regression}
    \text{NLL} = -\frac{1}{|\mathcal{D}|}\sum_{i=1}^{|\mathcal{D}|} \log \mathcal{N}(y_i| \hat{\mu}_{i}, \hat{\sigma}_{i}^2)
\end{equation}
where $\mathcal{N}$ is the normal distribution, $y_i$ is the ground truth value for the $i$-th sample, $\hat{\mu}_{i}$ is the predicted mean for the $i$-th sample and $\hat{\sigma}_{i}$ is the predicted standard deviation for the $i$-th sample.

\paragraph{Mean Squared Error}\label{sec:appendix:metrics:mse}

Mean squared error (MSE) is a regression metric measuring the average squared difference between the predicted and ground truth values.
We want to minimise MSE for regression tasks.
MSE is defined in Equation~\ref{eq:mse}.
\begin{equation}\label{eq:mse}
    \text{MSE} = \frac{1}{|\mathcal{D}|}\sum_{i=1}^{|\mathcal{D}|} (\hat{\mu}_i - y_i)^2
\end{equation}
where $\hat{\mu}_i$ is the predicted value for the $i$-th sample and $y_i$ is the ground truth value for the $i$-th sample.

We use the F1 score and CC-ECE to measure the performance under unbalanced settings instead of accuracy or ECE insensitive to class imbalance.

\section{Practical Implementation and Hardware Cost}\label{sec:appendix:hardware}

In this Section, we analyse the hardware cost of the SAE approach in detail, covering everything from modifying the input layer to adding the early exits.

\subsection{Input Layer}\label{sec:appendix:hardware:input_layer}

The input layer is modified to accept a set of $N$ inputs, one for each input in $N$.
In practice, considering image data of size $C\times H\times W$, the input layer is modified to accept a set of $N$ inputs of size $N\times C\times H\times W$.
That is done simply by concatenating the $N$ inputs along the channel dimension.

\paragraph{FC}\label{sec:appendix:hardware:input_layer:fc}

The input layer is, by default, a fully connected layer for the residual FC network.
The difference from standard NN is simply increasing the input dimension from $C$ to $ N\times C$ while leaving the output dimension $F$ unchanged.

The number of parameters for the input layer is:
\begin{itemize}[itemsep=0em]
    \item Fully Connected: $F\times N\times C + F$.
\end{itemize}

This is in comparison to the standard NN where the number of parameters for the input layer is: $F\times C + F$ or when $N=1$.

The number of FLOPs for the input layer is:
\begin{itemize}[itemsep=0em]
    \item Fully Connected: $F\times N\times C + F$.
\end{itemize}

This is in comparison to the standard NN where the number of FLOPs for the input layer is: $F\times C + F$ or when $N=1$.

\paragraph{ResNet \& VGG}\label{sec:appendix:hardware:input_layer:resnet_vgg}

For ResNet and VGG, the input layer is, by default, a convolutional layer. 
The difference to standard NN is simply increasing the input channels in the convolution from $C$ to $N\times C$ while leaving the output channel count $F$ unchanged.

The number of parameters for the input layer is:
\begin{itemize}[itemsep=0em]
    \item Convolution: $F\times N\times C\times K\times K + F$, $K$ is the kernel size.
\end{itemize}

This is compared to the standard NN where the input layer parameters are $F\times C\times K\times K + F$ or when $N=1$.

The number of FLOPs for the input layer is:
\begin{itemize}[itemsep=0em]
    \item Convolution: $F\times N\times C\times K\times K\times H\times W+ F\times H\times W$ and assuming stride 1.
\end{itemize}

This is in comparison to the standard NN where the number of FLOPs for the input layer is: $F\times C\times K\times K\times H\times W+ F\times H\times W$, again assuming stride 1 or when $N=1$.

\paragraph{ViT}\label{sec:appendix:hardware:input_layer:vit}

For ViT, the input layer creating the  $S$ embeddings from input patches is, by default, a fully connected layer.
The difference to standard NN is simply increasing the input dimension from $P\times P\times C$ to $P\times P\times N\times C$ where $P$ is the patch size.

The number of parameters for the input layer is:
\begin{itemize}[itemsep=0em]
    \item Fully Connected: $F\times P\times P\times N\times C+ F$.
\end{itemize}

This is in comparison to the standard NN where the number of parameters for the input layer is: $F\times P\times P\times C+ F$ or when $N=1$.

The number of FLOPs for the input layer is:
\begin{itemize}[itemsep=0em]
    \item Fully Connected: $F\times P\times P\times N\times  C\times S+ F\times S$.
\end{itemize}

This is in comparison to the standard NN where the number of FLOPs for the input layer is: $F\times P\times P\times C\times S+ F\times S$ or when $N=1$.

\subsection{Early Exits}\label{sec:appendix:hardware:early_exits}

The hardware cost of adding the early exits to the FC, ResNet, VGG and ViT are as follows. 
We denote $\{n^i\}_{i=1}^D$ as the number of inputs using the exit at depth $i$ for all depths $D$ and $F^{last}$ and $\{F^i\}_{i=1}^D$ as the feature size at the last layer and the depth $i$ respectively.
We define the $F^{last}$ as the last hidden dimension of the backbone, e.g. at the position of global average pooling for ResNet for the VGG or as the embedding dimension for ViT.

\paragraph{FC}\label{sec:appendix:hardware:early_exits:fc}

We add an early exit after a residual block for the residual FC network.
The early exit is composed of a fully connected layer with $F^{last}$ output nodes, batch normalisation layer, ReLU activation function and a prediction head with $F^{last}$ input nodes and $n^i\times O$ output nodes for the $i$-th exit.
Therefore, the number of parameters for the early exit, conditioned on the number of inputs $n^i$ using the exit at depth $i$, is:

\begin{itemize}[itemsep=0em]
    \item Fully Connected: $F^{i}\times F^{last} + F^{last}$.
    \item Batch Normalisation: $2\times F^{last}$.
    \item Prediction head: $F^{last} \times n^i\times O +  n^i\times O$.
\end{itemize}

The number of FLOPs for the early exit, conditioned on the number of inputs $n^i$ using the exit at depth $i$, is:

\begin{itemize}[itemsep=0em]
    \item Fully Connected: $F^{i}\times F^{last} + F^{last}$.
    \item Batch Normalisation: $2\times F^{last}$.
    \item ReLU: $F^{last}$.
    \item Prediction head: $F^{last} \times n^i\times O + n^i\times O$.

\end{itemize}

\paragraph{ResNet \& VGG}\label{sec:appendix:hardware:early_exits:resnet_vgg}

For ResNet and VGG, we add an early exit after a set of residual or feed-forward blocks.
The early exits are composed of a reshaping $1\times1$ convolutional layer with stride 1 with $F^{last}$ output channels, a batch normalisation layer, ReLU activation function, global average pooling followed by the prediction head with $F^{last}$ input nodes and $n^i\times O$ output nodes for the $i$-th exit.
Therefore, the number of parameters for the early exit, conditioned on the number of inputs $n^i$ using the exit at depth $i$, is:
\begin{itemize}[itemsep=0em]
    \item Convolution: $F^{i}\times F^{last}\times 1\times 1 + F^{last}$.
    \item Batch Normalisation: $2\times F^{last}$.
    \item Prediction head: $F^{last} \times n^i\times O +  n^i\times O$, where $O$ is the number of output nodes.
\end{itemize}

The number of FLOPs for the early exit, conditioned on the number of inputs $n^i$ using the exit at depth $i$, is:

\begin{itemize}[itemsep=0em]
    \item Convolution: $F^{i}\times F^{last}\times 1\times 1\times H^i\times W^i + F^{last}\times H^i\times W^i$, $H^i$ and $W^i$ are the height and width of the feature map at the given depth $i$.
    \item Batch Normalisation: $2\times F^{last}\times H^i\times W^i$.
    \item ReLU: $F^{last}\times H^i\times W^i$.
    \item Global Average Pooling: $F^{last}\times H^i\times W^i$.
    \item Prediction head: $F^{last} \times n^i\times O+ n^i\times O$.
\end{itemize}

\paragraph{ViT}\label{sec:appendix:hardware:early_exits:vit}

For ViT, the early exits are composed of a fully connected layer with $F$ input and output channels, where $F$ is the embedding dimension, followed by a GeLU activation, layer normalisation, reduction via average pooling across the sequence dimension, and a prediction head which is a linear layer with $F$ input nodes and $n^i\times O$ output nodes.
Therefore, the number of parameters for the early exit, conditioned on the number of inputs $n^i$ using the exit at depth $i$, is:
\begin{itemize}[itemsep=0em]
    \item Fully connected layer: $F \times F + F$.
    \item Layer normalisation: $2 \times F$.
    \item Prediction head: $F \times n^i\times O +  n^i\times O$.
\end{itemize}

The FLOPs for the early exit, conditioned on the number of inputs $n^i$ using the exit at depth $i$, are:

\begin{itemize}[itemsep=0em]
    \item Fully connected layer: $F \times F \times (S+N) +  F \times (S+N)$, where $S$ is the sequence length.
    \item GeLU: $F \times (S+N)$.
    \item Layer normalisation: $2 \times F \times (S+N)$.
    \item Reduction: $F \times (S+N)$.
    \item Prediction head: $F \times n^i\times O + n^i\times O$.
\end{itemize}

The early exits are added during inference only for the top $K$ chosen early exits decided by the learnt variational parameters $\mathbf{\theta}$ for each input in $N$.

\revision{In summary, increasing $N$ is simpler than increasing $K$.
If $N$ is increased, only the input and exit layers need to be modified, whereas if $K$ is increased, the early exits need to be added which favours increasing $N$ over $K$.
The overall training time of the model is marginally increased by $N$ or $K$ in comparison to an SE NN, however, the early exits introduce branching in the architecture and if not implemented efficiently, the training time can increase.
Likewise, the evaluation time is only marginally increased by $N$ or $K$ and caution should be taken to implement the early exits efficiently to avoid a significant increase in evaluation time.}

\section{Supplementary for Baseline Performance}\label{sec:appendix:baseline}

\subsection{NLL and OOD Data}\label{sec:appendix:baseline:nll_ood}

\begin{figure}
    \centering
    
    \begin{subfigure}[t]{0.22\textwidth}
        \centering
        \includegraphics[width=\textwidth]{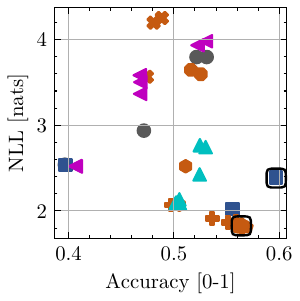}
        \caption{ID ACC, NLL}
        \label{fig:baselines:tinyimagenet:id:accuracy_nll}
    \end{subfigure}
    \hfill
    \begin{subfigure}[t]{0.23\textwidth}
        \centering
        \includegraphics[width=\textwidth]{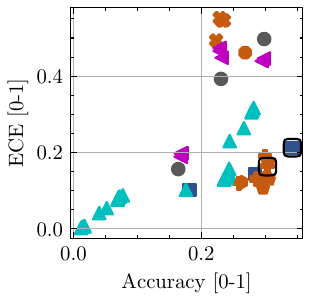}
        \caption{OOD ACC, CAL}
        \label{fig:baselines:tinyimagenet:ood:accuracy_calibration}
    \end{subfigure}
    \hfill
    \begin{subfigure}[t]{0.22\textwidth}
        \centering
        \includegraphics[width=\textwidth]{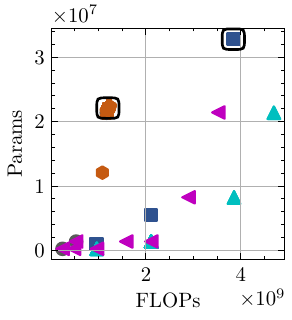}
        \caption{OOD FLOPs, Params}
        \label{fig:baselines:tinyimagenet:ood:flops_params}
    \end{subfigure}
    \hfill
    \begin{subfigure}[t]{0.22\textwidth}
        \centering
        \includegraphics[width=\textwidth]{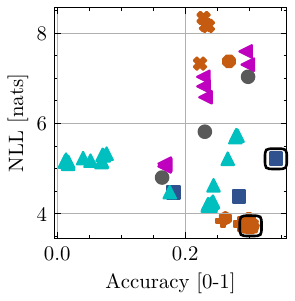}
        \caption{OOD ACC, NLL}
        \label{fig:baselines:tinyimagenet:ood:accuracy_nll}
    \end{subfigure}
    \vspace{0.2cm}
    \caption{Comparison on ID and OOD test sets for TinyImageNet, with respect to \textbf{\textcolor{Gray}{Standard NN $\newmoon$}}, \textbf{\textcolor{Blue}{NN Ensemble $\blacksquare$}}, 
    \textbf{\textcolor{Orange}{SAE: 
    I/B: $N\geq 2, 2 \leq K < D$ \PlusMarker, 
    EE: $N=1, K \geq 2$ \OctagonMarker, 
    MIMMO: $N\geq 2, K=D$ \PentagonMarker, 
    MIMO: $N\geq 2, K=1$ \HexagonMarker, 
    SE NN: $N=1, K=1$ \TimesMarker}}, 
    \textbf{\textcolor{Cyan}{MCD $\blacktriangle$}}, 
    \textbf{\textcolor{Pink}{BE $\blacktriangleleft$}}.
    The black outlines denote the configurations compared in the text.}
    \label{fig:baselines:tinyimagenet:id_ood}
\end{figure}

\begin{figure}
    \centering
    
    \begin{subfigure}[t]{0.22\textwidth}
        \centering
        \includegraphics[width=\textwidth]{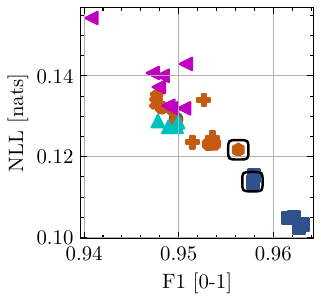}
        \caption{ID ACC, NLL}
        \label{fig:baselines:bloodmnist:id:f1_nll}
    \end{subfigure}
    \hfill
    \begin{subfigure}[t]{0.23\textwidth}
        \centering
        \includegraphics[width=\textwidth]{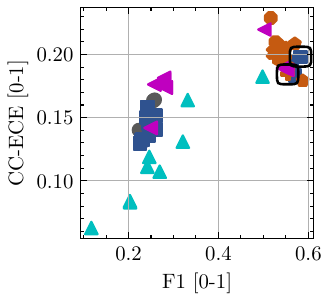}
        \caption{OOD ACC, CAL}
        \label{fig:baselines:bloodmnist:ood:f1_classconditionalcalibration}
    \end{subfigure}
    \hfill
    \begin{subfigure}[t]{0.24\textwidth}
        \centering
        \includegraphics[width=\textwidth]{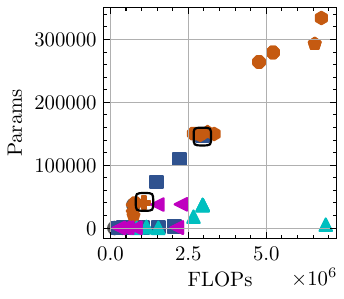}
        \caption{OOD FLOPs, Params}
        \label{fig:baselines:bloodmnist:ood:flops_params}
    \end{subfigure}
    \hfill
    \begin{subfigure}[t]{0.22\textwidth}
        \centering
        \includegraphics[width=\textwidth]{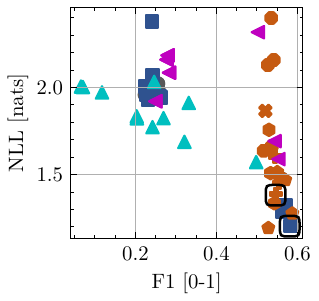}
        \caption{OOD ACC, NLL}
        \label{fig:baselines:bloodmnist:ood:f1_nll}
    \end{subfigure}
    \vspace{0.2cm}
    \caption{Comparison on ID and OOD test sets for BloodMNIST, with respect to \textbf{\textcolor{Gray}{Standard NN $\newmoon$}}, \textbf{\textcolor{Blue}{NN Ensemble $\blacksquare$}}, 
    \textbf{\textcolor{Orange}{SAE: 
    I/B: $N\geq 2, 2 \leq K < D$ \PlusMarker, 
    EE: $N=1, K \geq 2$ \OctagonMarker, 
    MIMMO: $N\geq 2, K=D$ \PentagonMarker, 
    MIMO: $N\geq 2, K=1$ \HexagonMarker, 
    SE NN: $N=1, K=1$ \TimesMarker}}, 
    \textbf{\textcolor{Cyan}{MCD $\blacktriangle$}}, 
    \textbf{\textcolor{Pink}{BE $\blacktriangleleft$}}.
    The black outlines denote the configurations compared in the text.}
    \label{fig:baselines:bloodmnist:id_ood}
\end{figure}

\begin{figure}
    \centering
    
    \begin{subfigure}[t]{0.22\textwidth}
        \centering
        \includegraphics[width=\textwidth]{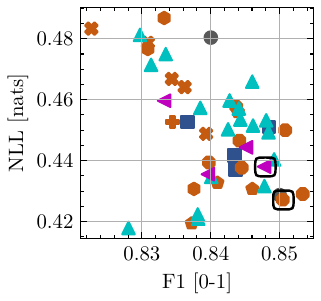}
        \caption{ID ACC, NLL}
        \label{fig:baselines:pneumoniamnist:id:f1_nll}
    \end{subfigure}
    \hfill
    \begin{subfigure}[t]{0.23\textwidth}
        \centering
        \includegraphics[width=\textwidth]{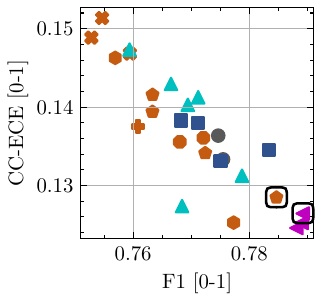}
        \caption{OOD ACC, CAL}
        \label{fig:baselines:pneumoniamnist:ood:f1_classconditionalcalibration}
    \end{subfigure}
    \hfill
    \begin{subfigure}[t]{0.22\textwidth}
        \centering
        \includegraphics[width=\textwidth]{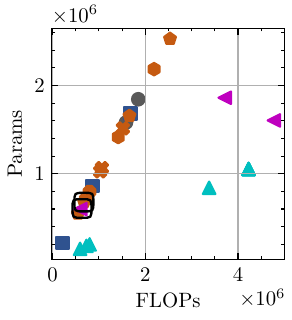}
        \caption{OOD FLOPs, Params}
        \label{fig:baselines:pneumoniamnist:ood:flops_params}
    \end{subfigure}
    \hfill
    \begin{subfigure}[t]{0.22\textwidth}
        \centering
        \includegraphics[width=\textwidth]{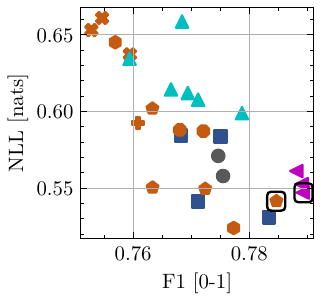}
        \caption{OOD ACC, NLL}
        \label{fig:baselines:pneumoniamnist:ood:f1_nll}
    \end{subfigure}
    \vspace{0.2cm}
    \caption{Comparison on ID and OOD test sets for PneumoniaMNIST, with respect to \textbf{\textcolor{Gray}{Standard NN $\newmoon$}}, \textbf{\textcolor{Blue}{NN Ensemble $\blacksquare$}}, 
    \textbf{\textcolor{Orange}{SAE: 
    I/B: $N\geq 2, 2 \leq K < D$ \PlusMarker, 
    EE: $N=1, K \geq 2$ \OctagonMarker, 
    MIMMO: $N\geq 2, K=D$ \PentagonMarker, 
    MIMO: $N\geq 2, K=1$ \HexagonMarker, 
    SE NN: $N=1, K=1$ \TimesMarker}}, 
    \textbf{\textcolor{Cyan}{MCD $\blacktriangle$}}, 
    \textbf{\textcolor{Pink}{BE $\blacktriangleleft$}}.
    The black outlines denote the configurations compared in the text.}
    \label{fig:baselines:pneumoniamnist:id_ood}
\end{figure}

\begin{figure}
    \centering
    \hfill
    \begin{subfigure}[t]{0.25\textwidth}
        \centering
        \includegraphics[width=\textwidth]{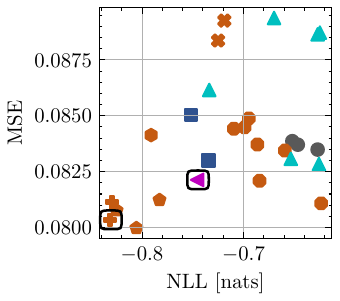}
        \caption{OOD NLL, MSE}
        \label{fig:baselines:retinamnist:ood:nll_mse}
    \end{subfigure}
    \hfill
    \begin{subfigure}[t]{0.23\textwidth}
        \centering
        \includegraphics[width=\textwidth]{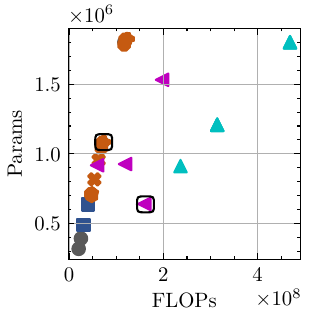}
        \caption{OOD FLOPs, Params}
        \label{fig:baselines:retinamnist:ood:flops_params}
    \end{subfigure}
    \hfill
    \vspace{0.2cm}
    \caption{Comparison on OOD test sets for RetinaMNIST, with respect to \textbf{\textcolor{Gray}{Standard NN $\newmoon$}}, \textbf{\textcolor{Blue}{NN Ensemble $\blacksquare$}}, 
    \textbf{\textcolor{Orange}{SAE: 
    I/B: $N\geq 2, 2 \leq K < D$ \PlusMarker, 
    EE: $N=1, K \geq 2$ \OctagonMarker, 
    MIMMO: $N\geq 2, K=D$ \PentagonMarker, 
    MIMO: $N\geq 2, K=1$ \HexagonMarker, 
    SE NN: $N=1, K=1$ \TimesMarker}}, 
    \textbf{\textcolor{Cyan}{MCD $\blacktriangle$}}, 
    \textbf{\textcolor{Pink}{BE $\blacktriangleleft$}}.
    The black outlines denote the configurations compared in the text.}
    \label{fig:baselines:rertinamnist:ood}
\end{figure}

In Figures~\ref{fig:baselines:tinyimagenet:id_ood},~\ref{fig:baselines:bloodmnist:id_ood},~\ref{fig:baselines:pneumoniamnist:id_ood}, and~\ref{fig:baselines:rertinamnist:ood} we provide the additional results for the baselines across all datasets with respect to NLL on ID test data and OOD test data.

For TinyImageNet and the ID test set in Figure~\ref{fig:baselines:tinyimagenet:id:accuracy_nll}, the \textcolor{Orange}{I/B \PlusMarker} configuration is better in NLL in comparison to the naive \textcolor{Blue}{ensemble $\blacksquare$} with $3.2\times$ fewer FLOPs and $1.5\times$ fewer parameters.
On OOD data in Figures~\ref{fig:baselines:tinyimagenet:ood:accuracy_calibration},~\ref{fig:baselines:tinyimagenet:ood:flops_params}, and~\ref{fig:baselines:tinyimagenet:ood:accuracy_nll}, the \textcolor{Orange}{I/B \PlusMarker} configuration is within the standard deviation of accuracy and with better mean ECE and NLL than the best \textcolor{Blue}{ensemble $\blacksquare$} while conserving the same FLOPs and parameters.

For BloodMNIST and the ID test set in Figure~\ref{fig:baselines:bloodmnist:id:f1_nll}, the \textcolor{Orange}{I/B \PlusMarker} configuration is comparable to the naive \textcolor{Blue}{ensemble $\blacksquare$} in NLL with $1.5\times$ fewer FLOPs and a comparable number of parameters.
On OOD data, seen in Figures~\ref{fig:baselines:bloodmnist:ood:f1_classconditionalcalibration},~\ref{fig:baselines:bloodmnist:ood:flops_params}, and~\ref{fig:baselines:bloodmnist:ood:f1_nll}, the SAE configurations all achieve high F1 score in comparison to the baselines, and a \textcolor{Orange}{I/B \PlusMarker} configuration is within 2\% of CC-ECE or F1 of the best \textcolor{Blue}{ensemble $\blacksquare$} with $2.7\times$ fewer FLOPs and $3.7\times$ fewer parameters.

For PneumoniaMNIST and the ID test set in Figure~\ref{fig:baselines:pneumoniamnist:id:f1_nll}, the \textcolor{Orange}{EE \OctagonMarker} configuration achieves one of the lowest NLLs with approximately half the FLOPs and parameters of the best \textcolor{Pink}{BE $\blacktriangleleft$}.
On OOD data, in Figures~\ref{fig:baselines:pneumoniamnist:ood:f1_classconditionalcalibration},~\ref{fig:baselines:pneumoniamnist:ood:flops_params}, and~\ref{fig:baselines:pneumoniamnist:ood:f1_nll}, the \textcolor{Orange}{MIMO \PentagonMarker} configuration is within 1\% of the best \textcolor{Pink}{BE $\blacktriangleleft$} with similar FLOPs and parameters.

Lastly, for RetinaMNIST and the OOD test set in Figures~\ref{fig:baselines:retinamnist:ood:nll_mse} and~\ref{fig:baselines:retinamnist:ood:flops_params}, the \textcolor{Orange}{I/B \PlusMarker} configuration is the best-performing method with approximately $2\times$ fewer FLOPs than the second best \textcolor{Pink}{BE $\blacktriangleleft$}.

\subsection{Overall Baseline Performance Ranks}\label{sec:appendix:baseline:overall_performance}

In Tables~\ref{tab:overall_performance:id:classification},~\ref{tab:overall_performance:ood:classification},~\ref{tab:overall_performance:id:regression} and~\ref{tab:overall_performance:ood:regression} we collected all the Pareto candidates for a method and a dataset, calculated the mean performance of the candidates across all metrics and then ranked the methods with respect to each other and their mean per-metric performance. 
Then we averaged the ranks across all the datasets per classification and regression. 
The best mean ranking, where lower ranks are better, is highlighted in bold. 
As it can be seen in the Tables, the methods generalised under SAE rank top or second across all datasets and metrics, demonstrating the robustness of SAE to different datasets and environments whether considering classification or regression tasks or ID or OOD data.
In case of FLOPs and Params, one needs to be careful with interpreting the results, as the methods with the lowest FLOPs and Params are not necessarily the best-performing methods in terms of the algorithmic metrics.

\subsection{Hyperparameter Optimisation}\label{sec:appendix:baseline:hpo}

\begin{figure}[t]
    \centering
    
    \begin{subfigure}[t]{0.24\textwidth}
        \centering
        \includegraphics[width=\textwidth]{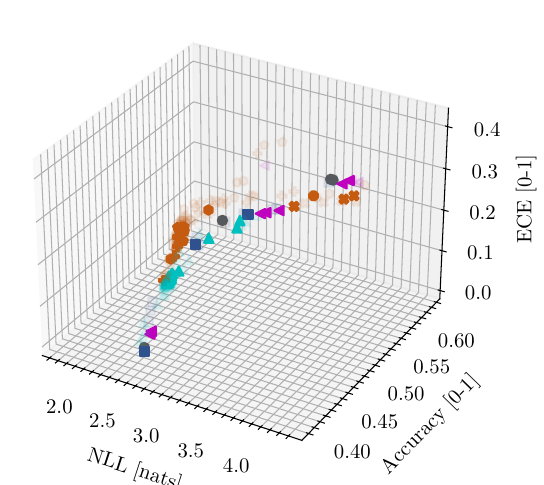}
        \caption{TinyImageNet ID.}
        \label{fig:baselines:tinyimagenet:id:hpo}
    \end{subfigure}
    \hfill
    \begin{subfigure}[t]{0.23\textwidth}
        \centering
        \includegraphics[width=\textwidth]{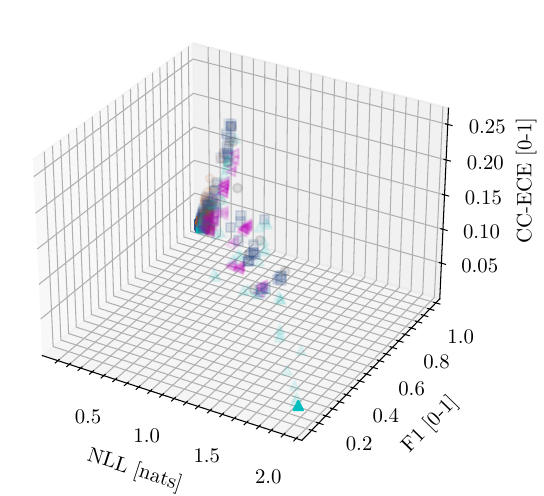}  
        \caption{BloodMNIST ID.}
        \label{fig:baselines:bloodmnist:id:hpo}
    \end{subfigure}
    \hfill
    \begin{subfigure}[t]{0.23\textwidth}
        \centering
        \includegraphics[width=\textwidth]{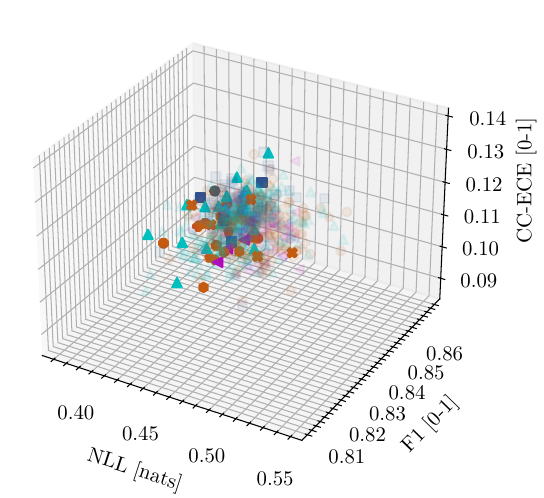}
        \caption{PneumoniaMNIST ID.}
        \label{fig:baselines:pneumoniamnist:id:hpo}
    \end{subfigure}
    \hfill
    \begin{subfigure}[t]{0.24\textwidth}
        \centering
        \includegraphics[width=\textwidth]{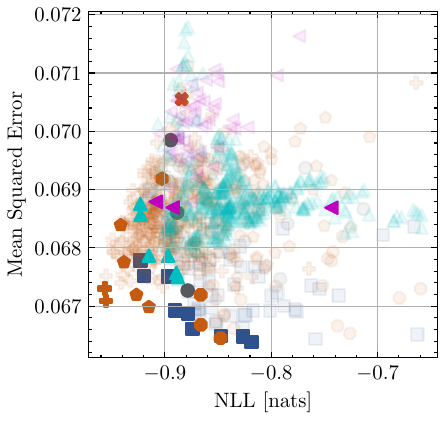}
        \caption{RetinaMNIST ID.}
        \label{fig:baselines:retinamnist:id:hpo}
    \end{subfigure}
    
    \begin{subfigure}[t]{0.23\textwidth}
        \centering
        \includegraphics[width=\textwidth]{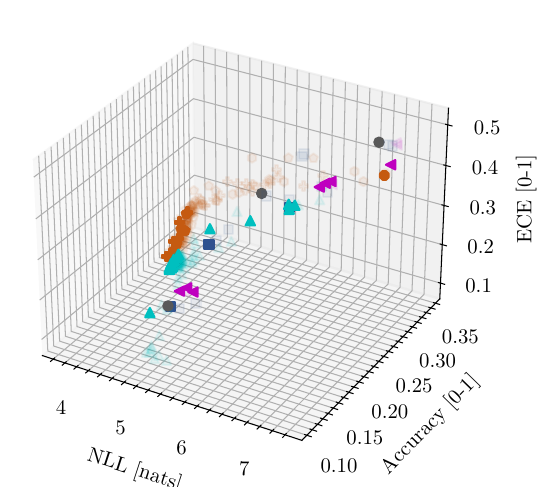}
        \caption{TinyImageNet OOD.}
        \label{fig:baselines:tinyimagenet:ood:hpo}
    \end{subfigure}
    \hfill
    \begin{subfigure}[t]{0.23\textwidth}
        \centering
        \includegraphics[width=\textwidth]{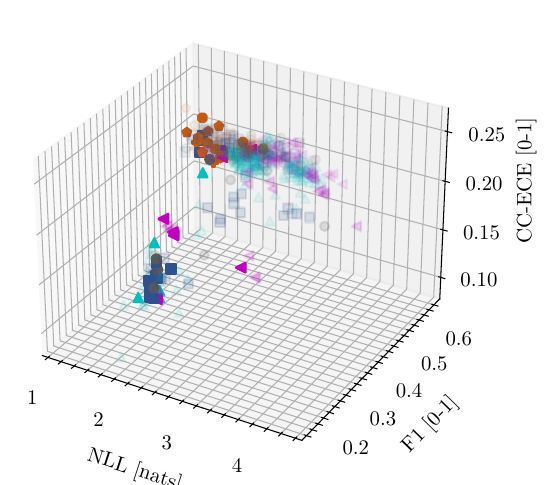}
        \caption{BloodMNIST OOD.}
        \label{fig:baselines:bloodmnist:ood:hpo}
    \end{subfigure}
    \hfill
    \begin{subfigure}[t]{0.23\textwidth}
        \centering
        \includegraphics[width=\textwidth]{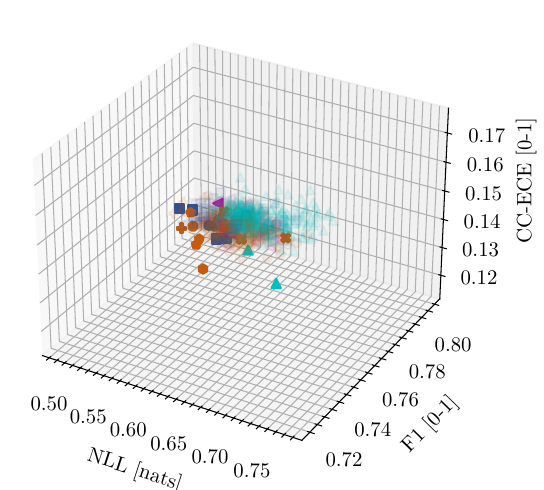}
        \caption{PneumoniaMNIST OOD.}
        \label{fig:baselines:pneumoniamnist:ood:hpo}
    \end{subfigure}
    \hfill
    \begin{subfigure}[t]{0.24\textwidth}
        \centering
        \includegraphics[width=\textwidth]{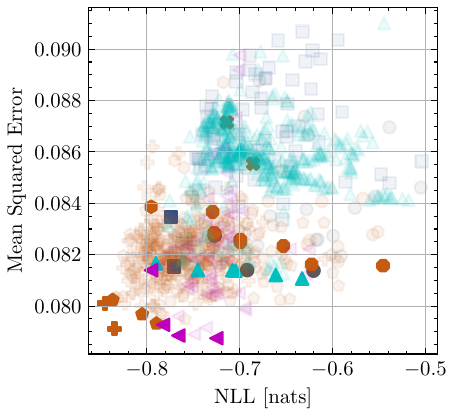}
        \caption{RetinaMNIST OOD.}
        \label{fig:baselines:retinamnist:ood:hpo}
    \end{subfigure}
    \vspace{0.2cm}
    \caption{Comparison on TinyImageNet, BloodMNIST, PneumoniaMNIST, and RetinaMNIST across ID (upper row) and OOD (lower row) datasets, with respect to \textbf{\textcolor{Gray}{Standard NN $\newmoon$}}, \textbf{\textcolor{Blue}{NN Ensemble $\blacksquare$}}, 
    \textbf{\textcolor{Orange}{SAE: 
    I/B: $N\geq 2, 2 \leq K < D$ \PlusMarker, 
    EE: $N=1, K \geq 2$ \OctagonMarker, 
    MIMMO: $N\geq 2, K=D$ \PentagonMarker, 
    MIMO: $N\geq 2, K=1$ \HexagonMarker, 
    SE NN: $N=1, K=1$ \TimesMarker}}, 
    \textbf{\textcolor{Cyan}{MCD $\blacktriangle$}}, 
    \textbf{\textcolor{Pink}{BE $\blacktriangleleft$}}.
    The plots contain all the configurations tried by hyperparameter optimisation.
    The high opacity configurations indicate the configurations used for the final evaluation over repeated seeds.
    }
    \label{fig:baseline:id_ood:hpo}
\end{figure}

In Figure~\ref{fig:baseline:id_ood:hpo}, we provide the hyperparameter optimisation (HPO) results for the baselines across all datasets.
The plots contain all the configurations tried by HPO, with high-opacity configurations indicating the configurations used for the final evaluation over repeated seeds.
The configurations were decided by looking at their empirical Pareto optimality across all the methods and then for each method individually across all the datasets.
These results demonstrate the HPO trends and algorithmic performance range that the baselines and SAE can achieve.
As the Figure shows, SAE tends to cluster around the Pareto front, demonstrating the adaptability of SAE to different search spaces and the robustness of the HPO process.
At the same time, this enables the practitioner to choose multiple configurations for SAE, which can be used to trade-off between the number of FLOPs, the number of parameters, and the algorithmic performance.

\subsection{Numerical Results}\label{sec:appendix:baseline:numerical_results}

In Tables~\ref{tab:tinyimagenet:id},~\ref{tab:tinyimagenet:ood},~\ref{tab:bloodmnist:id},~\ref{tab:bloodmnist:ood},~\ref{tab:pneumoniamnist:id},~\ref{tab:pneumoniamnist:ood},~\ref{tab:retinamnist:id}, and~\ref{tab:retinamnist:ood}, we provide the numerical results for the baselines across all datasets.
As in the main paper, we divide the SAE approach into the following categories: I/B, EE, MIMMO, MIMO, and SE NN.
We specify how many configurations were selected for the repeated seed evaluation.
We specify the configurations for each category where the first row of each method denotes the best performance on the first metric, the second row indicates the best performance on the second metric, and the third row represents the best performance on the third metric.
The \textbf{bold} font denotes the best performance across all the methods, and the \textit{italic} and \underline{underlined} font denote the configurations compared in the text.

\section{Supplementary for Changing K and N}\label{sec:appendix:changing_k_n}

\begin{figure}
    \centering
    \begin{subfigure}[t]{0.32\textwidth}
    \centering
    \includegraphics[width=\textwidth]{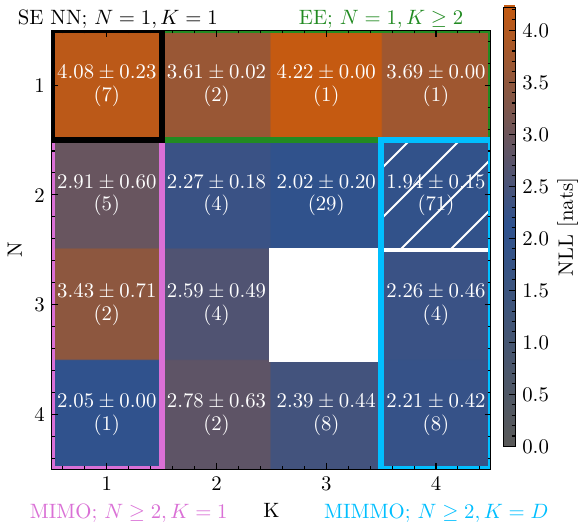}
    \caption{NLL on TinyImageNet ID.}
    \label{fig:changing_k_n:tinyimagenet:nll:id}
    \end{subfigure}
    \\
    \begin{subfigure}[t]{0.32\textwidth}
    \centering    
    \includegraphics[width=\textwidth]{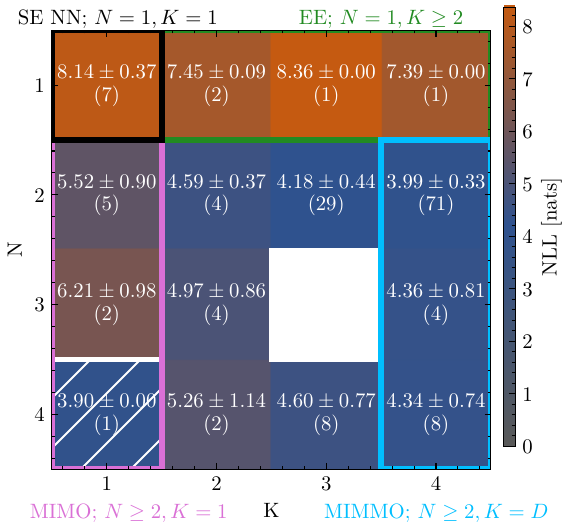}
    \caption{NLL on TinyImageNet OOD.}
    \label{fig:changing_k_n:tinyimagenet:nll:ood}
    \end{subfigure}
    \hfill
    \begin{subfigure}[t]{0.32\textwidth}
    \centering
    \includegraphics[width=\textwidth]{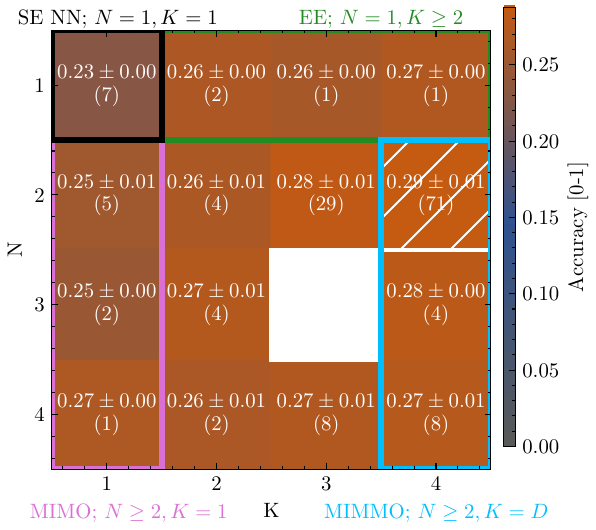}
    \caption{Accuracy on TinyImageNet OOD.}
    \label{fig:changing_k_n:tinyimagenet:accuracy:ood}
    \end{subfigure}
    \hfill
    \begin{subfigure}[t]{0.32\textwidth}
    \centering
    \includegraphics[width=\textwidth]{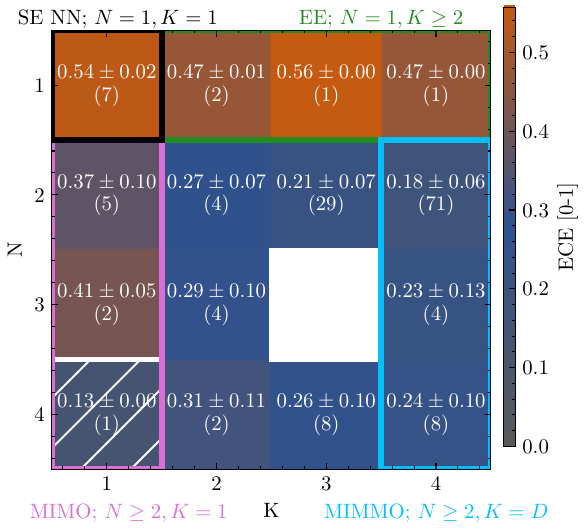}
    \caption{ECE on TinyImageNet OOD.}
    \label{fig:changing_k_n:tinyimagenet:ece:ood}
    \end{subfigure}
    \vspace{0.2cm}
    \caption{Varying $N,K$ on TinyImageNet ID and OOD test sets.
    The upper number is the average performance over $N$, $K$ combinations. 
    The number in brackets is the number of sampled configurations by HPO. 
    White box means no configurations sampled for that $N$, $K$. 
    Pattern signals best average performance.
    The coloured outlines signal the special cases for the generalised methods.}
    \label{fig:changing_k_n:tinyimagenet:nll_accuracy_ece}
\end{figure}

\begin{figure}
    \centering
    \begin{subfigure}[t]{0.32\textwidth}
    \centering    
    \includegraphics[width=\textwidth]{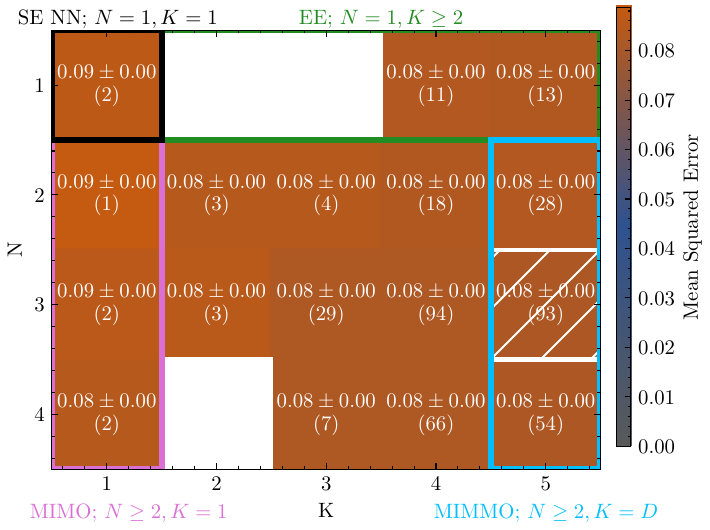}
    \caption{MSE on RetinaMNIST OOD.}
    \label{fig:changing_k_n:retinamnist:mse:ood}
    \end{subfigure}
    \hfill
    \begin{subfigure}[t]{0.32\textwidth}
    \centering
    \includegraphics[width=\textwidth]{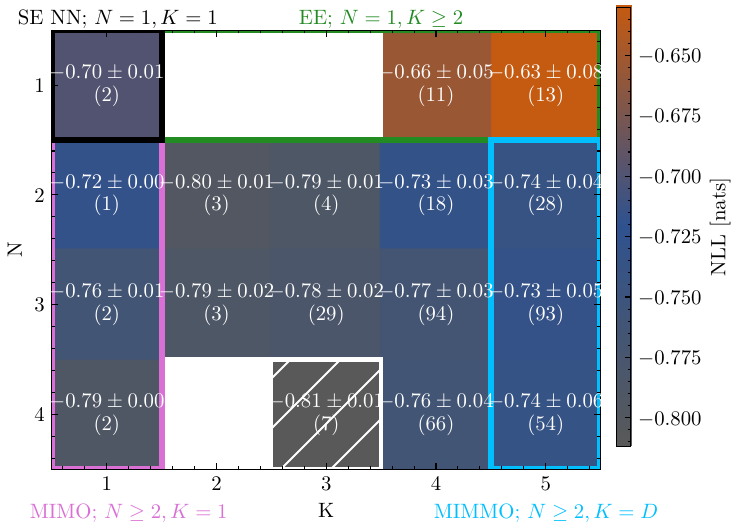}
    \caption{NLL on RetinaMNIST OOD.}
    \label{fig:changing_k_n:retinamnist:nll:ood}
    \end{subfigure}
    \vspace{0.2cm}
    \caption{Varying $N,K$ on RetinaMNIST OOD test sets.
    The upper number is the average performance over $N$, $K$ combinations. 
    The number in brackets is the number of sampled configurations by HPO. 
    White box means no configurations sampled for that $N$, $K$. 
    Pattern signals best average performance.
    The coloured outlines signal the special cases for the generalised methods.}
    \label{fig:changing_k_n:retinamnist:ood}
\end{figure}

\begin{figure}
    \centering
    \begin{subfigure}[t]{0.32\textwidth}
    \centering
    \includegraphics[width=\textwidth]{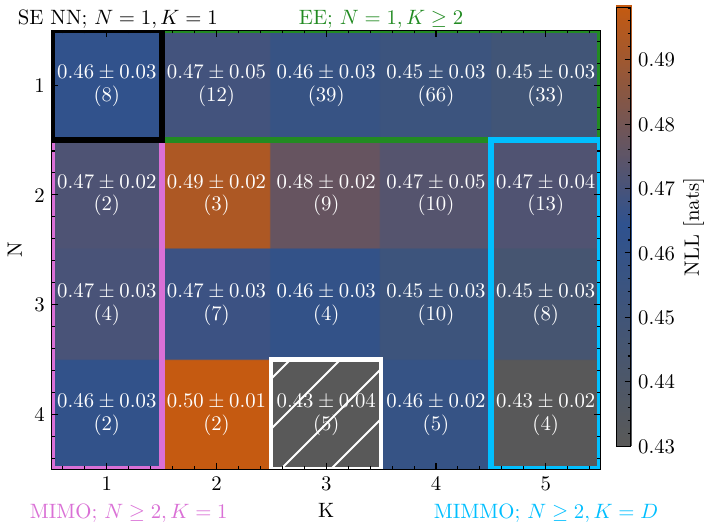}
    \caption{NLL on PneumoniaMNIST ID.}
    \label{fig:changing_k_n:pneumoniamnist:nll:id}
    \end{subfigure}
    \\
    \begin{subfigure}[t]{0.32\textwidth}
    \centering    
    \includegraphics[width=\textwidth]{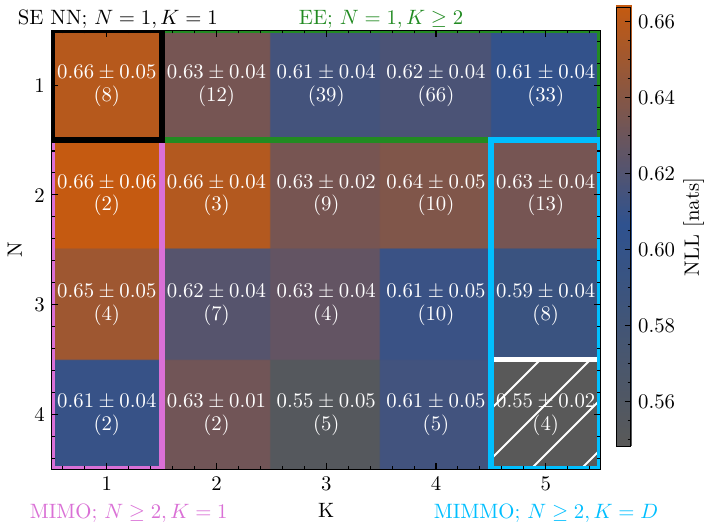}
    \caption{NLL on PneumoniaMNIST OOD.}
    \label{fig:changing_k_n:pneumoniamnist:nll:ood}
    \end{subfigure}
    \hfill
    \begin{subfigure}[t]{0.32\textwidth}
    \centering
    \includegraphics[width=\textwidth]{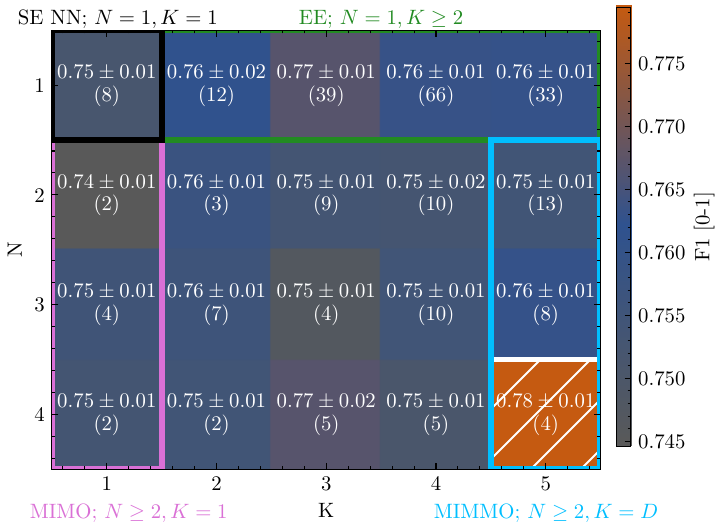}
    \caption{F1 on PneumoniaMNIST OOD.}
    \label{fig:changing_k_n:pneumoniamnist:f1:ood}
    \end{subfigure}
    \hfill
    \begin{subfigure}[t]{0.32\textwidth}
    \centering
    \includegraphics[width=\textwidth]{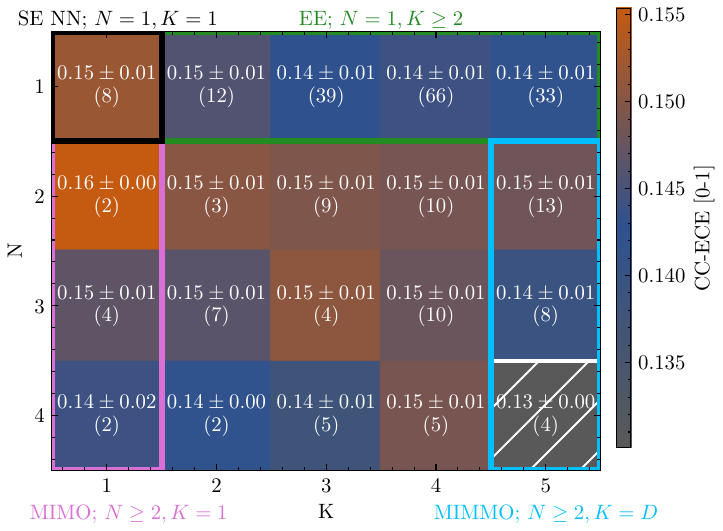}
    \caption{ECE on PneumoniaMNIST OOD.}
    \label{fig:changing_k_n:pneumoniamnist:ccece:ood}
    \end{subfigure}
    \vspace{0.2cm}
    \caption{Varying $N,K$ on PneumoniaMNIST ID and OOD test sets.
    The upper number is the average performance over $N$, $K$ combinations. 
    The number in brackets is the number of sampled configurations by HPO. 
    White box means no configurations sampled for that $N$, $K$. 
    Pattern signals best average performance.
    The coloured outlines signal the special cases for the generalised methods.}
    \label{fig:changing_k_n:pneumoniamnist:nll_f1_ece}
\end{figure}

\begin{figure}
    \centering
    \begin{subfigure}[t]{0.32\textwidth}
    \centering
    \includegraphics[width=\textwidth]{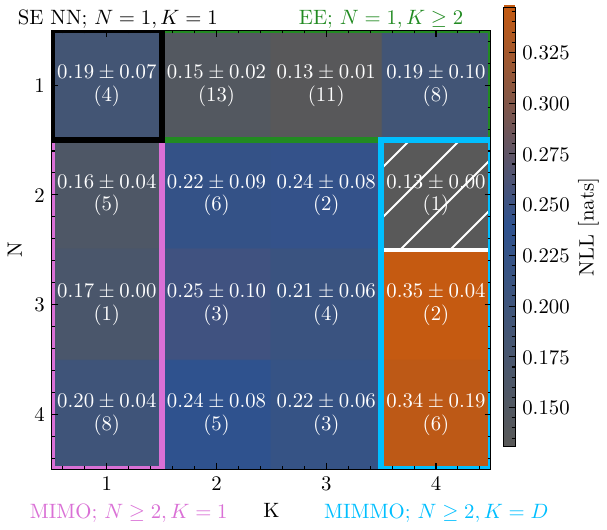}
    \caption{NLL on BloodMNIST ID.}
    \label{fig:changing_k_n:bloodmnist:nll:id}
    \end{subfigure}
    \\
    \begin{subfigure}[t]{0.32\textwidth}
    \centering    
    \includegraphics[width=\textwidth]{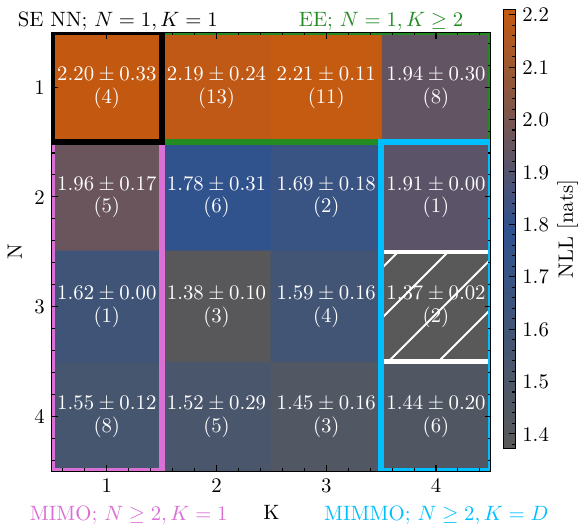}
    \caption{NLL on BloodMNIST OOD.}
    \label{fig:changing_k_n:bloodmnist:nll:ood}
    \end{subfigure}
    \hfill
    \begin{subfigure}[t]{0.32\textwidth}
    \centering
    \includegraphics[width=\textwidth]{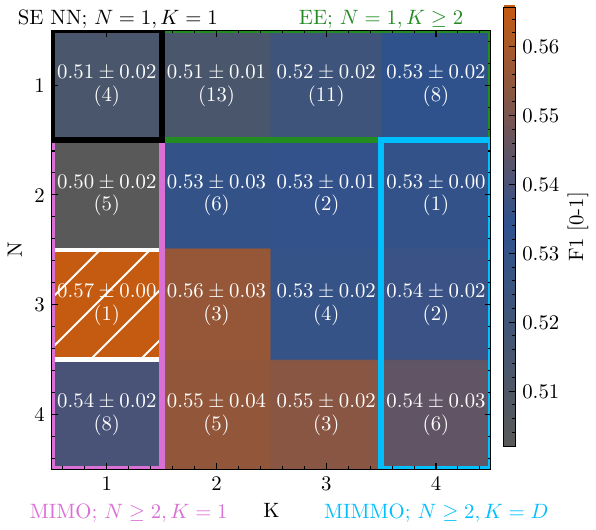}
    \caption{F1 on BloodMNIST OOD.}
    \label{fig:changing_k_n:bloodmnist:f1:ood}
    \end{subfigure}
    \hfill
    \begin{subfigure}[t]{0.32\textwidth}
    \centering
    \includegraphics[width=\textwidth]{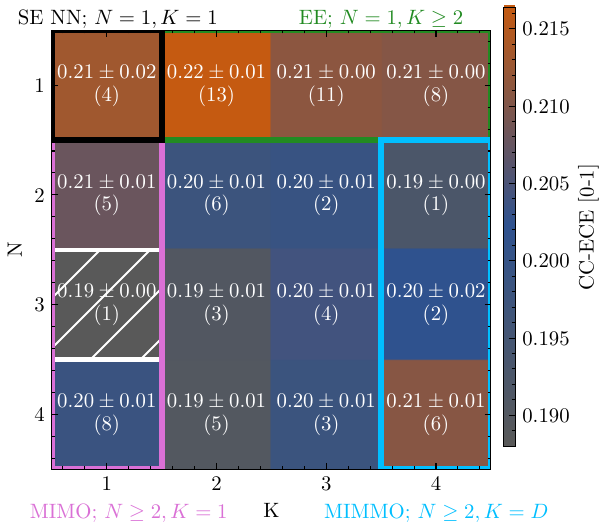}
    \caption{ECE on BloodMNIST OOD.}
    \label{fig:changing_k_n:bloodmnist:ccece:ood}
    \end{subfigure}
    \vspace{0.2cm}
    \caption{Varying $N,K$ on BloodMNIST ID and OOD test sets.
    The upper number is the average performance over $N$, $K$ combinations. 
    The number in brackets is the number of sampled configurations by HPO. 
    White box means no configurations sampled for that $N$, $K$. 
    Pattern signals best average performance.
    The coloured outlines signal the special cases for the generalised methods.}
    \label{fig:changing_k_n:bloodmnist:nll_f1_ece}
\end{figure}

In Figures~\ref{fig:changing_k_n:tinyimagenet:nll_accuracy_ece},~\ref{fig:changing_k_n:retinamnist:ood},~\ref{fig:changing_k_n:pneumoniamnist:nll_f1_ece}
and~\ref{fig:changing_k_n:bloodmnist:nll_f1_ece}  we show the Varying $N,K$ on TinyImageNet, RetinaMNIST, PneumoniaMNIST and BloodMNIST datasets concerning different metrics.
As discussed in the main part of the paper, there might be different suitable configurations for various metrics. 
We would point the reader's attention to the OOD results across all datasets and the ID results where we have shown that the $N$ and $K$ are different. 
This further indicates the necessity of SAE and the proposed search space and optimisation strategy, since the best configurations are not consistent across the same metric, dataset and backbone architecture.

\section{Supplementary for Depth Preference}\label{sec:appendix:depth}

\begin{figure}
    \centering
    \begin{subfigure}[t]{0.24\textwidth}
    \centering
    \includegraphics[width=\textwidth]{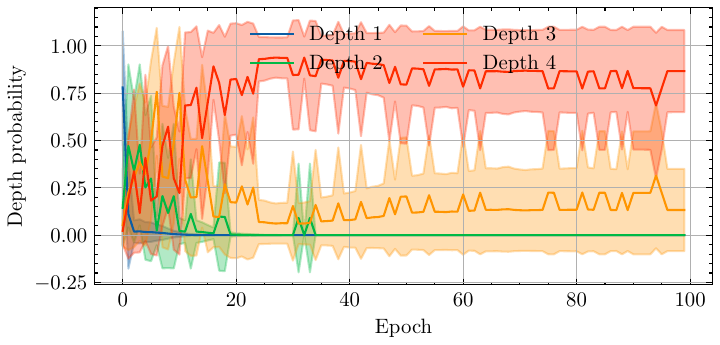}
    \caption{$N=1$.}
    \label{fig:depth:tinyimagenet:member_1}
    \end{subfigure}
    \hfill
    \begin{subfigure}[t]{0.24\textwidth}
    \centering
    \includegraphics[width=\textwidth]{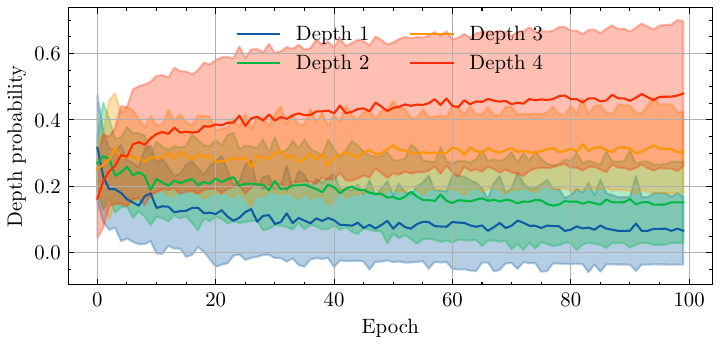}
    \caption{$N=2$.}
    \label{fig:depth:tinyimagenet:member_2}
    \end{subfigure}
    \hfill
    \begin{subfigure}[t]{0.24\textwidth}
    \centering
    \includegraphics[width=\textwidth]{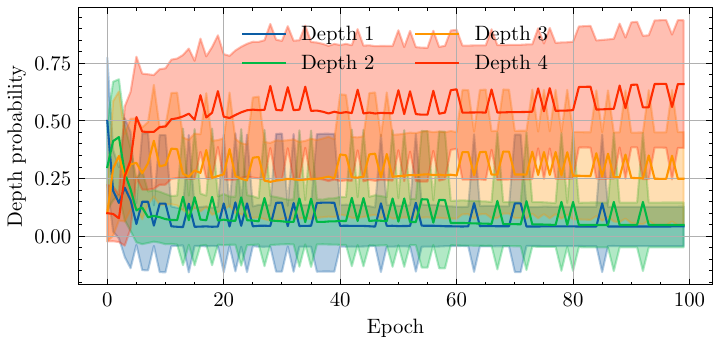}
    \caption{$N=3$.}
    \label{fig:depth:tinyimagenet:member_3}
    \end{subfigure}
    \hfill
    \begin{subfigure}[t]{0.24\textwidth}
    \centering
    \includegraphics[width=\textwidth]{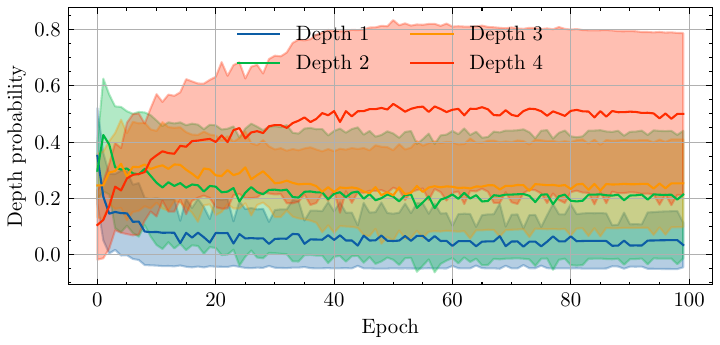}
    \caption{$N=4$.}
    \label{fig:depth:tinyimagenet:member_4}
    \end{subfigure}
    \begin{subfigure}[t]{0.24\textwidth}
    \centering
    \includegraphics[width=\textwidth]{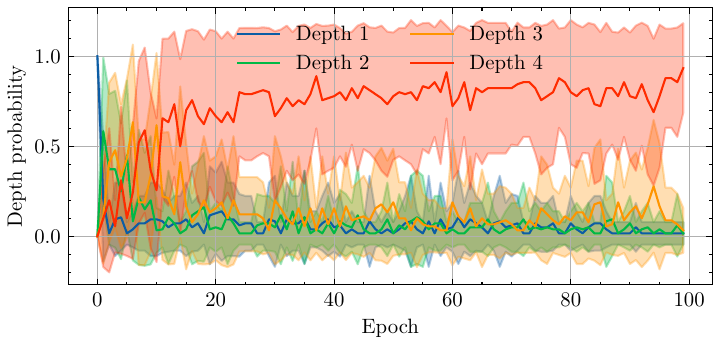}
    \caption{$K=1$.}
    \label{fig:depth:tinyimagenet:k_1}
    \end{subfigure}
    \hfill
    \begin{subfigure}[t]{0.24\textwidth}
    \centering
    \includegraphics[width=\textwidth]{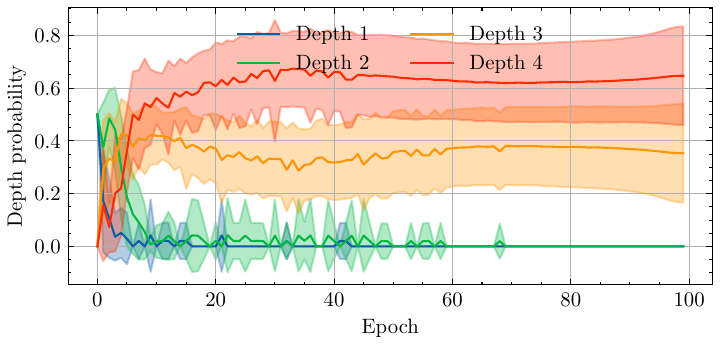}
    \caption{$K=2$.}
    \label{fig:depth:tinyimagenet:k_2}
    \end{subfigure}
    \hfill
    \begin{subfigure}[t]{0.24\textwidth}
    \centering
    \includegraphics[width=\textwidth]{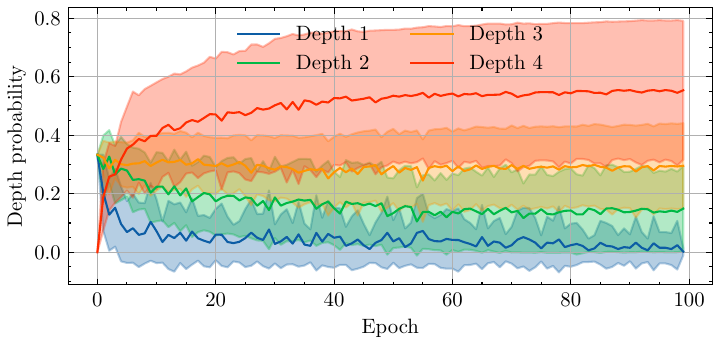}
    \caption{$K=3$.}
    \label{fig:depth:tinyimagenet:k_3}
    \end{subfigure}
    \hfill
    \begin{subfigure}[t]{0.24\textwidth}
    \centering
    \includegraphics[width=\textwidth]{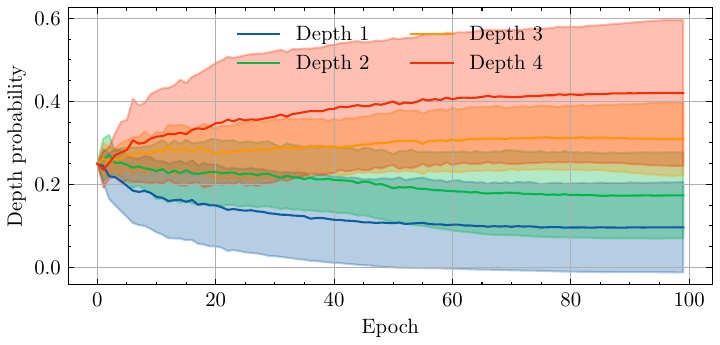}
    \caption{$K=D$.}
    \label{fig:depth:tinyimagenet:k_4}
    \end{subfigure}
    \vspace{0.2cm}
    \caption{Depth preference during training when averaging over different $N$ and $K$ for TinyImageNet.
    The lines denote the mean trend, and the shaded regions denote the standard deviation across configurations.}
    \label{fig:depth:tinyimagenet:members}
\end{figure}

\begin{figure}
    \centering
    \begin{subfigure}[t]{0.24\textwidth}
    \centering
    \includegraphics[width=\textwidth]{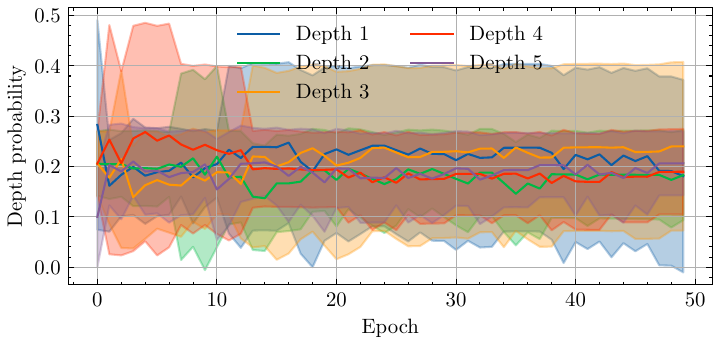}
    \caption{$N=1$.}
    \label{fig:depth:retinamnist:member_1}
    \end{subfigure}
    \hfill
    \begin{subfigure}[t]{0.24\textwidth}
    \centering
    \includegraphics[width=\textwidth]{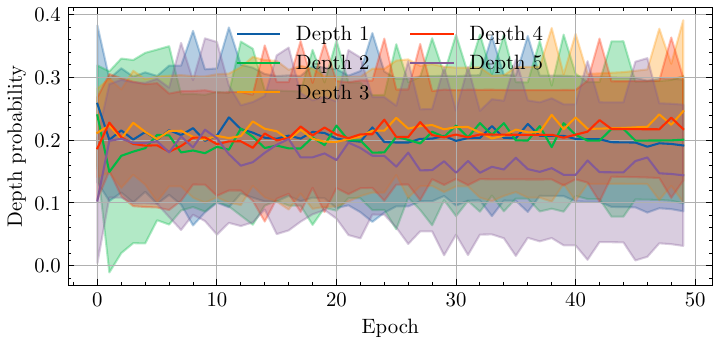}
    \caption{$N=2$.}
    \label{fig:depth:retinamnist:member_2}
    \end{subfigure}
    \hfill
    \begin{subfigure}[t]{0.24\textwidth}
    \centering
    \includegraphics[width=\textwidth]{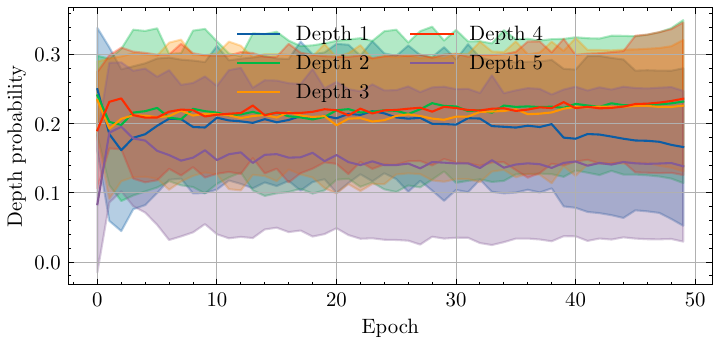}
    \caption{$N=3$.}
    \label{fig:depth:retinamnist:member_3}
    \end{subfigure}
    \hfill
    \begin{subfigure}[t]{0.24\textwidth}
    \centering
    \includegraphics[width=\textwidth]{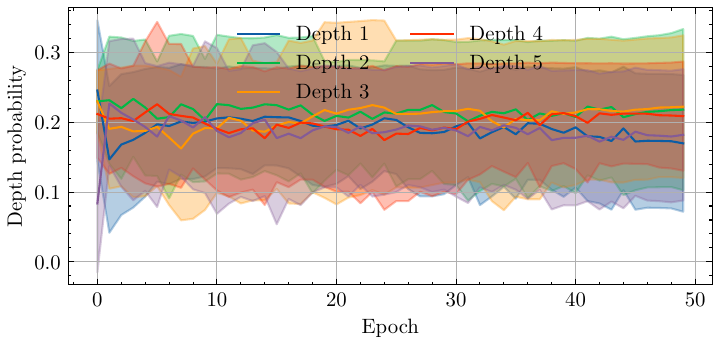}
    \caption{$N=4$.}
    \label{fig:depth:retinamnist:member_4}
    \end{subfigure}
    \newline
    \begin{subfigure}[t]{0.19\textwidth}
    \centering
    \includegraphics[width=\textwidth]{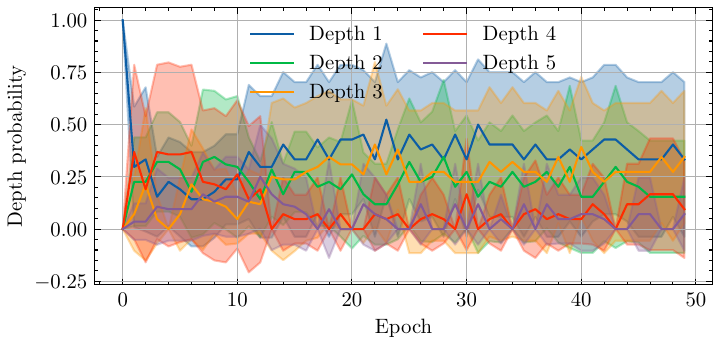}
    \caption{$K=1$.}
    \label{fig:depth:retinamnist:k_1}
    \end{subfigure}
    \hfill
    \begin{subfigure}[t]{0.19\textwidth}
    \centering
    \includegraphics[width=\textwidth]{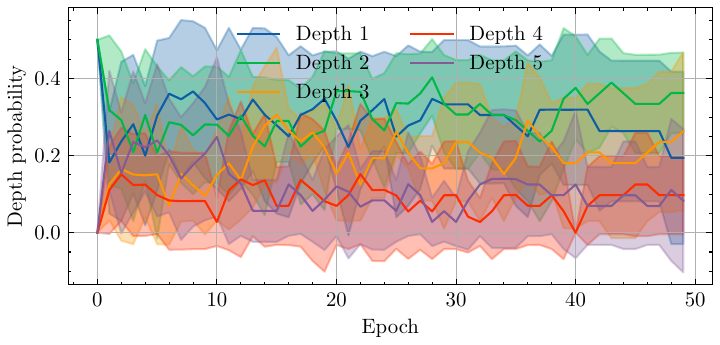}
    \caption{$K=2$.}
    \label{fig:depth:retinamnist:k_2}
    \end{subfigure}
    \hfill
    \begin{subfigure}[t]{0.19\textwidth}
    \centering
    \includegraphics[width=\textwidth]{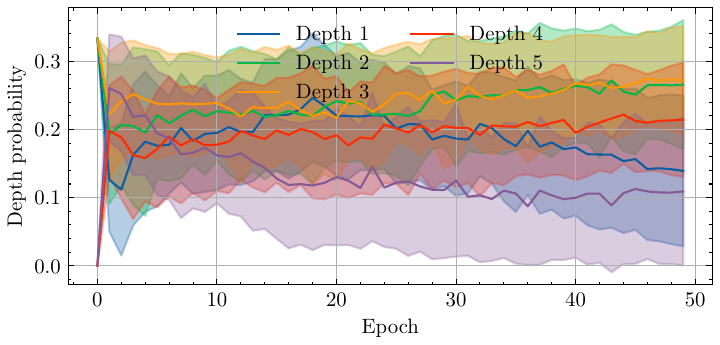}
    \caption{$K=3$.}
    \label{fig:depth:retinamnist:k_3}
    \end{subfigure}
    \hfill
    \begin{subfigure}[t]{0.19\textwidth}
    \centering
    \includegraphics[width=\textwidth]{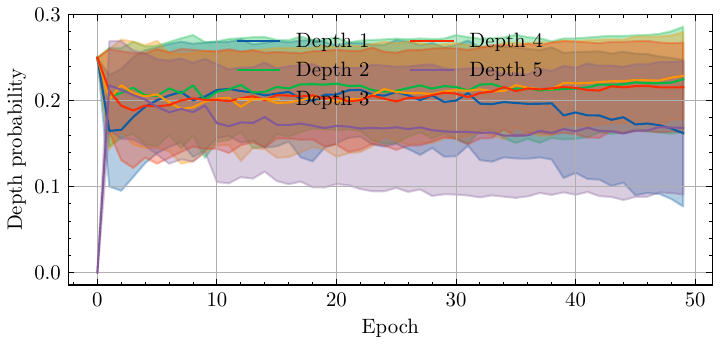}
    \caption{$K=4$.}
    \label{fig:depth:retinamnist:k_4}
    \end{subfigure}
    \begin{subfigure}[t]{0.19\textwidth}
    \centering
    \includegraphics[width=\textwidth]{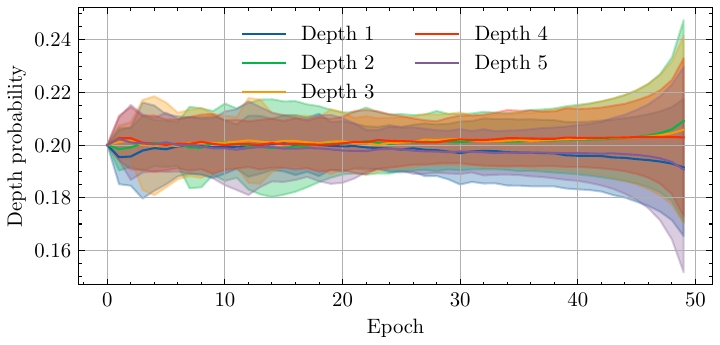}
    \caption{$K=D$.}
    \label{fig:depth:retinamnist:k_5}
    \end{subfigure}
    \vspace{0.2cm}
    \caption{Depth preference during training when averaging over different $N$ and $K$ for RetinaMNIST.
    The lines denote the mean trend, and the shaded regions denote the standard deviation across configurations.}
    \label{fig:depth:retinamnist:members}
\end{figure}

\begin{figure}
    \centering
    \begin{subfigure}[t]{0.24\textwidth}
    \centering
    \includegraphics[width=\textwidth]{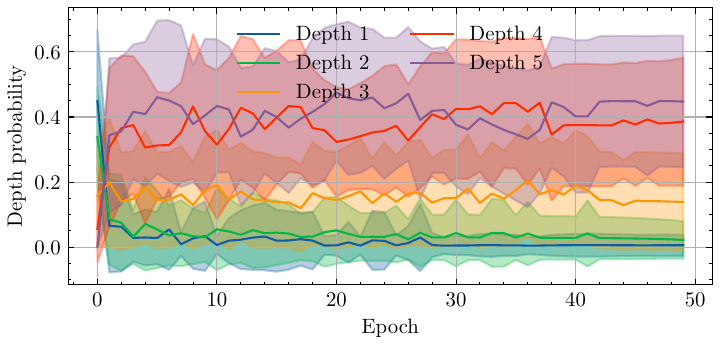}
    \caption{$N=1$.}
    \label{fig:depth:bloodmnist:member_1}
    \end{subfigure}
    \hfill
    \begin{subfigure}[t]{0.24\textwidth}
    \centering
    \includegraphics[width=\textwidth]{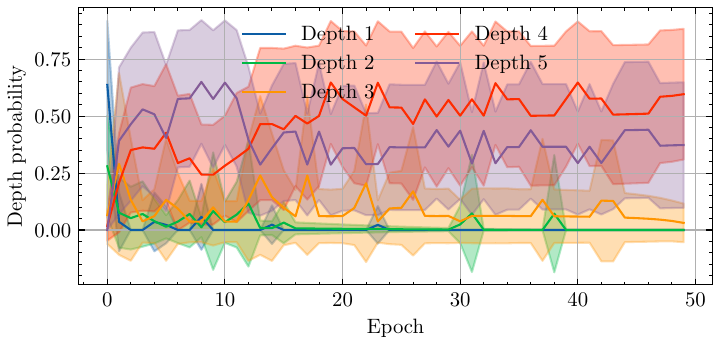}
    \caption{$N=2$.}
    \label{fig:depth:bloodmnist:member_2}
    \end{subfigure}
    \hfill
    \begin{subfigure}[t]{0.24\textwidth}
    \centering
    \includegraphics[width=\textwidth]{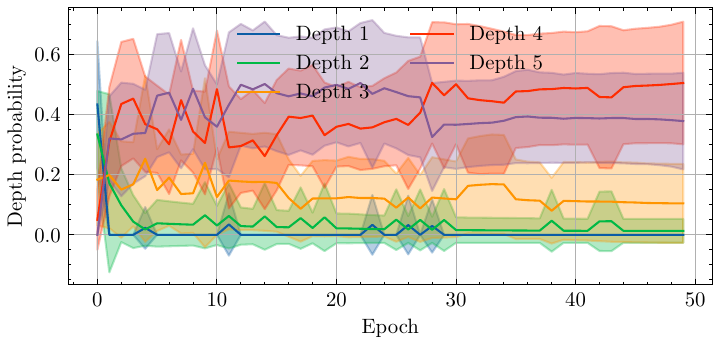}
    \caption{$N=3$.}
    \label{fig:depth:bloodmnist:member_3}
    \end{subfigure}
    \hfill
    \begin{subfigure}[t]{0.24\textwidth}
    \centering
    \includegraphics[width=\textwidth]{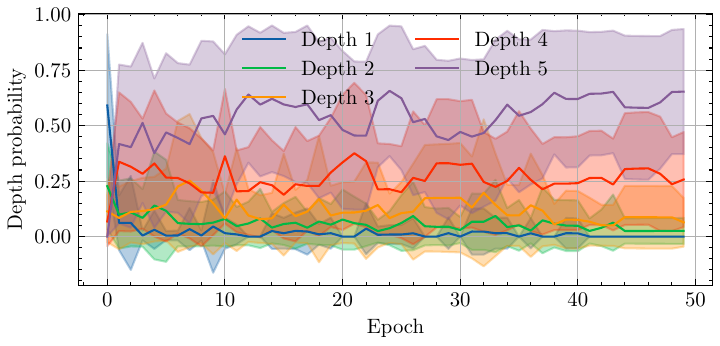}
    \caption{$N=4$.}
    \label{fig:depth:bloodmnist:member_4}
    \end{subfigure}
    \newline
    \begin{subfigure}[t]{0.24\textwidth}
    \centering
    \includegraphics[width=\textwidth]{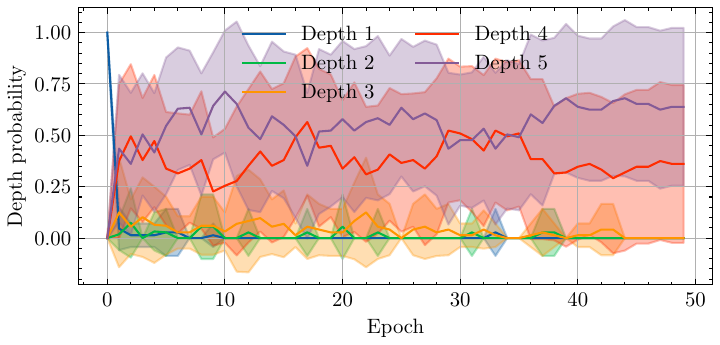}
    \caption{$K=1$.}
    \label{fig:depth:bloodmnist:k_1}
    \end{subfigure}
    \hfill
    \begin{subfigure}[t]{0.24\textwidth}
    \centering
    \includegraphics[width=\textwidth]{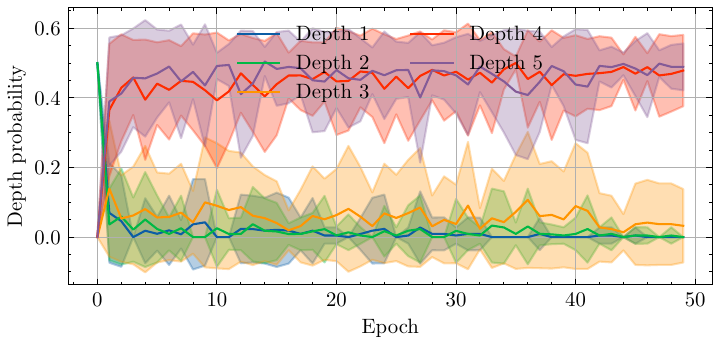}
    \caption{$K=2$.}
    \label{fig:depth:bloodmnist:k_2}
    \end{subfigure}
    \hfill
    \begin{subfigure}[t]{0.24\textwidth}
    \centering
    \includegraphics[width=\textwidth]{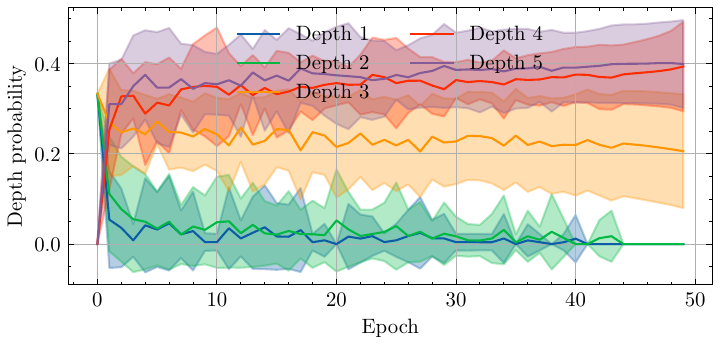}
    \caption{$K=3$.}
    \label{fig:depth:bloodmnist:k_3}
    \end{subfigure}
    \hfill
    \begin{subfigure}[t]{0.24\textwidth}
    \centering
    \includegraphics[width=\textwidth]{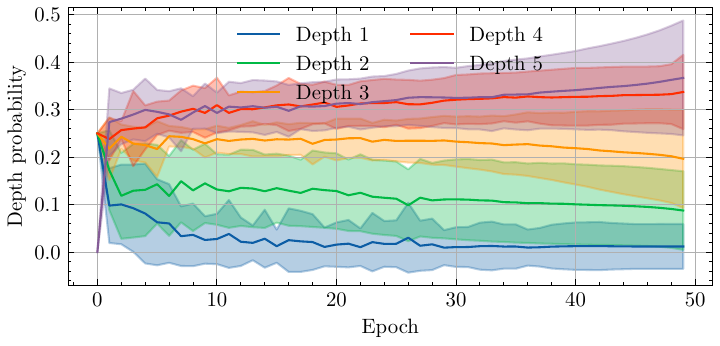}
    \caption{$K=D$.}
    \label{fig:depth:bloodmnist:k_4}
    \end{subfigure}
    \vspace{0.2cm}
    \caption{Depth preference during training when averaging over different $N$ and $K$ for BloodMNIST.
    The lines denote the mean trend, and the shaded regions denote the standard deviation across configurations.}
    \label{fig:depth:bloodmnist}
\end{figure}

\begin{figure}
    \centering
    \begin{subfigure}[t]{0.24\textwidth}
    \centering
    \includegraphics[width=\textwidth]{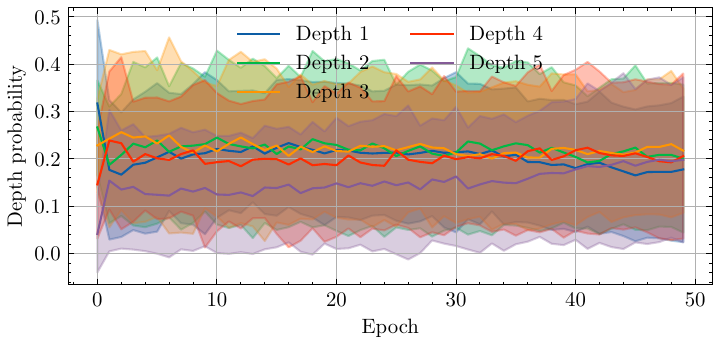}
    \caption{$N=1$.}
    \label{fig:depth:pneumoniamnist:member_1}
    \end{subfigure}
    \hfill
    \begin{subfigure}[t]{0.24\textwidth}
    \centering
    \includegraphics[width=\textwidth]{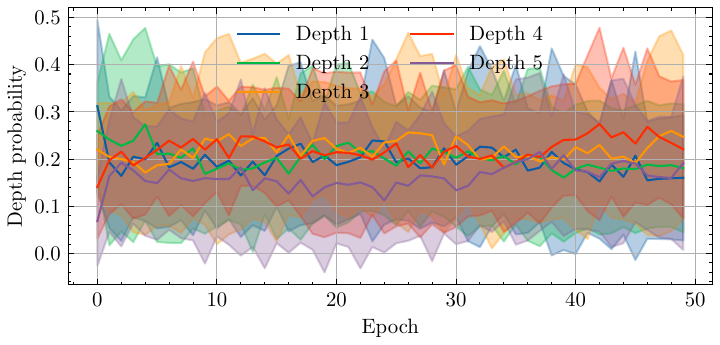}
    \caption{$N=2$.}
    \label{fig:depth:pneumoniamnist:member_2}
    \end{subfigure}
    \hfill
    \begin{subfigure}[t]{0.24\textwidth}
    \centering
    \includegraphics[width=\textwidth]{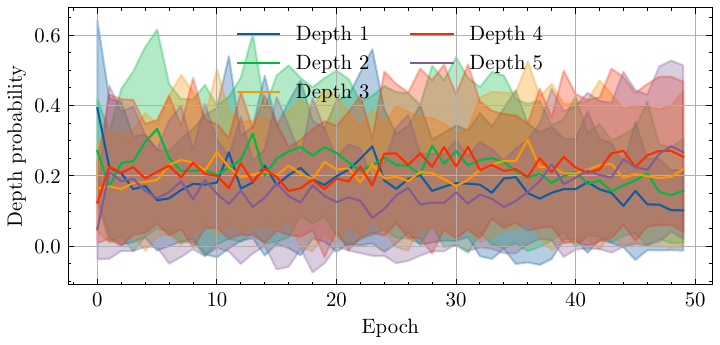}
    \caption{$N=3$.}
    \label{fig:depth:pneumoniamnist:member_3}
    \end{subfigure}
    \hfill
    \begin{subfigure}[t]{0.24\textwidth}
    \centering
    \includegraphics[width=\textwidth]{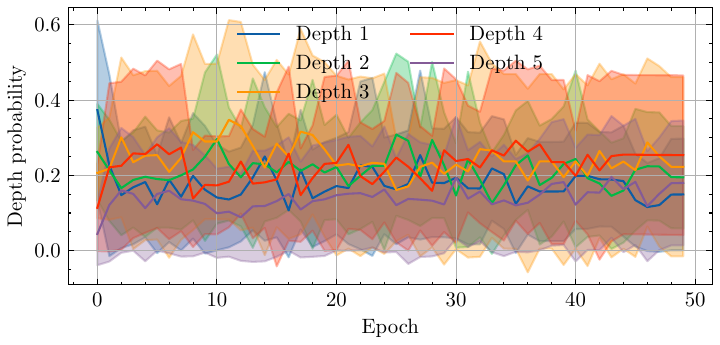}
    \caption{$N=4$.}
    \label{fig:depth:pneumoniamnist:member_4}
    \end{subfigure}
    \newline
    \begin{subfigure}[t]{0.19\textwidth}
    \centering
    \includegraphics[width=\textwidth]{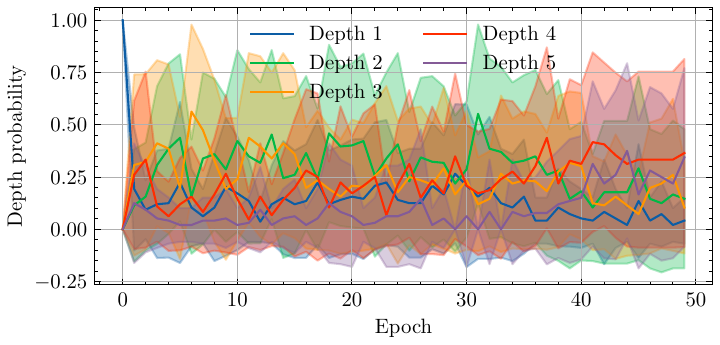}
    \caption{$K=1$.}
    \label{fig:depth:pneumoniamnist:k_1}
    \end{subfigure}
    \hfill
    \begin{subfigure}[t]{0.19\textwidth}
    \centering
    \includegraphics[width=\textwidth]{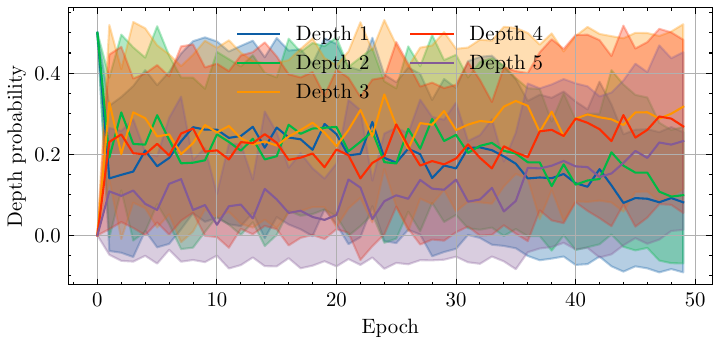}
    \caption{$K=2$.}
    \label{fig:depth:pneumoniamnist:k_2}
    \end{subfigure}
    \hfill
    \begin{subfigure}[t]{0.19\textwidth}
    \centering
    \includegraphics[width=\textwidth]{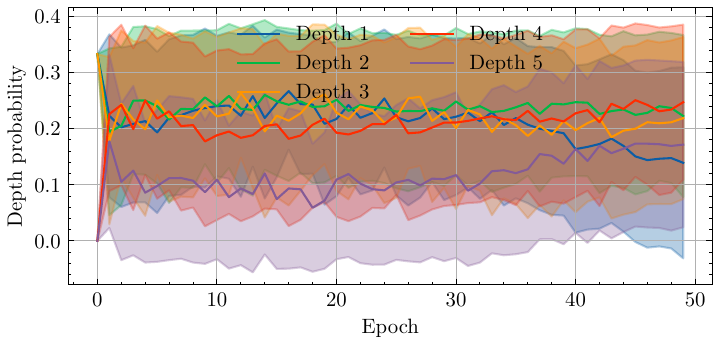}
    \caption{$K=3$.}
    \label{fig:depth:pneumoniamnist:k_3}
    \end{subfigure}
    \hfill
    \begin{subfigure}[t]{0.19\textwidth}
    \centering
    \includegraphics[width=\textwidth]{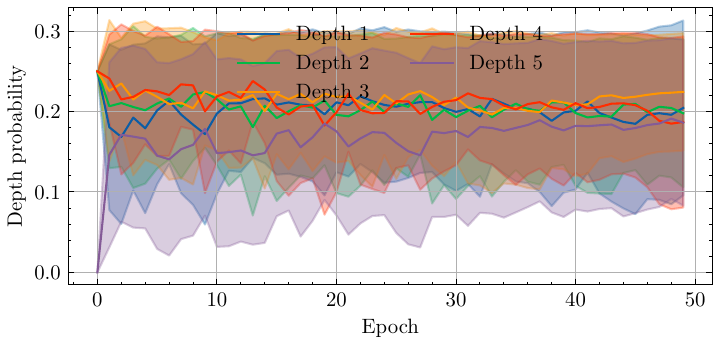}
    \caption{$K=4$.}
    \label{fig:depth:pneumoniamnist:k_4}
    \end{subfigure}
    \begin{subfigure}[t]{0.19\textwidth}
    \centering
    \includegraphics[width=\textwidth]{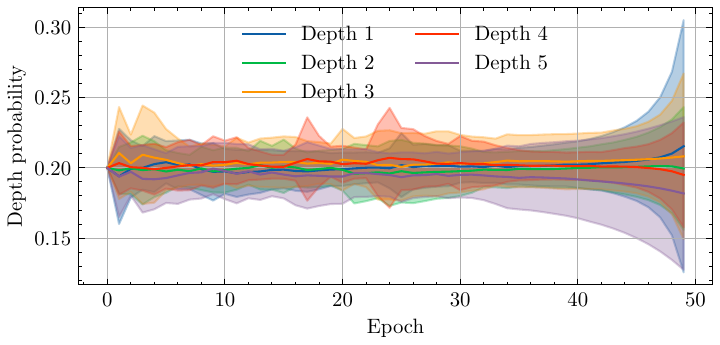}
    \caption{$K=D$.}
    \label{fig:depth:pneumoniamnist:k_5}
    \end{subfigure}
    \vspace{0.2cm}
    \caption{Depth preference during training when averaging over different $N$ and $K$ for PneumoniaMNIST.
    The lines denote the mean trend, and the shaded regions denote the standard deviation across configurations.}
    \label{fig:depth:pneumoniamnist}
\end{figure}

In Figures~\ref{fig:depth:tinyimagenet:members},~\ref{fig:depth:retinamnist:members},~\ref{fig:depth:bloodmnist} and~\ref{fig:depth:pneumoniamnist}, we show the depth preference for different $N$ and $K$ on TinyImageNet, RetinaMNIST, BloodMNIST and PneumoniaMNIST, respectively.
We categorise the results according to $N$ and $K$ and show the average depth preference across all $K$ or $N$, respectively.
The results show that the depth preference is consistent across different $N$ and $K$.
Depending on the complexity of the problem and the capacity of the backbone architecture, the depth preference can vary, but the general trend is that the deeper exits are preferred in comparison to the shallower ones.
As $K$ is increased, the networks also utilise the shallower exits more, however, the preference for the deeper exits is still present.
Interestingly, as $N$ is increased, the depth preference changes and we see that, for example, the network might prefer the deepest exit for processing one input and a shallower exit for processing another input.
This is best seen in the TinyImageNet dataset in Figure~\ref{fig:depth:tinyimagenet:members}.
We suspect that this is the network's way of adapting to the complexity of the input data and the capacity of the backbone architecture.

\section{Supplementary for Hyperparameter Influence}\label{sec:appendix:hyperparameter_influence}

In Tables~\ref{tab:correlation:tinyimagenet:id:test},~\ref{tab:correlation:bloodmnist:id:test},~\ref{tab:correlation:pneumoniamnist:id:test}, and~\ref{tab:correlation:retinamnist:id:test} we benchmark the influence of the hyperparameters on the performance of the SAE across all datasets and metrics.
The hyperparameters are the number of processed inputs $N$, the maximum number of early exits $K$, the width multiplier $W$, the start and end input repetition probability $i_{\text{start}}$ and $i_{\text{end}}$, the start and end temperature $T_{\text{start}}$ and $T_{\text{end}}$ and the start and end values of the trade-off regulariser $\alpha_{\text{start}}$ and $\alpha_{\text{end}}$.
We measured the correlation between the hyperparameters and the performance metrics via the Spearman rank correlation coefficient~\cite{zar2005spearman} when the hyperparameters are continuous such as $i(\cdot)$, $T(\cdot)$, and $\alpha(\cdot)$.
For categorical hyperparameters such as $N$, $K$, and $W$, we computed the ANOVA F-value~\cite{girden1992anova}.
Note that ANOVA is not a direct measure of correlation, but it is used to determine whether there are statistically significant differences between the means of two or more groups.
Hence it does not provide a direct measure of the strength of the relationship between the hyperparameter and the performance metric but it can be used to determine the importance of the hyperparameter's value on the performance metric.

The influence of input repetition probability $i$ varies across datasets in its impact on metrics such as NLL and Accuracy, suggesting its optimisation should be tailored per dataset characteristics.
The temperature $T$ demonstrates minimal but consistent influence across datasets, especially towards the end values $T_{\text{end}}$, indicating its role in the stability and fine-tuning of the SAE NN.
Correlations with performance metrics such as NLL and Accuracy fluctuate for $\alpha$, which points towards its role in controlling the trade-off between different data fitting and regularisation.
The variance of $N$ and $K$ show that they strongly influence predictive performance and computational overhead across datasets.
The width multiplier $W$ exerts substantial influence on the computational demand, FLOPs and number of parameters, particularly evident in the RetinaMNIST dataset.
Changes in $W$ also affect model accuracy and other performance metrics, which suggests that increasing the  network width is crucial for capturing adequate feature representations enabling the fitting of multiple predictors in the same architecture.

The $N$ and $K$ hyperparameters need to be observed in the context of $\alpha(t)$, $T(t)$ and $i(t)$, which balance the data fitting and regularisation. 
If $\alpha(t)$, is high the distribution of the exit weights will be more spread out across exits, showing that the network's representations are already useful for predictions at earlier depths. 
If the $T(t)$ is low, the exit weights will be sharper, showing focus only on the most confident depths. 
Lastly, if the $i(t)$ is high, the subnetworks will reuse more features between each other, as discussed in MIMO. 
In detail, we noticed that to balance high $N,K$, the HPO would decrease $\alpha(t)$ and $T(t)$ and increase $i(t)$ and vice versa. 
We want to stress that HPO chose different start and end points of the $\alpha(t)$, $T(t)$ and $i(t)$ demonstrating that for the best performance, the hyperparameters need to be scheduled to allow for the phase transitions between data fitting and regularisation.

\clearpage

\clearpage
\begin{table}
    \centering
        \begin{sc}
        \begin{adjustbox}{max width=\linewidth}
            \begin{tabular}{|l|ccc|cc|}
                \hline
                 \textbf{Method}                           & \textbf{NLL}        & \textbf{Accuracy/F1}          & \textbf{ECE/CC-ECE}      & \textbf{FLOPs}         & \textbf{Params}        \\
                 \hline\hline
                 \textcolor{Gray}{Standard NN $\newmoon$}        &     5.67 $\pm$ 0.47 & 4.67 $\pm$ 2.05 & 6.00 $\pm$ 1.63 & \textbf{1.00 $\pm$ 0.00} & 3.00 $\pm$ 0.00  \\
                 \textcolor{Blue}{NN Ensemble $\blacksquare$}        &  \textbf{3.00 $\pm$ 1.41} & \textbf{2.67 $\pm$ 1.25} & 3.67 $\pm$ 2.49 & 7.00 $\pm$ 2.16 & 6.33 $\pm$ 2.05 \\
                 \textcolor{Cyan}{MCD $\blacktriangle$} & 7.67 $\pm$ 1.89 & 9.00 $\pm$ 0.00 & 4.00 $\pm$ 2.83 & 8.00 $\pm$ 0.82 & \textbf{1.00 $\pm$ 0.00} \\
                 \textcolor{Pink}{BE $\blacktriangleleft$} & 6.00 $\pm$ 2.16 & 6.33 $\pm$ 1.25 & 6.33 $\pm$ 0.47 & 7.33 $\pm$ 0.47 & 2.67 $\pm$ 0.94 \\
                 \textcolor{Orange}{MIMO; $N\geq2, K=1$ \HexagonMarker} & 4.00 $\pm$ 2.16 & 4.67 $\pm$ 2.05 & 4.33 $\pm$ 0.94 & 4.33 $\pm$ 1.70 & 6.67 $\pm$ 1.25 \\
                 \textcolor{Orange}{MIMMO; $N\geq2, K=D$ \PentagonMarker}   &\textbf{3.00 $\pm$ 2.83} & 4.00 $\pm$  2.45 & 5.67$\pm$  2.36 & 3.67 $\pm$  1.70 & 5.33 $\pm$  2.87 \\
                 \textcolor{Orange}{I/B; $N\geq2, 2\leq K\leq D$ \PlusMarker} & 3.33 $\pm$ 1.25 & 3.67 $\pm$ 1.70 & \textbf{3.00 $\pm$ 2.83} & 6.67 $\pm$ 1.70 & 8.33 $\pm$ 0.47 \\
                \textcolor{Orange}{SE NN; $N=1, K=1$ \TimesMarker} & 7.67 $\pm$ 1.25 & 7.33 $\pm$ 0.94 & 6.33 $\pm$ 3.09 & 3.00 $\pm$ 0.82 & 5.00 $\pm$ 0.82 \\
                \textcolor{Orange}{EE; $N=1, K\geq2$ \OctagonMarker} & 4.67 $\pm$ 2.49 & \textbf{2.67 $\pm$ 1.25} & 5.67 $\pm$ 2.49 & 4.00 $\pm$ 0.82 & 6.67 $\pm$ 0.47 \\
    \hline
    \end{tabular}
    \end{adjustbox}
    \end{sc}
    \vspace{0.2cm}
    \caption{Average ranking across all classification ID test datasets. 
    We took all the Pareto candidates for a method and a dataset, calculated the mean performance of the candidates across all metrics and then ranked the methods with respect to each other and their mean per-metric performance. 
    Then we averaged the ranks across all the datasets.
    Lower rank is better.
    }
    \label{tab:overall_performance:id:classification}
\end{table}
\begin{table}
    \centering
        \begin{sc}
        \begin{adjustbox}{max width=\linewidth}
            \begin{tabular}{|l|ccc|cc|}
                \hline
                 \textbf{Method}                           & \textbf{NLL}        & \textbf{Accuracy/F1}          & \textbf{ECE/CC-ECE}      & \textbf{FLOPs}         & \textbf{Params}        \\
                 \hline\hline
                 \textcolor{Gray}{Standard NN $\newmoon$}        &     5.67 $\pm$ 1.25 & 5.00 $\pm$ 2.16 & 4.00 $\pm$ 1.41 & \textbf{2.00 $\pm$ 1.41} & 3.33 $\pm$ 2.05  \\
                 \textcolor{Blue}{NN Ensemble $\blacksquare$}        &  4.00 $\pm$ 1.63 & 4.67 $\pm$ 2.36 & 4.67 $\pm$ 2.05 & 6.00 $\pm$ 2.45 & 5.00 $\pm$ 2.45 \\
                 \textcolor{Cyan}{MCD $\blacktriangle$} & 5.67 $\pm$ 1.70 & 9.00 $\pm$ 0.00 & \textbf{1.33 $\pm$ 0.47} & 8.00 $\pm$ 0.82 & \textbf{2.33 $\pm$ 1.89} \\
                 \textcolor{Pink}{BE $\blacktriangleleft$} & 5.67 $\pm$ 3.40 & 5.00 $\pm$ 2.94 & 4.33 $\pm$ 2.05 & 6.33 $\pm$ 1.70 & 3.33 $\pm$ 1.25 \\
                 \textcolor{Orange}{MIMO; $N\geq2, K=1$ \HexagonMarker} & 3.67 $\pm$ 0.94 & 4.67 $\pm$ 1.25 & 3.67 $\pm$ 2.05 & 5.00 $\pm$ 2.45 & 6.67 $\pm$ 2.05 \\
                 \textcolor{Orange}{MIMMO; $N\geq2, K=max_K$ \PentagonMarker} & \textbf{1.67 $\pm$ 0.94} & \textbf{2.00 $\pm$ 1.41} & 5.33 $\pm$ 1.70 & 5.00 $\pm$ 1.41 & 6.33 $\pm$ 2.05 \\
                 \textcolor{Orange}{I/B; $N\geq2, 2\leq K\leq D$ \PlusMarker} & 3.67 $\pm$ 2.36 & 3.67 $\pm$ 2.36 & 5.67 $\pm$ 1.70 & 7.00 $\pm$ 1.63 & 8.33 $\pm$ 0.47 \\
                 \textcolor{Orange}{SE NN; $N=1, K=1$ \TimesMarker} & 7.67 $\pm$ 1.89 & 6.67 $\pm$ 1.25 & 8.67 $\pm$ 0.47 & 2.33 $\pm$ 0.47 & 4.33 $\pm$ 1.25 \\
                 \textcolor{Orange}{EE; $N=1, K\geq2$ \OctagonMarker} & 7.33 $\pm$ 0.94 & 4.33 $\pm$ 0.47 & 7.33 $\pm$ 1.70 & 3.33 $\pm$ 1.70 & 5.33 $\pm$ 2.36 \\
    \hline
    \end{tabular}
    \end{adjustbox}
    \end{sc}
    \vspace{0.2cm}
    \caption{Average ranking across all classification OOD test datasets. 
    We took all the Pareto candidates for a method and a dataset, calculated the mean performance of the candidates across all metrics and then ranked the methods with respect to each other and their mean per-metric performance. 
    Then we averaged the ranks across all the datasets.
    Lower rank is better.
    }
    \label{tab:overall_performance:ood:classification}
\end{table}
\begin{table}
    \centering
        \begin{sc}
        \begin{adjustbox}{max width=\linewidth}
            \begin{tabular}{|l|cc|cc|}
                \hline
                 \textbf{Method}                           & \textbf{NLL}        & \textbf{MSE}              & \textbf{FLOPs}         & \textbf{Params}        \\
                 \hline\hline
                 \textcolor{Gray}{Standard NN $\newmoon$}        & 6 & 5 & \textbf{1} & \textbf{1} \\
                 \textcolor{Blue}{NN Ensemble $\blacksquare$}         & 7 & 2 & 8 & 9 \\
                 \textcolor{Cyan}{MCD $\blacktriangle$} & 4 & 6 & 9 & 6 \\
                 \textcolor{Pink}{BE $\blacktriangleleft$} & 9 & 7 & 5 & 4 \\
                 \textcolor{Orange}{MIMO; $N\geq2, K=1$ \HexagonMarker} & 2 & 8 & 2 & 2  \\
                 \textcolor{Orange}{MIMMO; $N\geq2, K=D$ \PentagonMarker}   & 5 & 4 & 6 & 7  \\
                 \textcolor{Orange}{I/B; $N\geq2, 2\leq K\leq D$ \PlusMarker} & \textbf{1} &  3 & 7 & 8  \\
                \textcolor{Orange}{SE NN; $N=1, K=1$ \TimesMarker} & 3  &  9 & 3 & 3  \\
                \textcolor{Orange}{EE; $N=1, K\geq2$ \OctagonMarker} & 8 & \textbf{1} & 4 & 5 \\
    \hline
    \end{tabular}
    \end{adjustbox}
    \end{sc}
    \vspace{0.2cm}
    \caption{Ranking across RetinaMNIST ID test dataset. 
    We took all the Pareto candidates for a method, calculated the mean performance of the candidates across all metrics and then ranked the methods with respect to each other and their mean per-metric performance. 
    Lower rank is better.
    }
    \label{tab:overall_performance:id:regression}
\end{table}
\begin{table}
    \centering
        \begin{sc}
        \begin{adjustbox}{max width=\linewidth}
            \begin{tabular}{|l|cc|cc|}
                \hline
                 \textbf{Method}                           & \textbf{NLL}        & \textbf{MSE}              & \textbf{FLOPs}         & \textbf{Params}        \\
                 \hline\hline
                 \textcolor{Gray}{Standard NN $\newmoon$}        & 8 & 5 & 3 & 3 \\
                 \textcolor{Blue}{NN Ensemble $\blacksquare$}         & 4 & 6 & \textbf{1} & \textbf{1} \\
                 \textcolor{Cyan}{MCD $\blacktriangle$} & 7 & 8 & 9 & 8 \\
                 \textcolor{Pink}{BE $\blacktriangleleft$} & 9 & 3 & 8 & 5 \\
                 \textcolor{Orange}{MIMO; $N\geq2, K=1$ \HexagonMarker} & 3 & 7 & 2 & 2  \\
                 \textcolor{Orange}{MIMMO; $N\geq2, K=D$ \PentagonMarker}   & 2 & \textbf{1} & 5 & 6  \\
                 \textcolor{Orange}{I/B; $N\geq2, 2\leq K\leq D$ \PlusMarker} & \textbf{1} &  2 & 7 & 9  \\
                \textcolor{Orange}{SE NN; $N=1, K=1$ \TimesMarker} & 5  &  9 & 4 & 4  \\
                \textcolor{Orange}{EE; $N=1, K\geq2$ \OctagonMarker} & 6 & 4 & 6 & 7 \\
    \hline
    \end{tabular}
    \end{adjustbox}
    \end{sc}
    \vspace{0.2cm}
    \caption{Ranking across RetinaMNIST OOD test dataset. 
    We took all the Pareto candidates for a method, calculated the mean performance of the candidates across all metrics and then ranked the methods with respect to each other and their mean per-metric performance. 
    Lower rank is better.
    }
    \label{tab:overall_performance:ood:regression}
\end{table}
\begin{table}
    \centering
        \begin{sc}
        \begin{adjustbox}{max width=\linewidth}
            \begin{tabular}{|l|cccc|cc|}
                \hline
                 \textbf{Method}                           & \textbf{Count}      & \textbf{NLL [nats]}        & \textbf{Accuracy [0-1]}          & \textbf{ECE [0-1]}      & \textbf{FLOPs [M]}         & \textbf{Params [M]}        \\
                 \hline\hline
                 \multirow{3}{*}{\textcolor{Gray}{Standard NN $\newmoon$}}        & \multirow{3}{*}{4}  &     2.540 $\pm$ 0.006      &     0.396 $\pm$ 0.001      &     0.010 $\pm$ 0.002      & \textbf{240.137 $\pm$ 0.000} & \textbf{0.241 $\pm$ 0.000} \\
                 &                     &     3.796 $\pm$ 0.025      &     0.531 $\pm$ 0.001      &     0.330 $\pm$ 0.002      &     963.251 $\pm$ 0.000      &     8.218 $\pm$ 0.000      \\
                 &                     &     2.540 $\pm$ 0.006      &     0.396 $\pm$ 0.001      &     0.010 $\pm$ 0.002      &     240.137 $\pm$ 0.000      &     0.241 $\pm$ 0.000      \\ \hline
                 \multirow{3}{*}{\textcolor{Blue}{NN Ensemble $\blacksquare$}}        & \multirow{3}{*}{3}  &     2.022 $\pm$ 0.008      &     0.556 $\pm$ 0.001      &     0.037 $\pm$ 0.002      &     2105.251 $\pm$ 0.000     &     5.481 $\pm$ 0.000      \\
                 &                     &     \underline{2.388 $\pm$ 0.006}      & \underline{\textbf{0.597 $\pm$ 0.002}} &     \underline{0.083 $\pm$ 0.003}      &     \underline{3853.002 $\pm$ 0.000}     &     \underline{32.873 $\pm$ 0.000}     \\
                 &                     &     2.538 $\pm$ 0.015      &     0.397 $\pm$ 0.003      &     0.009 $\pm$ 0.002      &     240.137 $\pm$ 0.000      &     0.241 $\pm$ 0.000      \\ \hline
                 \multirow{3}{*}{\textcolor{Cyan}{MCD $\blacktriangle$}}            & \multirow{3}{*}{19} &     2.086 $\pm$ 0.009      &     0.505 $\pm$ 0.003      &     0.015 $\pm$ 0.003      &     2110.102 $\pm$ 0.000     &     1.370 $\pm$ 0.000      \\
                 &                     &     2.744 $\pm$ 0.020      &     0.530 $\pm$ 0.002      &     0.172 $\pm$ 0.004      &     3858.836 $\pm$ 0.000     &     8.218 $\pm$ 0.000      \\
                 &                     &     5.139 $\pm$ 0.002      &     0.012 $\pm$ 0.001      & \textbf{0.003 $\pm$ 0.001} &     3858.836 $\pm$ 0.000     &     8.218 $\pm$ 0.000      \\ \hline
                 \multirow{3}{*}{\textcolor{Pink}{BE $\blacktriangleleft$}}      & \multirow{3}{*}{7}  &     2.520 $\pm$ 0.005      &     0.407 $\pm$ 0.003      &     0.032 $\pm$ 0.005      &     483.445 $\pm$ 0.000      &     0.244 $\pm$ 0.000      \\
                 &                     &     3.980 $\pm$ 0.020      &     0.530 $\pm$ 0.004      &     0.338 $\pm$ 0.002      &     3863.801 $\pm$ 0.000     &     8.259 $\pm$ 0.000      \\
                 &                     &     2.523 $\pm$ 0.011      &     0.406 $\pm$ 0.002      &     0.027 $\pm$ 0.003      &     966.889 $\pm$ 0.000      &     0.247 $\pm$ 0.000      \\ \hline
                 \multirow{3}{*}{\textcolor{Orange}{MIMO; $N\geq2, K=1$ \HexagonMarker}}    & \multirow{3}{*}{2}  &     2.071 $\pm$ 0.048      &     0.505 $\pm$ 0.009      &     0.051 $\pm$ 0.014      &    1081.378 $\pm$ 81.796     &     12.078 $\pm$ 5.669     \\
                 &                     &     2.522 $\pm$ 0.048      &     0.511 $\pm$ 0.004      &     0.210 $\pm$ 0.003      &     1180.377 $\pm$ 0.000     &     21.488 $\pm$ 0.000     \\
                 &                     &     2.071 $\pm$ 0.048      &     0.505 $\pm$ 0.009      &     0.051 $\pm$ 0.014      &    1081.378 $\pm$ 81.796     &     12.078 $\pm$ 5.669     \\ \hline
                 \multirow{3}{*}{\textcolor{Orange}{MIMMO; $N\geq2, K=D$ \PentagonMarker}}   & \multirow{3}{*}{4}  &     1.807 $\pm$ 0.013      &     0.563 $\pm$ 0.005      &     0.063 $\pm$ 0.003      &     1243.154 $\pm$ 0.000     &     22.338 $\pm$ 0.000     \\
                 &                     &     1.808 $\pm$ 0.003      &     0.569 $\pm$ 0.001      &     0.046 $\pm$ 0.002      &     1243.154 $\pm$ 0.000     &     22.338 $\pm$ 0.000     \\
                 &                     &     1.815 $\pm$ 0.016      &     0.561 $\pm$ 0.004      &     0.026 $\pm$ 0.003      &     1243.051 $\pm$ 0.000     &     22.235 $\pm$ 0.000     \\ \hline
                 \multirow{3}{*}{\textcolor{Orange}{SE NN; $N=1, K=1$ \TimesMarker}}     & \multirow{3}{*}{3}  &     3.566 $\pm$ 0.056      &     0.474 $\pm$ 0.004      &     0.319 $\pm$ 0.006      &     1173.197 $\pm$ 0.000     &     21.384 $\pm$ 0.000     \\
                 &                     &     4.252 $\pm$ 0.011      &     0.488 $\pm$ 0.002      &     0.363 $\pm$ 0.002      &     1173.197 $\pm$ 0.000     &     21.384 $\pm$ 0.000     \\
                 &                     &     3.566 $\pm$ 0.056      &     0.474 $\pm$ 0.004      &     0.319 $\pm$ 0.006      &     1173.197 $\pm$ 0.000     &     21.384 $\pm$ 0.000     \\ \hline
                 \multirow{3}{*}{\textcolor{Orange}{EE; $N=1, K\geq2$ \OctagonMarker}}     & \multirow{3}{*}{2}  &     3.596 $\pm$ 0.020      &     0.526 $\pm$ 0.005      &     0.337 $\pm$ 0.092      &     1181.852 $\pm$ 0.000     &     21.619 $\pm$ 0.000     \\
                 &                     &     3.596 $\pm$ 0.020      &     0.526 $\pm$ 0.005      &     0.337 $\pm$ 0.092      &     1181.852 $\pm$ 0.000     &     21.619 $\pm$ 0.000     \\
                 &                     &     3.596 $\pm$ 0.020      &     0.526 $\pm$ 0.005      &     0.337 $\pm$ 0.092      &     1181.852 $\pm$ 0.000     &     21.619 $\pm$ 0.000     \\ \hline
                 \multirow{3}{*}{\textcolor{Orange}{I/B; $N\geq2, 2\leq K\leq D$ \PlusMarker}} & \multirow{3}{*}{5}  & \textit{\textbf{1.801 $\pm$ 0.026}} &     \textit{0.564 $\pm$ 0.009}      &     \textit{0.040 $\pm$ 0.006}      &     \textit{1206.773 $\pm$ 0.000}     &     \textit{22.098 $\pm$ 0.000}     \\
                 &                     &     1.801 $\pm$ 0.026      &     0.564 $\pm$ 0.009      &     0.040 $\pm$ 0.006      &     1206.773 $\pm$ 0.000     &     22.098 $\pm$ 0.000     \\
                 &                     &     2.068 $\pm$ 0.010      &     0.497 $\pm$ 0.003      &     0.018 $\pm$ 0.001      &     1196.418 $\pm$ 0.000     &     22.033 $\pm$ 0.000     \\
    \hline
    \end{tabular}
    \end{adjustbox}
    \end{sc}
    \vspace{0.2cm}
    \caption{Best results for empirically Pareto optimal configurations on TinyImageNet ID dataset. 
    Each row represents the best configuration for each method for the first metric, the second metric, etc.
    The count represents how many configurations were in the Pareto front.
    The \textit{italic} and \underline{underlined} values represent the compared configurations in the main text, respectively.
    The \textbf{bold} values represent the best value across all methods for each metric.
    }
    \label{tab:tinyimagenet:id}
    \end{table}

\begin{table}
    \centering
        \begin{sc}
        \begin{adjustbox}{max width=\linewidth}
            \begin{tabular}{|l|cccc|cc|}
                \hline
                 \textbf{Method}                           & \textbf{Count}      & \textbf{NLL [nats]}        & \textbf{Accuracy [0-1]}          & \textbf{ECE [0-1]}      & \textbf{FLOPs [M]}         & \textbf{Params [M]}        \\
                 \hline\hline
                 \multirow{3}{*}{\textcolor{Gray}{Standard NN $\newmoon$}}        & \multirow{3}{*}{3}  &     4.803 $\pm$ 1.586      &     0.164 $\pm$ 0.104      &     0.157 $\pm$ 0.094      & \textbf{240.137 $\pm$ 0.000} & \textbf{0.241 $\pm$ 0.000} \\
                 &                     &     7.033 $\pm$ 2.485      &     0.298 $\pm$ 0.135      &     0.497 $\pm$ 0.105      &     1173.197 $\pm$ 0.000     &     21.384 $\pm$ 0.000     \\
                 &                     &     4.803 $\pm$ 1.586      &     0.164 $\pm$ 0.104      &     0.157 $\pm$ 0.094      &     240.137 $\pm$ 0.000      &     0.241 $\pm$ 0.000      \\ \hline
                 \multirow{3}{*}{\textcolor{Blue}{NN Ensemble $\blacksquare$}}        & \multirow{3}{*}{3}  &     4.387 $\pm$ 1.809      &     0.284 $\pm$ 0.151      &     0.143 $\pm$ 0.090      &     2105.251 $\pm$ 0.000     &     5.481 $\pm$ 0.000      \\
                 &                     &     \underline{5.219 $\pm$ 2.330}      & \underline{\textbf{0.342 $\pm$ 0.158}} &     \underline{0.211 $\pm$ 0.100}      &     \underline{3853.002 $\pm$ 0.000}     &     \underline{32.873 $\pm$ 0.000}     \\
                 &                     &     4.477 $\pm$ 1.460      &     0.182 $\pm$ 0.115      &     0.102 $\pm$ 0.070      &     960.548 $\pm$ 0.000      &     0.964 $\pm$ 0.000      \\ \hline
                 \multirow{3}{*}{\textcolor{Cyan}{MCD $\blacktriangle$}}            & \multirow{3}{*}{26} &     4.180 $\pm$ 1.517      &     0.240 $\pm$ 0.136      &     0.134 $\pm$ 0.097      &     2110.102 $\pm$ 0.000     &     1.370 $\pm$ 0.000      \\
                 &                     &     5.731 $\pm$ 2.304      &     0.282 $\pm$ 0.143      &     0.317 $\pm$ 0.099      &     3858.836 $\pm$ 0.000     &     8.218 $\pm$ 0.000      \\
                 &                     &     5.158 $\pm$ 0.019      &     0.012 $\pm$ 0.001      & \textbf{0.002 $\pm$ 0.001} &     3858.836 $\pm$ 0.000     &     8.218 $\pm$ 0.000      \\ \hline
                 \multirow{3}{*}{\textcolor{Pink}{BE $\blacktriangleleft$}}      & \multirow{3}{*}{8}  &     5.050 $\pm$ 1.891      &     0.168 $\pm$ 0.107      &     0.189 $\pm$ 0.112      &     966.889 $\pm$ 0.000      &     0.247 $\pm$ 0.000      \\
                 &                     &     7.302 $\pm$ 2.542      &     0.298 $\pm$ 0.136      &     0.446 $\pm$ 0.069      &     3528.084 $\pm$ 0.000     &     21.435 $\pm$ 0.000     \\
                 &                     &     5.050 $\pm$ 1.891      &     0.168 $\pm$ 0.107      &     0.189 $\pm$ 0.112      &     966.889 $\pm$ 0.000      &     0.247 $\pm$ 0.000      \\ \hline
                 \multirow{3}{*}{\textcolor{Orange}{MIMO; $N\geq2, K=1$ \HexagonMarker}}    & \multirow{3}{*}{1}  &     3.902 $\pm$ 1.371      &     0.263 $\pm$ 0.136      &     0.122 $\pm$ 0.078      &    1081.378 $\pm$ 81.796     &     12.078 $\pm$ 5.669     \\
                 &                     &     3.902 $\pm$ 1.371      &     0.263 $\pm$ 0.136      &     0.122 $\pm$ 0.078      &    1081.378 $\pm$ 81.796     &     12.078 $\pm$ 5.669     \\
                 &                     &     3.902 $\pm$ 1.371      &     0.263 $\pm$ 0.136      &     0.122 $\pm$ 0.078      &    1081.378 $\pm$ 81.796     &     12.078 $\pm$ 5.669     \\ \hline
                 \multirow{3}{*}{\textcolor{Orange}{MIMMO; $N\geq2, K=D$ \PentagonMarker}} & \multirow{3}{*}{5}   & \textbf{3.686 $\pm$ 1.425} &     0.301 $\pm$ 0.151      &     0.138 $\pm$ 0.087      &     1243.051 $\pm$ 0.000     &     22.235 $\pm$ 0.000     \\
                 &                     &     3.704 $\pm$ 1.445      &     0.306 $\pm$ 0.153      &     0.129 $\pm$ 0.086      &     1243.154 $\pm$ 0.000     &     22.338 $\pm$ 0.000     \\
                 &                     &     3.697 $\pm$ 1.420      &     0.296 $\pm$ 0.152      &     0.106 $\pm$ 0.078      &     1243.154 $\pm$ 0.000     &     22.338 $\pm$ 0.000     \\ \hline
                 \multirow{3}{*}{\textcolor{Orange}{SE NN; $N=1, K=1$ \TimesMarker}}     & \multirow{3}{*}{4}  &     7.324 $\pm$ 2.499      &     0.223 $\pm$ 0.127      &     0.494 $\pm$ 0.091      &     1173.197 $\pm$ 0.000     &     21.384 $\pm$ 0.000     \\
                 &                     &     8.162 $\pm$ 2.635      &     0.235 $\pm$ 0.129      &     0.546 $\pm$ 0.098      &     1173.197 $\pm$ 0.000     &     21.384 $\pm$ 0.000     \\
                 &                     &     7.324 $\pm$ 2.499      &     0.223 $\pm$ 0.127      &     0.494 $\pm$ 0.091      &     1173.197 $\pm$ 0.000     &     21.384 $\pm$ 0.000     \\ \hline
                 \multirow{3}{*}{\textcolor{Orange}{EE; $N=1, K\geq2$ \OctagonMarker}}     & \multirow{3}{*}{1}  &     7.375 $\pm$ 2.668      &     0.268 $\pm$ 0.143      &     0.462 $\pm$ 0.089      &     1181.852 $\pm$ 0.000     &     21.619 $\pm$ 0.000     \\
                 &                     &     7.375 $\pm$ 2.668      &     0.268 $\pm$ 0.143      &     0.462 $\pm$ 0.089      &     1181.852 $\pm$ 0.000     &     21.619 $\pm$ 0.000     \\
                 &                     &     7.375 $\pm$ 2.668      &     0.268 $\pm$ 0.143      &     0.462 $\pm$ 0.089      &     1181.852 $\pm$ 0.000     &     21.619 $\pm$ 0.000     \\ \hline
                 \multirow{3}{*}{\textcolor{Orange}{I/B; $N\geq2, 2\leq K\leq D$ \PlusMarker}} & \multirow{3}{*}{5}  &     \textit{3.723 $\pm$ 1.443}      &     \textit{0.303 $\pm$ 0.151}      &    \textit{0.164 $\pm$ 0.088}     &     \textit{1206.773 $\pm$ 0.000}     &     \textit{22.098 $\pm$ 0.000}     \\
                 &                     &     3.723 $\pm$ 1.443      &     0.303 $\pm$ 0.151      &     0.164 $\pm$ 0.088      &     1206.773 $\pm$ 0.000     &     22.098 $\pm$ 0.000     \\
                 &                     &     3.849 $\pm$ 1.334      &     0.259 $\pm$ 0.133      &     0.120 $\pm$ 0.092      &     1221.467 $\pm$ 0.044     &     22.640 $\pm$ 0.044     \\
    \hline
    \end{tabular}
    \end{adjustbox}
    \end{sc}    
    \vspace{0.2cm}
    \caption{Best results for empirically Pareto optimal configurations on TinyImageNet OOD dataset. 
    Each row represents the best configuration for each method for the first metric, the second metric, etc.
    The count represents how many configurations were in the Pareto front.
    The \textit{italic} and \underline{underlined} values represent the compared configurations in the main text, respectively.
    The \textbf{bold} values represent the best value across all methods for each metric.
    }
    \label{tab:tinyimagenet:ood}
    \end{table}

\begin{table}
\centering
    \begin{sc}
    \begin{adjustbox}{max width=\linewidth}
        \begin{tabular}{|l|cccc|cc|}
            \hline
             \textbf{Method}                           & \textbf{Count}      & \textbf{NLL [nats]}        & \textbf{F1 [0-1]}          & \textbf{CC-ECE [0-1]}      & \textbf{FLOPs [M]}         & \textbf{Params [M]}        \\
             \hline\hline
             \multirow{3}{*}{\textcolor{Gray}{Standard NN $\newmoon$}}        & \multirow{3}{*}{1} &     0.129 $\pm$ 0.004      &     0.950 $\pm$ 0.003      &     0.023 $\pm$ 0.003      & \textbf{4.697 $\pm$ 0.000} &     0.246 $\pm$ 0.000      \\
             &                    &     0.129 $\pm$ 0.004      &     0.950 $\pm$ 0.003      &     0.023 $\pm$ 0.003      &     4.697 $\pm$ 0.000      &     0.246 $\pm$ 0.000      \\
             &                    &     0.129 $\pm$ 0.004      &     0.950 $\pm$ 0.003      &     0.023 $\pm$ 0.003      &     4.697 $\pm$ 0.000      &     0.246 $\pm$ 0.000      \\ \hline
             \multirow{3}{*}{\textcolor{Blue}{NN Ensemble $\blacksquare$}}        & \multirow{3}{*}{6} & \textbf{0.102 $\pm$ 0.002} & \textbf{0.963 $\pm$ 0.001} &     0.023 $\pm$ 0.002      &     40.102 $\pm$ 0.000     &     2.315 $\pm$ 0.000      \\
             &                    &     0.103 $\pm$ 0.003      &     0.963 $\pm$ 0.001      &     0.023 $\pm$ 0.002      &     35.371 $\pm$ 0.000     &     1.134 $\pm$ 0.000      \\
             &                    &     \underline{0.115 $\pm$ 0.003}      &     \underline{0.958 $\pm$ 0.001}      & \underline{\textbf{0.020 $\pm$ 0.001}} &     \underline{17.686 $\pm$ 0.000}     &     \underline{0.567 $\pm$ 0.000}      \\ \hline
             \multirow{3}{*}{\textcolor{Cyan}{MCD $\blacktriangle$}}            & \multirow{3}{*}{5} &     0.127 $\pm$ 0.002      &     0.949 $\pm$ 0.001      &     0.023 $\pm$ 0.001      &     18.820 $\pm$ 0.000     &     0.246 $\pm$ 0.000      \\
             &                    &     0.129 $\pm$ 0.002      &     0.950 $\pm$ 0.003      &     0.020 $\pm$ 0.001      &     18.820 $\pm$ 0.000     &     0.246 $\pm$ 0.000      \\
             &                    &     0.129 $\pm$ 0.002      &     0.950 $\pm$ 0.003      &     0.020 $\pm$ 0.001      &     18.820 $\pm$ 0.000     &     0.246 $\pm$ 0.000      \\ \hline
             \multirow{3}{*}{\textcolor{Pink}{BE $\blacktriangleleft$}}      & \multirow{3}{*}{8} &     0.132 $\pm$ 0.002      &     0.951 $\pm$ 0.004      &     0.022 $\pm$ 0.002      &     14.218 $\pm$ 0.000     &     0.249 $\pm$ 0.000      \\
             &                    &     0.143 $\pm$ 0.004      &     0.951 $\pm$ 0.004      &     0.028 $\pm$ 0.001      &     30.243 $\pm$ 0.000     &     0.583 $\pm$ 0.000      \\
             &                    &     0.132 $\pm$ 0.002      &     0.951 $\pm$ 0.004      &     0.022 $\pm$ 0.002      &     14.218 $\pm$ 0.000     &     0.249 $\pm$ 0.000      \\ \hline
             \multirow{3}{*}{\textcolor{Orange}{MIMO; $N\geq2, K=1$ \HexagonMarker}}    & \multirow{3}{*}{2} &     \textit{0.122 $\pm$ 0.005}      &     \textit{0.956 $\pm$ 0.004}      &     \textit{0.022 $\pm$ 0.001}      &     \textit{11.424 $\pm$ 0.000}     &     \textit{0.600 $\pm$ 0.000}      \\
             &                    &     0.122 $\pm$ 0.005      &     0.956 $\pm$ 0.004      &     0.022 $\pm$ 0.001      &     11.424 $\pm$ 0.000     &     0.600 $\pm$ 0.000      \\
             &                    &     0.122 $\pm$ 0.005      &     0.956 $\pm$ 0.004      &     0.022 $\pm$ 0.001      &     11.424 $\pm$ 0.000     &     0.600 $\pm$ 0.000      \\ \hline
             \multirow{3}{*}{\textcolor{Orange}{MIMMO; $N\geq2, K=D$ \PentagonMarker}}   & \multirow{3}{*}{1} &     0.135 $\pm$ 0.006      &     0.948 $\pm$ 0.003      &     0.029 $\pm$ 0.001      &     5.621 $\pm$ 0.289      & \textbf{0.212 $\pm$ 0.072} \\
             &                    &     0.135 $\pm$ 0.006      &     0.948 $\pm$ 0.003      &     0.029 $\pm$ 0.001      &     5.621 $\pm$ 0.289      &     0.212 $\pm$ 0.072      \\
             &                    &     0.135 $\pm$ 0.006      &     0.948 $\pm$ 0.003      &     0.029 $\pm$ 0.001      &     5.621 $\pm$ 0.289      &     0.212 $\pm$ 0.072      \\ \hline
             \multirow{3}{*}{\textcolor{Orange}{SE NN; $N=1, K=1$ \TimesMarker}}     & \multirow{3}{*}{1} &     0.133 $\pm$ 0.006      &     0.948 $\pm$ 0.004      &     0.022 $\pm$ 0.001      &     9.468 $\pm$ 0.558      &     0.440 $\pm$ 0.139      \\
             &                    &     0.133 $\pm$ 0.006      &     0.948 $\pm$ 0.004      &     0.022 $\pm$ 0.001      &     9.468 $\pm$ 0.558      &     0.440 $\pm$ 0.139      \\
             &                    &     0.133 $\pm$ 0.006      &     0.948 $\pm$ 0.004      &     0.022 $\pm$ 0.001      &     9.468 $\pm$ 0.558      &     0.440 $\pm$ 0.139      \\ \hline
             \multirow{3}{*}{\textcolor{Orange}{EE; $N=1, K\geq2$ \OctagonMarker}}     & \multirow{3}{*}{4} &     0.123 $\pm$ 0.006      &     0.953 $\pm$ 0.004      &     0.025 $\pm$ 0.004      &     10.465 $\pm$ 0.132     &     0.610 $\pm$ 0.002      \\
             &                    &     0.123 $\pm$ 0.005      &     0.954 $\pm$ 0.004      &     0.024 $\pm$ 0.002      &     10.237 $\pm$ 0.000     &     0.606 $\pm$ 0.000      \\
             &                    &     0.132 $\pm$ 0.002      &     0.948 $\pm$ 0.001      &     0.021 $\pm$ 0.001      &     5.789 $\pm$ 0.000      &     0.336 $\pm$ 0.000      \\ \hline
             \multirow{3}{*}{\textcolor{Orange}{I/B; $N\geq2, 2\leq K\leq D$ \PlusMarker}} & \multirow{3}{*}{3} &     0.124 $\pm$ 0.003      &     0.951 $\pm$ 0.002      &     0.024 $\pm$ 0.002      &     10.539 $\pm$ 0.000     &     0.599 $\pm$ 0.000      \\
             &                    &     0.125 $\pm$ 0.006      &     0.954 $\pm$ 0.003      &     0.026 $\pm$ 0.003      &     11.499 $\pm$ 0.071     &     0.608 $\pm$ 0.004      \\
             &                    &     0.124 $\pm$ 0.003      &     0.951 $\pm$ 0.002      &     0.024 $\pm$ 0.002      &     10.539 $\pm$ 0.000     &     0.599 $\pm$ 0.000      \\
\hline
\end{tabular}
\end{adjustbox}
\end{sc}    
\vspace{0.2cm}
\caption{Best results for empirically Pareto optimal configurations on BloodMNIST ID dataset. 
Each row represents the best configuration for each method for the first metric, the second metric, etc.
The count represents how many configurations were in the Pareto front.
The \textit{italic} and \underline{underlined} values represent the compared configurations in the main text, respectively.
The \textbf{bold} values represent the best value across all methods for each metric.
}
\label{tab:bloodmnist:id}
\end{table}

\begin{table}
\centering
    \begin{sc}
    \begin{adjustbox}{max width=\linewidth}
        \begin{tabular}{|l|cccc|cc|}
            \hline
             \textbf{Method}                           & \textbf{Count}      & \textbf{NLL [nats]}        & \textbf{F1 [0-1]}          & \textbf{CC-ECE [0-1]}      & \textbf{FLOPs [M]}         & \textbf{Params [M]}        \\
             \hline\hline
             \multirow{3}{*}{\textcolor{Gray}{Standard NN $\newmoon$}}        & \multirow{3}{*}{5}  &     1.478 $\pm$ 1.255      &     0.560 $\pm$ 0.258      &     0.187 $\pm$ 0.092      &     0.737 $\pm$ 0.000      &     0.037 $\pm$ 0.000      \\
             &                     &     1.478 $\pm$ 1.255      &     0.560 $\pm$ 0.258      &     0.187 $\pm$ 0.092      &     0.737 $\pm$ 0.000      &     0.037 $\pm$ 0.000      \\
             &                     &     1.959 $\pm$ 0.990      &     0.223 $\pm$ 0.080      &     0.140 $\pm$ 0.025      & \textbf{0.128 $\pm$ 0.000} & \textbf{0.000 $\pm$ 0.000} \\ \hline
             \multirow{3}{*}{\textcolor{Blue}{NN Ensemble $\blacksquare$}}        & \multirow{3}{*}{17} &     \underline{1.206 $\pm$ 1.040}      &     \underline{0.582 $\pm$ 0.269}      &     \underline{0.198 $\pm$ 0.078}      &     \underline{2.948 $\pm$ 0.000}      &     \underline{0.147 $\pm$ 0.000}      \\
             &                     &     1.206 $\pm$ 1.040      &     0.582 $\pm$ 0.269      &     0.198 $\pm$ 0.078      &     2.948 $\pm$ 0.000      &     0.147 $\pm$ 0.000      \\
             &                     &     2.006 $\pm$ 1.530      &     0.225 $\pm$ 0.109      &     0.129 $\pm$ 0.030      &     0.256 $\pm$ 0.000      &     0.000 $\pm$ 0.000      \\ \hline
             \multirow{3}{*}{\textcolor{Cyan}{MCD $\blacktriangle$}}            & \multirow{3}{*}{11} &     1.571 $\pm$ 1.203      &     0.498 $\pm$ 0.221      &     0.182 $\pm$ 0.065      &     2.962 $\pm$ 0.000      &     0.037 $\pm$ 0.000      \\
             &                     &     1.571 $\pm$ 1.203      &     0.498 $\pm$ 0.221      &     0.182 $\pm$ 0.065      &     2.962 $\pm$ 0.000      &     0.037 $\pm$ 0.000      \\
             &                     &     2.005 $\pm$ 0.001      &     0.065 $\pm$ 0.004      & \textbf{0.011 $\pm$ 0.005} &     0.842 $\pm$ 0.000      &     0.000 $\pm$ 0.000      \\ \hline
             \multirow{3}{*}{\textcolor{Pink}{BE $\blacktriangleleft$}}      & \multirow{3}{*}{13} &     1.589 $\pm$ 1.425      &     0.553 $\pm$ 0.262      &     0.188 $\pm$ 0.095      &     2.259 $\pm$ 0.000      &     0.038 $\pm$ 0.000      \\
             &                     &     1.589 $\pm$ 1.425      &     0.553 $\pm$ 0.262      &     0.188 $\pm$ 0.095      &     2.259 $\pm$ 0.000      &     0.038 $\pm$ 0.000      \\
             &                     &     1.921 $\pm$ 1.127      &     0.249 $\pm$ 0.077      &     0.142 $\pm$ 0.026      &     0.270 $\pm$ 0.000      &     0.000 $\pm$ 0.000      \\ \hline
             \multirow{3}{*}{\textcolor{Orange}{MIMO; $N\geq2, K=1$ \HexagonMarker}}    & \multirow{3}{*}{4}  &     1.335 $\pm$ 1.217      &     0.543 $\pm$ 0.265      &     0.201 $\pm$ 0.074      &     3.327 $\pm$ 0.009      &     0.150 $\pm$ 0.002      \\
             &                     &     1.485 $\pm$ 1.445      &     0.556 $\pm$ 0.289      &     0.190 $\pm$ 0.091      &     6.767 $\pm$ 0.019      &     0.334 $\pm$ 0.005      \\
             &                     &     1.485 $\pm$ 1.445      &     0.556 $\pm$ 0.289      &     0.190 $\pm$ 0.091      &     6.767 $\pm$ 0.019      &     0.334 $\pm$ 0.005      \\ \hline
             \multirow{3}{*}{\textcolor{Orange}{MIMMO; $N\geq2, K=D$ \PentagonMarker}}   & \multirow{3}{*}{4}  & \textbf{1.194 $\pm$ 0.849} &     0.528 $\pm$ 0.233      &     0.208 $\pm$ 0.044      &     0.741 $\pm$ 0.005      &     0.019 $\pm$ 0.000      \\
             &                     &     1.277 $\pm$ 1.422      & \textbf{0.587 $\pm$ 0.281} &     0.179 $\pm$ 0.094      &     3.116 $\pm$ 0.000      &     0.153 $\pm$ 0.000      \\
             &                     &     1.277 $\pm$ 1.422      &     0.587 $\pm$ 0.281      &     0.179 $\pm$ 0.094      &     3.116 $\pm$ 0.000      &     0.153 $\pm$ 0.000      \\\hline
             \multirow{3}{*}{\textcolor{Orange}{SE NN; $N=1, K=1$ \TimesMarker}}     & \multirow{3}{*}{1}  &     1.866 $\pm$ 1.869      &     0.521 $\pm$ 0.267      &     0.201 $\pm$ 0.099      &     0.719 $\pm$ 0.030      &     0.032 $\pm$ 0.008      \\
             &                     &     1.866 $\pm$ 1.869      &     0.521 $\pm$ 0.267      &     0.201 $\pm$ 0.099      &     0.719 $\pm$ 0.030      &     0.032 $\pm$ 0.008      \\
             &                     &     1.866 $\pm$ 1.869      &     0.521 $\pm$ 0.267      &     0.201 $\pm$ 0.099      &     0.719 $\pm$ 0.030      &     0.032 $\pm$ 0.008      \\ \hline
             \multirow{3}{*}{\textcolor{Orange}{EE; $N=1, K\geq2$ \OctagonMarker}}     & \multirow{3}{*}{5}  &     1.636 $\pm$ 1.412      &     0.516 $\pm$ 0.272      &     0.229 $\pm$ 0.055      &     0.792 $\pm$ 0.020      &     0.040 $\pm$ 0.000      \\
             &                     &     2.158 $\pm$ 2.128      &     0.541 $\pm$ 0.290      &     0.201 $\pm$ 0.095      &     5.213 $\pm$ 0.000      &     0.279 $\pm$ 0.000      \\
             &                     &     1.664 $\pm$ 1.409      &     0.537 $\pm$ 0.283      &     0.197 $\pm$ 0.094      &     2.672 $\pm$ 0.000      &     0.150 $\pm$ 0.000      \\ \hline
             \multirow{3}{*}{\textcolor{Orange}{I/B; $N\geq2, 2\leq K\leq D$ \PlusMarker}} & \multirow{3}{*}{3}  &     1.387 $\pm$ 1.373      &     0.547 $\pm$ 0.266      &     0.198 $\pm$ 0.069      &     1.083 $\pm$ 0.005      &     0.040 $\pm$ 0.000      \\
             &                     &     \textit{1.451 $\pm$ 1.529}      &    \textit{0.554 $\pm$ 0.265}      &     \textit{0.185 $\pm$ 0.083}      &     1\textit{.075 $\pm$ 0.000}      &     \textit{0.040 $\pm$ 0.000}      \\
             &                     &     1.451 $\pm$ 1.529      &     0.554 $\pm$ 0.265      &     0.185 $\pm$ 0.083      &     1.075 $\pm$ 0.000      &     0.040 $\pm$ 0.000      \\
\hline
\end{tabular}
\end{adjustbox}
\end{sc}    
\vspace{0.2cm}
\caption{Best results for empirically Pareto optimal configurations on BloodMNIST OOD dataset. 
Each row represents the best configuration for each method for the first metric, the second metric, etc.
The count represents how many configurations were in the Pareto front.
The \textit{italic} and \underline{underlined} values represent the compared configurations in the main text, respectively.
The \textbf{bold} values represent the best value across all methods for each metric.
}
\label{tab:bloodmnist:ood}
\end{table}

\begin{table}
\centering
    \begin{sc}
    \begin{adjustbox}{max width=\linewidth}
        \begin{tabular}{|l|cccc|cc|}
            \hline
             \textbf{Method}                           & \textbf{Count}      & \textbf{NLL [nats]}        & \textbf{F1 [0-1]}          & \textbf{CC-ECE [0-1]}      & \textbf{FLOPs [M]}         & \textbf{Params [M]}        \\
             \hline\hline
             \multirow{3}{*}{\textcolor{Gray}{Standard NN $\newmoon$}}              & \multirow{3}{*}{2}  & 0.429 $\pm$ 0.015          & 0.850 $\pm$ 0.007          & 0.107 $\pm$ 0.006          & 1.850 $\pm$ 0.000          & 1.844 $\pm$ 0.000          \\
             &                     & 0.429 $\pm$ 0.015          & 0.850 $\pm$ 0.007          & 0.107 $\pm$ 0.006          & 1.850 $\pm$ 0.000          & 1.844 $\pm$ 0.000          \\
             &                     & 0.429 $\pm$ 0.015          & 0.850 $\pm$ 0.007          & 0.107 $\pm$ 0.006          & 1.850 $\pm$ 0.000          & 1.844 $\pm$ 0.000          \\ \hline
\multirow{3}{*}{\textcolor{Blue}{NN Ensemble $\blacksquare$}}              & \multirow{3}{*}{4}  & 0.437 $\pm$ 0.018          & 0.844 $\pm$ 0.006          & 0.106 $\pm$ 0.002          & \textbf{0.217 $\pm$ 0.000} & 0.215 $\pm$ 0.000          \\
             &                     & 0.451 $\pm$ 0.024          & 0.848 $\pm$ 0.011          & 0.112 $\pm$ 0.003          & 2.111 $\pm$ 0.000          & 2.106 $\pm$ 0.000          \\
             &                     & 0.437 $\pm$ 0.018          & 0.844 $\pm$ 0.006          & 0.106 $\pm$ 0.002          & 0.217 $\pm$ 0.000          & 0.215 $\pm$ 0.000          \\ \hline
\multirow{3}{*}{\textcolor{Cyan}{MCD $\blacktriangle$}}                      & \multirow{3}{*}{20} & \textbf{0.418 $\pm$ 0.017} & 0.828 $\pm$ 0.003          & 0.112 $\pm$ 0.008          & 0.595 $\pm$ 0.000          & \textbf{0.148 $\pm$ 0.000} \\
             &                     & 0.440 $\pm$ 0.030          & 0.849 $\pm$ 0.008          & 0.108 $\pm$ 0.004          & 4.229 $\pm$ 0.000          & 1.053 $\pm$ 0.000          \\
             &                     & 0.678 $\pm$ 0.002          & 0.385 $\pm$ 0.000          & \textbf{0.042 $\pm$ 0.002} & 0.595 $\pm$ 0.000          & 0.148 $\pm$ 0.000          \\ \hline
\multirow{3}{*}{\textcolor{Pink}{BE $\blacktriangleleft$}}           & \multirow{3}{*}{4}  & 0.436 $\pm$ 0.011          & 0.840 $\pm$ 0.010          & 0.109 $\pm$ 0.009          & 0.450 $\pm$ 0.000          & 0.153 $\pm$ 0.000          \\
             &                     & 0.438 $\pm$ 0.018          & 0.848 $\pm$ 0.007          & 0.106 $\pm$ 0.010          & 3.388 $\pm$ 0.000          & 0.861 $\pm$ 0.000          \\
             &                     & \underline{0.444 $\pm$ 0.026}          & \underline{0.845 $\pm$ 0.014}          & \underline{0.103 $\pm$ 0.010}          & \underline{4.774 $\pm$ 0.000}          & \underline{1.604 $\pm$ 0.000}          \\ \hline
\multirow{3}{*}{\textcolor{Orange}{MIMO; $N\geq2, K=1$ \HexagonMarker}}      & \multirow{3}{*}{3}  & 0.431 $\pm$ 0.010          & 0.838 $\pm$ 0.005          & 0.105 $\pm$ 0.010          & 1.418 $\pm$ 0.033          & 1.416 $\pm$ 0.033          \\
             &                     & 0.431 $\pm$ 0.010          & 0.838 $\pm$ 0.005          & 0.105 $\pm$ 0.010          & 1.418 $\pm$ 0.033          & 1.416 $\pm$ 0.033          \\
             &                     & 0.431 $\pm$ 0.010          & 0.838 $\pm$ 0.005          & 0.105 $\pm$ 0.010          & 1.418 $\pm$ 0.033          & 1.416 $\pm$ 0.033          \\ \hline
\multirow{3}{*}{\textcolor{Orange}{MIMMO; $N\geq2, K=D$ \PentagonMarker}}     & \multirow{3}{*}{4}  & 0.419 $\pm$ 0.025          & 0.837 $\pm$ 0.008          & 0.106 $\pm$ 0.009          & 1.660 $\pm$ 0.000          & 1.656 $\pm$ 0.000          \\
             &                     & 0.431 $\pm$ 0.017          & 0.846 $\pm$ 0.005          & 0.104 $\pm$ 0.009          & 2.062 $\pm$ 0.075          & 2.057 $\pm$ 0.075          \\
             &                     & 0.431 $\pm$ 0.017          & 0.846 $\pm$ 0.005          & 0.104 $\pm$ 0.009          & 2.062 $\pm$ 0.075          & 2.057 $\pm$ 0.075          \\ \hline
\multirow{3}{*}{\textcolor{Orange}{SE NN; $N=1, K=1$ \TimesMarker}}        & \multirow{3}{*}{5}  & 0.449 $\pm$ 0.029          & 0.839 $\pm$ 0.016          & 0.109 $\pm$ 0.009          & 1.519 $\pm$ 0.220          & 1.515 $\pm$ 0.219          \\
             &                     & 0.449 $\pm$ 0.029          & 0.839 $\pm$ 0.016          & 0.109 $\pm$ 0.009          & 1.519 $\pm$ 0.220          & 1.515 $\pm$ 0.219          \\
             &                     & 0.449 $\pm$ 0.029          & 0.839 $\pm$ 0.016          & 0.109 $\pm$ 0.009          & 1.519 $\pm$ 0.220          & 1.515 $\pm$ 0.219          \\ \hline
\multirow{3}{*}{\textcolor{Orange}{EE; $N=1, K\geq2$ \OctagonMarker}}        & \multirow{3}{*}{7}  & \textit{0.427 $\pm$ 0.029}          & \textit{0.850 $\pm$ 0.013}          & \textit{0.097 $\pm$ 0.009}          & \textit{2.115 $\pm$ 0.000}          & \textit{2.109 $\pm$ 0.000}          \\
             &                     & 0.429 $\pm$ 0.028          & \textbf{0.853 $\pm$ 0.008} & 0.105 $\pm$ 0.008          & 2.645 $\pm$ 0.000          & 2.638 $\pm$ 0.000          \\
             &                     & 0.427 $\pm$ 0.029          & 0.850 $\pm$ 0.013          & 0.097 $\pm$ 0.009          & 2.115 $\pm$ 0.000          & 2.109 $\pm$ 0.000          \\ \hline
\multirow{3}{*}{\textcolor{Orange}{I/B; $N\geq2, 2\leq K\leq D$ \PlusMarker}} & \multirow{3}{*}{1}  & 0.453 $\pm$ 0.058          & 0.834 $\pm$ 0.018          & 0.104 $\pm$ 0.017          & 4.094 $\pm$ 0.132          & 4.088 $\pm$ 0.132           \\
             &                     & 0.453 $\pm$ 0.058          & 0.834 $\pm$ 0.018          & 0.104 $\pm$ 0.017          & 4.094 $\pm$ 0.132          & 4.088 $\pm$ 0.132           \\
             &                     & 0.453 $\pm$ 0.058          & 0.834 $\pm$ 0.018          & 0.104 $\pm$ 0.017          & 4.094 $\pm$ 0.132          & 4.088 $\pm$ 0.132          \\
\hline
\end{tabular}
\end{adjustbox}
\end{sc}   
\vspace{0.2cm}
\caption{Best results for empirically Pareto optimal configurations on PneumoniaMNIST ID dataset. 
Each row represents the best configuration for each method for the first metric, the second metric, etc.
The count represents how many configurations were in the Pareto front.
The \textit{italic} and \underline{underlined} values represent the compared configurations in the main text, respectively.
The \textbf{bold} values represent the best value across all methods for each metric.
}
\label{tab:pneumoniamnist:id}
\end{table}

\begin{table}
    \centering
        \begin{sc}
        \begin{adjustbox}{max width=\linewidth}
            \begin{tabular}{|l|cccc|cc|}
                \hline
                 \textbf{Method}                           & \textbf{Count}      & \textbf{NLL [nats]}        & \textbf{F1 [0-1]}          & \textbf{CC-ECE [0-1]}      & \textbf{FLOPs [M]}         & \textbf{Params [M]}        \\
                 \hline\hline
                 \multirow{3}{*}{\textcolor{Gray}{Standard NN $\newmoon$}}        & \multirow{3}{*}{2} &     0.558 $\pm$ 0.184      &     0.775 $\pm$ 0.100      &     0.133 $\pm$ 0.041      &     1.585 $\pm$ 0.000      &     1.581 $\pm$ 0.000      \\
                 &                    &     0.558 $\pm$ 0.184      &     0.775 $\pm$ 0.100      &     0.133 $\pm$ 0.041      &     1.585 $\pm$ 0.000      &     1.581 $\pm$ 0.000      \\
                 &                    &     0.558 $\pm$ 0.184      &     0.775 $\pm$ 0.100      &     0.133 $\pm$ 0.041      &     1.585 $\pm$ 0.000      &     1.581 $\pm$ 0.000      \\ \hline
                 \multirow{3}{*}{\textcolor{Blue}{NN Ensemble $\blacksquare$}}        & \multirow{3}{*}{4} &     0.531 $\pm$ 0.195      &     0.783 $\pm$ 0.096      &     0.135 $\pm$ 0.042      &     0.867 $\pm$ 0.000      &     0.861 $\pm$ 0.000      \\
                 &                    &     0.531 $\pm$ 0.195      &     0.783 $\pm$ 0.096      &     0.135 $\pm$ 0.042      &     0.867 $\pm$ 0.000      &     0.861 $\pm$ 0.000      \\
                 &                    &     0.584 $\pm$ 0.247      &     0.775 $\pm$ 0.100      &     0.133 $\pm$ 0.044      & \textbf{0.217 $\pm$ 0.000} &     0.215 $\pm$ 0.000      \\ \hline
                 \multirow{3}{*}{\textcolor{Cyan}{MCD $\blacktriangle$}}            & \multirow{3}{*}{8} &     0.599 $\pm$ 0.415      &     0.779 $\pm$ 0.104      &     0.131 $\pm$ 0.062      &     0.733 $\pm$ 0.000      &     0.182 $\pm$ 0.000      \\
                 &                    &     0.599 $\pm$ 0.415      &     0.779 $\pm$ 0.104      &     0.131 $\pm$ 0.062      &     0.733 $\pm$ 0.000      &     0.182 $\pm$ 0.000      \\
                 &                    &     0.678 $\pm$ 0.005      &     0.385 $\pm$ 0.000      & \textbf{0.049 $\pm$ 0.021} &     0.595 $\pm$ 0.000      & \textbf{0.148 $\pm$ 0.000} \\ \hline
                 \multirow{3}{*}{\textcolor{Pink}{BE $\blacktriangleleft$}}      & \multirow{3}{*}{3} &     \underline{0.547 $\pm$ 0.171}      & \underline{\textbf{0.789 $\pm$ 0.082}} &     \underline{0.126 $\pm$ 0.042}      &     \underline{0.601 $\pm$ 0.000}      &     \underline{0.599 $\pm$ 0.000}      \\
                 &                    &     0.547 $\pm$ 0.171      &     0.789 $\pm$ 0.082      &     0.126 $\pm$ 0.042      &     0.601 $\pm$ 0.000      &     0.599 $\pm$ 0.000      \\
                 &                    &     0.561 $\pm$ 0.210      &     0.788 $\pm$ 0.089      &     0.125 $\pm$ 0.045      &     4.774 $\pm$ 0.000      &     1.604 $\pm$ 0.000      \\ \hline
                 \multirow{3}{*}{\textcolor{Orange}{MIMO; $N\geq2, K=1$ \HexagonMarker}}    & \multirow{3}{*}{2} & \textbf{0.524 $\pm$ 0.223} &     0.777 $\pm$ 0.098      &     0.125 $\pm$ 0.050      &     1.418 $\pm$ 0.033      &     1.416 $\pm$ 0.033      \\
                 &                    &     0.524 $\pm$ 0.223      &     0.777 $\pm$ 0.098      &     0.125 $\pm$ 0.050      &     1.418 $\pm$ 0.033      &     1.416 $\pm$ 0.033      \\
                 &                    &     0.524 $\pm$ 0.223      &     0.777 $\pm$ 0.098      &     0.125 $\pm$ 0.050      &     1.418 $\pm$ 0.033      &     1.416 $\pm$ 0.033      \\\hline
                 \multirow{3}{*}{\textcolor{Orange}{MIMMO; $N\geq2, K=D$ \PentagonMarker}}   & \multirow{3}{*}{4} &     \textit{0.542 $\pm$ 0.299}      &     \textit{0.785 $\pm$ 0.108}      &     \textit{0.128 $\pm$ 0.063}      &     \textit{0.682 $\pm$ 0.000}      &     \textit{0.680 $\pm$ 0.000}      \\
                 &                    &     0.542 $\pm$ 0.299      &     0.785 $\pm$ 0.108      &     0.128 $\pm$ 0.063      &     0.682 $\pm$ 0.000      &     0.680 $\pm$ 0.000      \\
                 &                    &     0.542 $\pm$ 0.299      &     0.785 $\pm$ 0.108      &     0.128 $\pm$ 0.063      &     0.682 $\pm$ 0.000      &     0.680 $\pm$ 0.000      \\ \hline
                 \multirow{3}{*}{\textcolor{Orange}{SE NN; $N=1, K=1$ \TimesMarker}}     & \multirow{3}{*}{3} &     0.637 $\pm$ 0.280      &     0.759 $\pm$ 0.107      &     0.147 $\pm$ 0.051      &     1.519 $\pm$ 0.220      &     1.515 $\pm$ 0.219      \\
                 &                    &     0.637 $\pm$ 0.280      &     0.759 $\pm$ 0.107      &     0.147 $\pm$ 0.051      &     1.519 $\pm$ 0.220      &     1.515 $\pm$ 0.219      \\
                 &                    &     0.637 $\pm$ 0.280      &     0.759 $\pm$ 0.107      &     0.147 $\pm$ 0.051      &     1.519 $\pm$ 0.220      &     1.515 $\pm$ 0.219      \\ \hline
                 \multirow{3}{*}{\textcolor{Orange}{EE; $N=1, K\geq2$ \OctagonMarker}}     & \multirow{3}{*}{2} &     0.587 $\pm$ 0.238      &     0.772 $\pm$ 0.110      &     0.136 $\pm$ 0.053      &     0.715 $\pm$ 0.029      &     0.711 $\pm$ 0.029      \\
                 &                    &     0.587 $\pm$ 0.238      &     0.772 $\pm$ 0.110      &     0.136 $\pm$ 0.053      &     0.715 $\pm$ 0.029      &     0.711 $\pm$ 0.029      \\
                 &                    &     0.588 $\pm$ 0.257      &     0.768 $\pm$ 0.108      &     0.136 $\pm$ 0.045      &     0.798 $\pm$ 0.000      &     0.795 $\pm$ 0.000      \\ \hline
                 \multirow{3}{*}{\textcolor{Orange}{I/B; $N\geq2, 2\leq K\leq D$ \PlusMarker}} & \multirow{3}{*}{1} &     0.592 $\pm$ 0.248      &     0.761 $\pm$ 0.110      &     0.138 $\pm$ 0.051      &     4.094 $\pm$ 0.132      &     4.088 $\pm$ 0.132      \\
                 &                    &     0.592 $\pm$ 0.248      &     0.761 $\pm$ 0.110      &     0.138 $\pm$ 0.051      &     4.094 $\pm$ 0.132      &     4.088 $\pm$ 0.132      \\
                 &                    &     0.592 $\pm$ 0.248      &     0.761 $\pm$ 0.110      &     0.138 $\pm$ 0.051      &     4.094 $\pm$ 0.132      &     4.088 $\pm$ 0.132      \\
    \hline
    \end{tabular}
    \end{adjustbox}
    \end{sc}
    \vspace{0.2cm}
    \caption{Best results for empirically Pareto optimal configurations on PneumoniaMNIST OOD dataset. 
    Each row represents the best configuration for each method for the first metric, the second metric, etc.
    The count represents how many configurations were in the Pareto front.
    The \textit{italic} and \underline{underlined} values represent the compared configurations in the main text, respectively.
    The \textbf{bold} values represent the best value across all methods for each metric.
    }
    \label{tab:pneumoniamnist:ood}
\end{table}

\begin{table}
    \centering
        \begin{sc}
        \begin{adjustbox}{max width=\linewidth}
            \begin{tabular}{|l|ccc|cc|}
                \hline
                 \textbf{Method}                           & \textbf{Count}      & \textbf{NLL [nats]}        & \textbf{MSE}             & \textbf{FLOPs [M]}         & \textbf{Params [M]}        \\
                 \hline\hline
                 \multirow{2}{*}{\textcolor{Gray}{Standard NN $\newmoon$}}        & \multirow{2}{*}{3} &     -0.891 $\pm$ 0.008      &       0.069 $\pm$ 0.001       & \textbf{15.315 $\pm$ 0.000} & \textbf{0.246 $\pm$ 0.000} \\
                 &                    &     -0.843 $\pm$ 0.032      &       0.068 $\pm$ 0.001       &     58.925 $\pm$ 0.000      &     0.912 $\pm$ 0.000      \\ \hline
                 \multirow{2}{*}{\textcolor{Blue}{NN Ensemble $\blacksquare$}}        & \multirow{2}{*}{9} &     \underline{-0.901 $\pm$ 0.015}      &       \underline{0.068 $\pm$ 0.000}       &     \underline{49.949 $\pm$ 0.000}      &     \underline{0.787 $\pm$ 0.000}      \\
                 &                    &     -0.858 $\pm$ 0.017      &  \textbf{0.067 $\pm$ 0.000}   &     220.672 $\pm$ 0.000     &     3.403 $\pm$ 0.000      \\ \hline
                 \multirow{2}{*}{\textcolor{Cyan}{MCD $\blacktriangle$}}            & \multirow{2}{*}{6} &     -0.898 $\pm$ 0.015      &       0.068 $\pm$ 0.000       &     488.560 $\pm$ 0.000     &     1.875 $\pm$ 0.000      \\
                 &                    &     -0.898 $\pm$ 0.015      &       0.068 $\pm$ 0.000       &     488.560 $\pm$ 0.000     &     1.875 $\pm$ 0.000      \\ \hline
                 \multirow{2}{*}{\textcolor{Pink}{BE $\blacktriangleleft$}}      & \multirow{2}{*}{3} &     -0.880 $\pm$ 0.027      &       0.069 $\pm$ 0.001       &     44.969 $\pm$ 0.000      &     0.697 $\pm$ 0.000      \\
                 &                    &     -0.880 $\pm$ 0.027      &       0.069 $\pm$ 0.001       &     44.969 $\pm$ 0.000      &     0.697 $\pm$ 0.000      \\ \hline
                 \multirow{2}{*}{\textcolor{Orange}{MIMO; $N\geq2, K=1$ \HexagonMarker}}    & \multirow{2}{*}{1} &     \textit{-0.896 $\pm$ 0.007}      &       \textit{0.070 $\pm$ 0.001}       &     \textit{47.839 $\pm$ 11.736}     &     \textit{0.699 $\pm$ 0.171}      \\
                 &                    &     -0.896 $\pm$ 0.007      &       0.070 $\pm$ 0.001       &     47.839 $\pm$ 11.736     &     0.699 $\pm$ 0.171      \\ \hline
                 \multirow{2}{*}{\textcolor{Orange}{MIMMO; $N\geq2, K=D$ \PentagonMarker}}   & \multirow{2}{*}{4} &     -0.890 $\pm$ 0.042      &       0.068 $\pm$ 0.001       &     122.704 $\pm$ 0.000     &     1.824 $\pm$ 0.000      \\
                                           &                    &     -0.868 $\pm$ 0.065      &       0.068 $\pm$ 0.001       &     122.704 $\pm$ 0.000     &     1.824 $\pm$ 0.000      \\ \hline
                 \multirow{2}{*}{\textcolor{Orange}{SE NN; $N=1, K=1$ \TimesMarker}}     & \multirow{2}{*}{1} &     -0.886 $\pm$ 0.004      &       0.070 $\pm$ 0.000       &     53.502 $\pm$ 12.028     &     0.817 $\pm$ 0.184      \\
                 &                    &     -0.886 $\pm$ 0.004      &       0.070 $\pm$ 0.000       &     53.502 $\pm$ 12.028     &     0.817 $\pm$ 0.184      \\ \hline
                 \multirow{2}{*}{\textcolor{Orange}{EE; $N=1, K\geq2$ \OctagonMarker}}     & \multirow{2}{*}{3} &     -0.856 $\pm$ 0.011      &       0.067 $\pm$ 0.000       &     88.794 $\pm$ 6.317      &     1.357 $\pm$ 0.096      \\
                 &                    &     -0.856 $\pm$ 0.011      &       0.067 $\pm$ 0.000       &     88.794 $\pm$ 6.317      &     1.357 $\pm$ 0.096      \\ \hline
                 \multirow{2}{*}{\textcolor{Orange}{I/B; $N\geq2, 2\leq K\leq D$ \PlusMarker}} & \multirow{2}{*}{2} & \textbf{-0.910 $\pm$ 0.044} &       0.068 $\pm$ 0.001       &     122.702 $\pm$ 0.000     &     1.823 $\pm$ 0.000      \\
                 &                    &     -0.902 $\pm$ 0.049      &       0.068 $\pm$ 0.001       &     122.702 $\pm$ 0.000     &     1.823 $\pm$ 0.000      \\
    \hline
    \end{tabular}
    \end{adjustbox}
    \end{sc}
    \vspace{0.2cm}
    \caption{Best results for empirically Pareto optimal configurations on RetinaMNIST ID dataset. 
    Each row represents the best configuration for each method for the first metric, the second metric, etc.
    The count represents how many configurations were in the Pareto front.
    The \textit{italic} and \underline{underlined} values represent the compared configurations in the main text, respectively.
    The \textbf{bold} values represent the best value across all methods for each metric.
    }
    \label{tab:retinamnist:id}
\end{table}

\begin{table}
    \centering
        \begin{sc}
        \begin{adjustbox}{max width=\linewidth}
            \begin{tabular}{|l|ccc|cc|}
                \hline
                 \textbf{Method}                           & \textbf{Count}      & \textbf{NLL [nats]}        & \textbf{MSE}             & \textbf{FLOPs [M]}         & \textbf{Params [M]}        \\
                 \hline\hline
                 \multirow{2}{*}{\textcolor{Gray}{Standard NN $\newmoon$}}        & \multirow{2}{*}{3} &     -0.652 $\pm$ 0.742      &       0.084 $\pm$ 0.027       & \textbf{20.145 $\pm$ 0.000} & \textbf{0.320 $\pm$ 0.000} \\
                 &                    &     -0.627 $\pm$ 0.602      &       0.083 $\pm$ 0.030       &     117.167 $\pm$ 0.000     &     1.800 $\pm$ 0.000      \\ \hline
                 \multirow{2}{*}{\textcolor{Blue}{NN Ensemble $\blacksquare$}}        & \multirow{2}{*}{2} &     -0.752 $\pm$ 0.404      &       0.085 $\pm$ 0.028       &     30.630 $\pm$ 0.000      &     0.492 $\pm$ 0.000      \\
                 &                    &     -0.735 $\pm$ 0.468      &       0.083 $\pm$ 0.024       &     40.290 $\pm$ 0.000      &     0.640 $\pm$ 0.000      \\ \hline
                 \multirow{2}{*}{\textcolor{Cyan}{MCD $\blacktriangle$}}            & \multirow{2}{*}{6} &     -0.734 $\pm$ 0.345      &       0.086 $\pm$ 0.032       &     236.272 $\pm$ 0.000     &     0.913 $\pm$ 0.000      \\
                 &                    &     -0.626 $\pm$ 0.542      &       0.083 $\pm$ 0.030       &     468.671 $\pm$ 0.000     &     1.800 $\pm$ 0.000      \\ \hline
                 \multirow{2}{*}{\textcolor{Pink}{BE $\blacktriangleleft$}}      & \multirow{2}{*}{4} &     \underline{-0.746 $\pm$ 0.423}      &       \underline{0.082 $\pm$ 0.023}       &     \underline{159.686 $\pm$ 0.000}     &     \underline{0.642 $\pm$ 0.000}      \\
                 &                    &     -0.746 $\pm$ 0.423      &       0.082 $\pm$ 0.023       &     159.686 $\pm$ 0.000     &     0.642 $\pm$ 0.000      \\ \hline
                 \multirow{2}{*}{\textcolor{Orange}{MIMO; $N\geq2, K=1$ \HexagonMarker}}    & \multirow{2}{*}{1} &     -0.791 $\pm$ 0.202      &       0.084 $\pm$ 0.028       &     47.839 $\pm$ 11.736     &     0.699 $\pm$ 0.171      \\
                 &                    &     -0.791 $\pm$ 0.202      &       0.084 $\pm$ 0.028       &     47.839 $\pm$ 11.736     &     0.699 $\pm$ 0.171      \\ \hline
                 \multirow{2}{*}{\textcolor{Orange}{MIMMO; $N\geq2, K=D$ \PentagonMarker}}   & \multirow{2}{*}{3} &     -0.825 $\pm$ 0.182      &       0.081 $\pm$ 0.026       &     48.747 $\pm$ 0.000      &     0.718 $\pm$ 0.000      \\
                                           &                    &     -0.806 $\pm$ 0.212      &       0.080 $\pm$ 0.025       &     122.704 $\pm$ 0.000     &     1.824 $\pm$ 0.000      \\ \hline
                 \multirow{2}{*}{\textcolor{Orange}{SE NN; $N=1, K=1$ \TimesMarker}}     & \multirow{2}{*}{2} &     -0.725 $\pm$ 0.371      &       0.088 $\pm$ 0.033       &     62.532 $\pm$ 28.371     &     0.955 $\pm$ 0.433      \\
                 &                    &     -0.725 $\pm$ 0.371      &       0.088 $\pm$ 0.033       &     62.532 $\pm$ 28.371     &     0.955 $\pm$ 0.433      \\ \hline
                 \multirow{2}{*}{\textcolor{Orange}{EE; $N=1, K\geq2$ \OctagonMarker}}     & \multirow{2}{*}{7} &     -0.710 $\pm$ 0.305      &       0.084 $\pm$ 0.035       &     68.293 $\pm$ 0.000      &     1.044 $\pm$ 0.000      \\
                 &                    &     -0.624 $\pm$ 0.362      &       0.081 $\pm$ 0.028       &     119.037 $\pm$ 0.000     &     1.821 $\pm$ 0.000      \\ \hline
                 \multirow{2}{*}{\textcolor{Orange}{I/B; $N\geq2, 2\leq K\leq D$ \PlusMarker}} & \multirow{2}{*}{2} & \textit{\textbf{-0.832 $\pm$ 0.175}} &  \textit{\textbf{0.080 $\pm$ 0.025}}   &    \textit{74.009 $\pm$ 0.000}      &     \textit{1.086 $\pm$ 0.000}      \\
                 &                    &     -0.832 $\pm$ 0.175      &       0.080 $\pm$ 0.025       &     74.009 $\pm$ 0.000      &     1.086 $\pm$ 0.000      \\
    \hline
    \end{tabular}
    \end{adjustbox}
    \end{sc} 
    \vspace{0.2cm}
    \caption{Best results for empirically Pareto optimal configurations on RetinaMNIST OOD dataset. 
    Each row represents the best configuration for each method for the first metric, the second metric, etc.
    The count represents how many configurations were in the Pareto front.
    The \textit{italic} and \underline{underlined} values represent the compared configurations in the main text, respectively.
    The \textbf{bold} values represent the best value across all methods for each metric.
    }
    \label{tab:retinamnist:ood}
\end{table}

\clearpage


\clearpage

\begin{table}
    \centering
        \begin{sc}
        \begin{adjustbox}{max width=\linewidth}
            \begin{tabular}{|l|cccccc|cc|}
                \hline
                \textbf{Metric}                    &   \textbf{$i_{\text{start}}(\cdot)$} &   \textbf{$i_{\text{end}}(\cdot)$} &   \textbf{$T_{\text{start}}(\cdot)$} &   \textbf{$T_{\text{end}}(\cdot)$} &   \textbf{$\alpha_{\text{start}}(\cdot)$} &   \textbf{$\alpha_{\text{end}}(\cdot)$} &   \textbf{$N^{c}$} &   \textbf{$K^{c}$} \\
                \hline\hline
                NLL                 &                -0.03 &               0.32 &                 0.27 &               0.03 &                      -0.26 &                   -0.29 &    100.09 &     49.48 \\
                Accuracy            &                  0.24 &              -0.16 &                -0.28 &              -0.07 &                       0.15 &                    0.3  &     38.78 &     61.01 \\
                ECE                 &        0.27 &               0.35 &                 0.16 &               0.06 &                      -0.28 &                   -0.3  &     46.72 &     31.64 \\
                FLOPs &                 0.02 &              -0.12 &                -0.25 &              -0.03 &                       0.28 &                    0.45 &      8.54 &     42.09 \\
                Params                &                0.07 &              -0.1  &                -0.23 &              -0.01 &                       0.26 &                    0.4  &      1.19 &     10.72 \\
                \hline
            \end{tabular}
        \end{adjustbox}
        \end{sc}
    \vspace{0.2cm}
    \caption{Correlation of hyperparameters with metrics on the TinyImageNet ID test set.
    The default correlation measure is Spearman correlation, for the columns with categorical hyperparameters ANOVA is used.
    $i$ denotes the input repetition probability, $T$ the temperature, $\alpha$ the alpha parameter, $N^{c}$ the number of members and $K^{c}$ the maximum number of exits.
    $c$ stands for categorical variables.
    The start and end, denote the start and end value of the hyperparameter, respectively.
    }
    \label{tab:correlation:tinyimagenet:id:test}
\end{table}
\begin{table}
    \centering
        \begin{sc}
        \begin{adjustbox}{max width=\linewidth}
            \begin{tabular}{|l|cccccc|ccc|}
                \hline
                \textbf{Metric}                    &   \textbf{$i_{\text{start}}(\cdot)$} &   \textbf{$i_{\text{end}}(\cdot)$} &   \textbf{$T_{\text{start}}(\cdot)$} &   \textbf{$T_{\text{end}}(\cdot)$} &   \textbf{$\alpha_{\text{start}}(\cdot)$} &   \textbf{$\alpha_{\text{end}}(\cdot)$} &   \textbf{$N^{c}$} &   \textbf{$K^{c}$} &   \textbf{$W^{c}$} \\
                \hline\hline
                F1                   &                 0.31 &               0.33 &                -0.05 &               0.03 &                       0.13 &                    0.06 &      5.01 &      1.49 &     43.7  \\
                CC-ECE              &                --0.3  &              -0.33 &                -0.1  &              -0.1  &                      -0.05 &                   -0.11 &      5.39 &      9.75 &     20.12 \\
                NLL                 &                -0.29 &              -0.35 &                -0.03 &              -0.01 &                      -0.06 &                   -0.08 &      5.81 &      2.97 &     47.86 \\
                FLOPs &                 0.2  &               0.1  &                -0.08 &              -0.09 &                       0.06 &                    0.03 &      0.26 &      1.51 &     68.27 \\
               Params               &                 0.18 &               0.16 &                -0.1  &              -0.09 &                       0.1  &                    0.04 &      1.07 &      0.89 &     50.71 \\
                \hline
            \end{tabular}
        \end{adjustbox}
        \end{sc}
    \vspace{0.2cm}
    \caption{Correlation of hyperparameters with metrics on the BloodMNIST ID test set.
    The default correlation measure is Spearman correlation, for the columns with categorical hyperparameters ANOVA is used.
    $i$ denotes the input repetition probability, $T$ the temperature, $\alpha$ the alpha parameter, $N^{c}$ the number of members, $K^{c}$ the maximum number of exits and $W^{c}$ the width multiplier.
    $c$ stands for categorical variables.
    The start and end, denote the start and end value of the hyperparameter, respectively.
    }
    \label{tab:correlation:bloodmnist:id:test}
\end{table}
\begin{table}
    \centering
        \begin{sc}
        \begin{adjustbox}{max width=\linewidth}
            \begin{tabular}{|l|cccccc|ccc|}
                \hline
                \textbf{Metric}                    &   \textbf{$i_{\text{start}}(\cdot)$} &   \textbf{$i_{\text{end}}(\cdot)$} &   \textbf{$T_{\text{start}}(\cdot)$} &   \textbf{$T_{\text{end}}(\cdot)$} &   \textbf{$\alpha_{\text{start}}(\cdot)$} &   \textbf{$\alpha_{\text{end}}(\cdot)$} &   \textbf{$N^{c}$} &   \textbf{$K^{c}$} &   \textbf{$W^{c}$} \\
                \hline\hline
                F1                 &                 0.01 &              -0.04 &                -0.04 &               0.01 &                       0.01 &                    0.07 &      8    &      2.62 &      3.18 \\
                CC-ECE              &                0.11 &               0.01 &                 0.01 &              -0.02 &                       0    &                   -0.02 &      2.15 &      0.44 &      0.5  \\
                NLL                &               0.1  &               0.02 &                -0.01 &               0    &                       0.01 &                   -0.01 &      4.02 &      2.11 &      2.55 \\
                FLOPs &                -0.01 &              -0.12 &                -0.05 &              -0.1  &                      -0.11 &                    0.13 &      3.95 &      0.59 &    219.72 \\
                Params                &                -0.01 &              -0.12 &                -0.04 &              -0.1  &                      -0.12 &                    0.12 &      3.97 &      0.59 &    219.34 \\
                \hline
            \end{tabular}
        \end{adjustbox}
        \end{sc}
    \vspace{0.2cm}
    \caption{Correlation of hyperparameters with metrics on the PneumoniaMNIST ID test set. 
    The default correlation measure is Spearman correlation, for the columns with categorical hyperparameters ANOVA is used.
    $i$ denotes the input repetition probability, $T$ the temperature, $\alpha$ the alpha parameter, $N^{c}$ the number of members, $K^{c}$ the maximum number of exits and $W^{c}$ the width multiplier.
    $c$ stands for categorical variables.
    The start and end, denote the start and end value of the hyperparameter, respectively.
    }
    \label{tab:correlation:pneumoniamnist:id:test}
\end{table}
\begin{table}
    \centering
        \begin{sc}
        \begin{adjustbox}{max width=\linewidth}
            \begin{tabular}{|l|cccccc|ccc|}
                \hline
                \textbf{Metric}                    &   \textbf{$i_{\text{start}}(\cdot)$} &   \textbf{$i_{\text{end}}(\cdot)$} &   \textbf{$T_{\text{start}}(\cdot)$} &   \textbf{$T_{\text{end}}(\cdot)$} &   \textbf{$\alpha_{\text{start}}(\cdot)$} &   \textbf{$\alpha_{\text{end}}(\cdot)$} &   \textbf{$N^{c}$} &   \textbf{$K^{c}$} &   \textbf{$W^{c}$} \\
                \hline\hline
                NLL                 &                0.33 &               0.12 &                -0.12 &              -0.07 &                      -0.05 &                    0.04 &     53.86 &     11.14 &      5.61 \\
                MSE                 &                0.08 &               0.08 &                -0.03 &               0    &                       0.02 &                   -0.06 &     48.99 &     18.59 &      9.39 \\
                FLOPs &                -0.02 &              -0.01 &                 0.13 &               0.05 &                      -0.11 &                   -0.12 &     69.89 &      9.45 &   1268.52 \\
                Params                &                -0.03 &              -0.02 &                 0.13 &               0.05 &                      -0.11 &                   -0.12 &     70.62 &      9.39 &   1303.6  \\
                \hline
            \end{tabular}
        \end{adjustbox}
        \end{sc}
    \vspace{0.2cm}
    \caption{Correlation of hyperparameters with metrics on the RetinaMNIST ID test set.
    The default correlation measure is Spearman correlation, for the columns with
    categorical hyperparameters ANOVA is used.
    $i$ denotes the input repetition probability, $T$ the temperature, $\alpha$ the alpha parameter, $N^{c}$ the number of members, $K^{c}$ the maximum number of exits and $W^{c}$ the width multiplier.
    $c$ stands for categorical variables.
    The start and end, denote the start and end value of the hyperparameter, respectively.
    }
    \label{tab:correlation:retinamnist:id:test}
\end{table}

\end{document}